\documentclass[10pt,twocolumn,letterpaper]{article}

\usepackage{cvpr}

\usepackage[table]{xcolor}

\newcommand{\todo}[1]{{\color{red}#1}}

\definecolor{cvprblue}{rgb}{0.21,0.49,0.74}
\usepackage[pagebackref,breaklinks,colorlinks,citecolor=cvprblue]{hyperref}

\usepackage{algorithm}  
\usepackage{algpseudocode}  
\usepackage{amsfonts}       
\usepackage{amsmath}       
\usepackage{amssymb}
\usepackage{pifont}
\usepackage{nicefrac}       
\usepackage{microtype}

\usepackage{multirow}
\usepackage{nicefrac}
\usepackage{float}
\usepackage{caption}
\usepackage{subcaption}
\newcounter{cw}

\newcommand\yb{\mathbf{y}}
\newcommand\tb{\mathbf{t}}

\newcommand\vb{\mathbf{v}}
\newcommand\xb{\mathbf{x}}
\newcommand\Xb{\mathbf{X}}

\newcommand\Pb{\mathbf{P}}

\newcommand\Gb{\mathbf{G}}
\newcommand\ub{\mathbf{u}}

\newcommand\Rb{\mathbf{R}}

\newcommand\db{\mathbf{d}}

\newcommand\rb{\mathbf{r}}
\newcommand\nb{\mathbf{n}}

\DeclareMathOperator*{\argmin}{arg\,min}

\definecolor{myred}{rgb}{1,0.3,0.3}    
\definecolor{myorange}{rgb}{1,0.7,0.4}  
\definecolor{myyellow}{rgb}{1,1,0.5}

\newcommand{\xmark}{\ding{55}}
\newcommand{\myparagraph}[1]{\vspace{0.2cm}\noindent {\bf #1.}}

\title{Fine Dense Alignment of Image Bursts \\ through Camera Pose and Depth Estimation}

\author{
Bruno Lecouat$^{1,2,\ast}$ $\: \: \: \:$
Yann Dubois de Mont-Marin$^{2,\ast}$  $\: \: \: \:$
Théo Bodrito$^{2,}$\thanks{Authors marked with an asterisk (*) contributed equally to this work. This project started when B. Lecouat was a PhD student at Inria Paris.} \\
Julien Mairal$^{4}$ $\: \: \: \:$
Jean Ponce$^{2,3}$\\
}

\begin{document}

\maketitle

\begin{abstract}

This paper introduces a novel approach to the fine alignment of images
in a burst captured by a handheld camera. In contrast to traditional techniques that
estimate two-dimensional transformations between frame pairs or rely on
discrete correspondences, 
the proposed algorithm establishes dense correspondences by optimizing
both the camera motion and surface depth and orientation at every pixel.
This approach improves alignment,
particularly in scenarios with parallax challenges.
Extensive experiments with synthetic bursts featuring small and even
tiny baselines demonstrate that it outperforms the best
optical flow methods available today in this setting, {\em without}
requiring any training. 
Beyond enhanced alignment, our
method opens avenues for tasks beyond simple image restoration, such as depth estimation and 3D reconstruction, as supported by promising preliminary results. 
This positions our approach as a versatile tool for various burst
image processing applications.
\end{abstract}

\addtocounter{footnote}{0} 
\stepcounter{footnote}\footnotetext{Enhance Lab.}
\stepcounter{footnote}\footnotetext{Inria and DIENS (ENS-PSL, CNRS, Inria).}
\stepcounter{footnote}\footnotetext{Courant Institute and Center for Data Science, New York University.}
\stepcounter{footnote}\footnotetext{Univ. Grenoble Alpes, Inria, CNRS, Grenoble INP, LJK.}
\stepcounter{footnote}\footnotetext{Corresponding author: bruno.lecouat@enhancelab.fr}

\section{Introduction}

This paper tackles the challenge of dense alignment in burst photography, a domain characterized by minimal camera movement and predominantly static scenes. We aim to align these image sequences accurately, quickly, and reliably. 

Burst photography is increasingly pivotal in a range of image enhancement applications, as evidenced by recent advancements in high dynamic range imaging \cite{hasinoff2016burst,lecouat2022high}, night photography \cite{liba2019handheld}, deblurring \cite{delbracio2015burst}, or super-resolution \cite{wronski2019handheld,lecouat2021lucas,bhat2021deep}. In this context, a handheld camera captures a rapid sequence of images with slightly different viewpoints due to hand tremor, possibly with varying camera settings, over a brief duration. 
The alignment of these frames is a critical precursor for these methods. However, current approaches to image registration between image pairs, such as homography or optical flow estimation, do not fully leverage the nature of burst sequences (multiple views of a quasi-static three-dimensional scene with slight camera motion). This limitation potentially leads to suboptimal outcomes. Precision in alignment is crucial for the quality of the enhanced images, and inaccuracies can significantly impair the final results, introducing artifacts like ghosting or zipping \cite{lecouat2021lucas}.

In this paper, instead of relying on traditional pairwise dense alignment of frames, we propose a novel global estimation approach tailored for image bursts, which explicitly considers the three-dimensional nature of the scene.
Specifically, our approach takes full advantage of the small baseline feature by introducing a new parametrization of optical flows, consistent across different views, based on the image formation model. 
This model assumes a perspective camera with known intrinsic parameters, capturing a static scene comprising surfaces approximated as small planar patches. Given the small baseline, we anticipate minimal occlusions between views. Consequently, we simplify the representation of the 3D scene into a concise two-dimensional grid that encodes the depth and normals of these planar surfaces.

\begin{figure*}[htpb]
    \centering
    \vspace{-0.2cm}
    \includegraphics[width=0.95\textwidth]{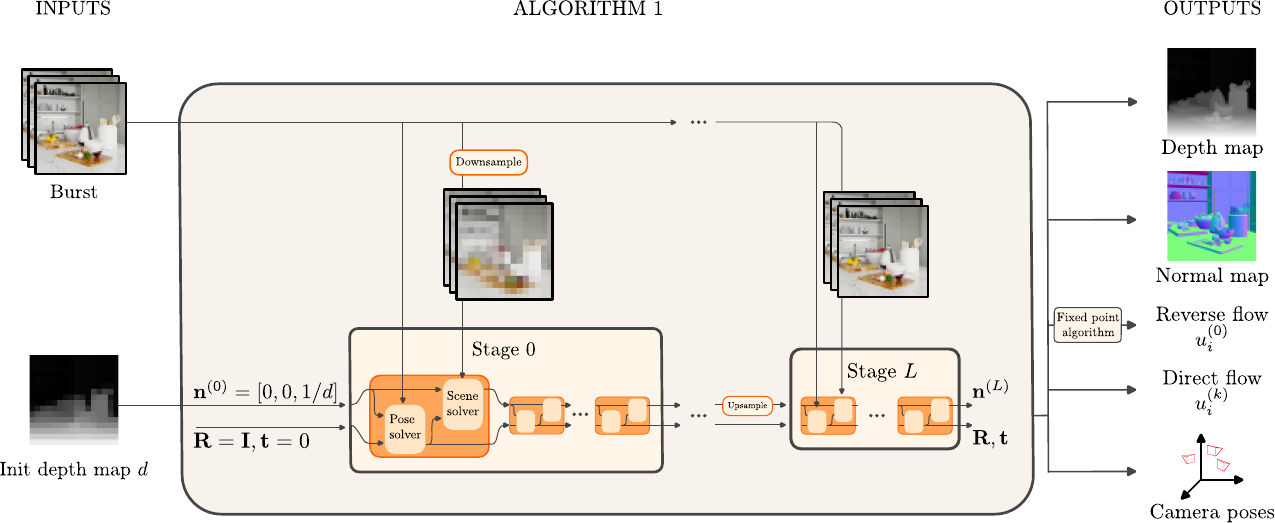} 
    \caption{The global pipeline of our optimization-based method. It inputs a burst of images and an initialization depth map and outputs the direct and reverse flow between each image and the first one. Our method estimates the optical flow using the camera's pose and 3D scene structure as optimization variables of the photometric reprojection errors in a reference frame then poses and depth maps can also be retrieved. }\label{fig:pipeline}
    \vspace{-0.5cm}
\end{figure*}

More precisely, our method employs a 2D grid to represent depth, normals from a reference view, and camera poses. While camera poses are optimized individually for each frame, structural parameters are shared across all views. This shared parameterization requires fewer parameters than traditional pairwise optical flow methods. It enhances the overall consistency and effectiveness of our alignment method while still preserving the expressivity necessary for accurately modeling motion induced from 3D scenes.

In practice, we solve a global optimization problem to align frames, minimizing patch photometric reprojection errors across all views within the reference frame. Optimizing for camera pose, depth, and normal parameters. In situations with parallax, our model adapts to determine camera motion and scene geometry that accounts for the relative movements between frames. When no parallax effects are present, the model defaults to fitting pose parameters for each frame with constant depth, similar to homography fitting.

To achieve efficient optimization, we propose a new coarse-to-fine block-coordinate descent algorithm inspired by the parametric Lucas-Kanade algorithm \cite{baker2004lucas} in its structure, using a variant of the Gauss-Newton algorithm for precise pose optimization on $SO(3)$ and gradient descent for depth and normal adjustments. We also introduce a novel fixed-point algorithm to infer depth maps for new camera positions. This algorithm is particularly advantageous for our specific needs but also holds potential for broader applications. It enables us to calculate reverse optical flows and adapt reference views to other views, which is essential for downstream tasks like super-resolution and low-light photography and can also be used to detect occlusions.

We validate our approach with synthetic bursts built with photorealistic rendering software. To validate our approach with real-world data, we also demonstrate applications with real bursts captured with a Pixel 6 pro smartphone to night photography denoising and super-resolution.
Quantitative and qualitative experiments with synthetic and real data show that our method consistently gives accurate registration results even when little or no parallax is present and consistently outperforms the state-of-the-art in the burst setting, outperforming learning-based methods for flow estimation such as RAFT \cite{teed2020raft}.

Beyond flow estimation, our model demonstrates exceptional versatility and efficiency in small baseline scenarios. It not only achieves convergence in pose and depth to meaningful values but also surpasses specialized methods in these areas. Distinct from conventional 3D methods that typically separate camera pose estimation and dense reconstruction into different steps, our method directly tackles dense optimization within a joint estimation framework. 

Essentially, our approach acts as a multifunctional tool in burst photography, with the dual capability to accurately estimate flow and precisely determine depth and pose. It is helpful across a wide range of downstream tasks and sets a new benchmark for processing small motion scenes---characterized by its simplicity, accuracy, and robustness.

\myparagraph{Contributions}
Below, we summarize our key contributions, highlighting how our approach serves as a versatile tool applicable to various burst image processing tasks:
\begin{enumerate}
    \item \textbf{State-of-the-art dense alignment for burst imagery}: we propose a novel optimization algorithm that outperforms deep-learning methods in dense alignment. This precision is especially useful for tasks requiring fine alignments, like burst super-resolution.
    \item \textbf{Accurate pose and depth estimation in small motion}: our algorithm provides state-of-the-art camera pose and depth estimation results in scenarios with minimal motion, effectively capturing 3D scene structures from bursts with small baselines. This performance is achieved where standard SFM methods such as COLMAP~\cite{schonberger2016structure} struggle.
    \item \textbf{Novel fixed-point algorithm for depth inference}: we propose a new fixed-point algorithm for deducing depth maps at novel camera positions, enhancing our method's utility in reversing optical flows and warping reference views onto other views, with potential applications beyond the scope of this paper.
\end{enumerate}

\section{Related work}
\paragraph{Burst photography.}

Burst photography is a technique that involves capturing a sequence of images to improve the overall quality of a photograph by reducing noise~\cite{hasinoff2016burst}, enhancing details~\cite{wronski2019handheld,mehta2023gated,lecouat2021lucas,bhat2021deep,luo2022bsrt,bhat2021deep2,dudhane2023burstormer}, and improving dynamic range~\cite{lecouat2022high}. Traditionally, algorithms for burst photography rely on a registration step to align frames~\cite{delbracio2021mobile}.

Recent advancements explore machine learning, specifically deep learning, for burst photography, often eliminating the need for traditional registration~\cite{luo2022bsrt,dudhane2023burstormer}. However, many such algorithms are based on supervised learning, demanding paired datasets of degraded raw bursts and high-quality sRGB images for training. The reliance on simulated raw bursts generated from ground truth sRGB images introduces a potential mismatch between training and real-world data distributions. Real-world bursts may exhibit different degradations or involve a sensor mismatch, leading to artifacts~\cite{bhat2023self}. Self-supervised learning methods~\cite{bhat2023self,nguyen2021self} have emerged to address these issues.
Furthermore, the computational demands of deep learning models pose challenges for integration into embedded devices~\cite{delbracio2021mobile}, limiting their practical utility under constraints of limited ressources.

In contrast, we present an efficient approach to image alignment specifically designed for burst image data. This approach does not rely on machine learning and can serve as a versatile tool in various burst processing applications, whether they involve learning-based components or not.

\myparagraph{Multi-frame image registration} 

A straightforward method for image alignment in burst photography involves aligning frames with a reference frame, as demonstrated in \cite{wronski2019handheld,hasinoff2016burst}. Some works have explored the multi-view setting to enhance registration quality, such as \cite{aguerrebere2016fundamental, farsiu2005constrained, aguerrebere2018practical}, which introduced various optimization-based approaches for multi-view image registration. However, these approaches are limited to simple motion models, such as translations. In contrast, our method is more general and takes into account the three-dimensional nature of the scene.

\myparagraph{Depth reconstruction from small motions}
Popular 3D reconstruction methods rely on geometric approaches such as structure from motion (SfM) \cite{schonberger2016structure}. These methods use geometric constraints and depend on keypoint correspondences to reconstruct a sparse 3D scene. Subsequently, dense 3D representations can be estimated based on the sparse reconstruction, as done by Colmap~\cite{schonberger2016structure}. Bundle adjustment is a critical step for refining the estimated 3D structure and camera poses of a scene. This process involves optimizing 2D image keypoints, corresponding 3D points, and camera calibration parameters iteratively to minimize the reprojection error, leading to a more accurate scene reconstruction.

Several 3D reconstruction methods have been specifically tailored for scenarios involving small motions to reconstruct depth maps. 
For instance, Im et al. \cite{im2015high} have adapted SfM to small motion settings, whereas \cite{ha2016high} have proposed an efficient method using feature tracking for pairwise key points and bundle adjustment algorithms adapted to small motions.
Additionally, this method estimates the intrinsic parameters of the camera as well as distortion parameters to achieve a better fit with the data.
In a different approach, \cite{chugunov2022implicit} introduces a neural depth model and uses an inertial measurement unit (IMU) and lidar measurements to respectively initialize camera poses and the depth map. Then, \cite{chugunov2022shakes} eliminates the need to initialize with a depth map model, although initialization with such a model may still yield improved results.

In contrast, our method serves a different purpose than depth estimation, with our primary goal being accurate image alignment. As shown in Sec.~\ref{sec:experiment}, our dense depth estimation procedure is more suitable for this task than approaches based on bundle adjustment with sparse keypoints.

\section{Method}\label{sec:meth}

The proposed method aims to robustly and accurately estimate the optical flow and its inverse between a reference image and other images within a burst sequence. Given the nature of a burst, where movements are small, this approach provides the opportunity to directly address the problem densely, in contrast to \cite{ha2016high}, which relies on prior sparse matching between pairs of views. Densely approaching the problem enables high flow accuracy compared to other existing methods. In order to address the problem both densely and robustly, the key idea of the method is to parameterize the flows for each view using a common dense map characterizing the scene in the reference view and the relative positions of the views with respect to the reference frame. 

Our formation model is detailed in the first paragraph below, leading to the formulation of flow estimation by optimizing the dense structure map and the relative positions of the views. This optimization is achieved by minimizing the photometric reprojection error through the direct flow induced by the parameters, which is the loss that best characterizes the quality of the induced flow. The challenges of this minimization problem are outlined in the second paragraph. The minimization procedure uses a block coordinate descent between the dense structure map and the relative poses, described in the third paragraph. This approach stabilizes the optimization process.

It also enables a coarse-to-fine approach for the dense parameterization of the scene. Finally, our formation model also allows the calculation of inverse optical flow through a fixed-point algorithm, detailed in the fourth paragraph. The global pipeline is illustrated in Fig. \ref{fig:pipeline}.

\begin{figure}
    \centering
        \includegraphics[trim=40 17 0 0, clip, width=0.6\textwidth]{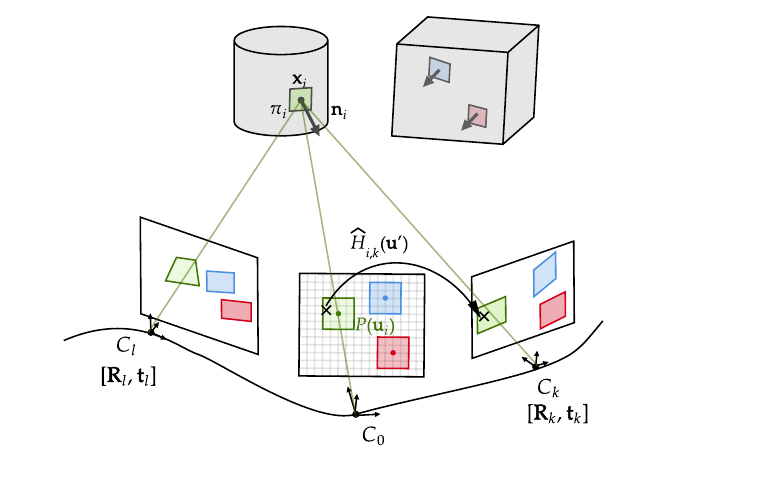}
    \caption{Image formation model with a patch and its local homography flow.}\label{fig:formation_model}
    \vspace{-0.5cm}
\end{figure}

\myparagraph{Image formation model}
We consider a rigid scene described by a piecewise surface and $K+1$ internally calibrated pinhole cameras $(C_k)_{k=0..K}$.
A point $\ub_i$ in $\Gb$ a regular grid of the $C_0$ camera plane, is the projection of a point $\xb_i$ of the scene surface. We denote by $\pi_i$, the affine plane tangent to the scene in $\xb_i$ parameterized, by its (non-unit) normal $n_i$ such that $\pi_i=\{\yb\in\mathbb{R}^3, \nb_i^\top\yb=1\}$. A patch $P(\ub_i)$ around $\ub_i$ is the projection of a patch around $\xb_i$ in $\pi_i$, and its image in the camera plane $C_k$ is given by a homography uniquely defined by the plane $\pi_i$ and the extrinsic parameters $\Rb_k, \tb_k$ of the other camera (Fig. \ref{fig:formation_model}).
For $\ub'$ in the patch $P(\ub_i)$, we have the direct flow locally expressed as a homography:
\begin{align}
    \hat{H}_{i,k}(\ub') &= \psi(H_{i,k} [\ub',1]^\top)\\
    H_{i,k} &= \Rb_k + \tb_k \nb_i^\top,\label{eq:homo}
\end{align}
where $H_{i,k}$ is the homography matrix for the patch $P(\ub_i)$ in the camera plane of $C_k$, $[\ub',1]$ is the homogeneous representation of $\ub'$ and $\psi: x,y,z\rightarrow x/z, y/z$ is the standard projection. The parameters of this flow are the non-unit normal $\nb_i$ characterizing the plane $\pi_i$ and the pose $\Rb_k, t_k$. It is important to note, as in \cite{Hartley_book, chugunov2022shakes}, that  $\nb_i$ is not a homogeneous vector defined up to scale and has three full degrees of freedom. The formation model is summed up in Fig. \ref{fig:formation_model}.

\myparagraph{Minimization problem}
The parameters of our formation model is $n=(\nb_i)_{i \in G}$ the dense map over a regular grid $\Gb$ parametrizing the scene structure and $R,t = (\Rb_k, t_k)_{k=1..K}$ the pose parameters of each $C_k$ relative to $C_0$. As the objective is to estimate the optical flow between view $k$ and the reference view, we optimize the parameters $\mathbf{R}, \mathbf{t}, \mathbf{n}$ so that the flows derived from local homographies $H_{i,k}$ minimize the photometric reprojection error between the images $I_0$ and $I_k$. More specifically, we solve the minimization problem:
\begin{equation}
   \min_{\nb,\Rb,\tb}  \frac{1}{2}\sum_{k=1}^K\sum_{i\in \Gb}\sum_{\ub'\in 
 \Pb(\ub_i)} \rho(|I_0(\ub') - I_k(\hat{H}_{i,k}(\ub'))|^2)\label{eq:min_prob},
\end{equation}
where $\rho$ is a robust loss function as in \cite{ipolinverse}. Indeed, the formation model does not account for occlusion phenomena. When a pixel $\mathbf{u}'$ in the $C_0$ plane is the projection of a point $\mathbf{x}$ that is not visible in camera $C_k$, $\hat{H}_{i,k}(\mathbf{u}')$ is essentially the projection of another point $y$ that is not on the same scene element as $x$. Consequently, it is likely that the color $I_k(\hat{H}_{i,k}(\mathbf{u'}))$ deviates significantly from $I_0(\mathbf{u})$. The function $\rho$ reduces the importance of large values, effectively filtering out such cases.

\myparagraph{Optimization procedure}
As usual in structure from motion litterature \cite{Hartley_book}, there is a global scale ambiguity since for every $\lambda>0$, jointly replacing $\tb$ by $\lambda \tb$ and $\nb$  by $1/\lambda\nb$
does not change the homography matrices $H_{i,k}$ and nor the loss.
To prevent this ambiguity from hindering the convergence of the optimization procedure, our algorithm relies on a \emph{block coordinate descent} that alternates between steps on the plane map $\mathbf{n}$ and steps on the relative poses $\mathbf{R}, \mathbf{t}$. Indeed, when $\mathbf{n}$ is fixed, there is no longer any ambiguity about the value that $\mathbf{t}$ can take, and the same applies to $\mathbf{n}$ when $\mathbf{t}$ is fixed. Gradually, the ambiguity boiles down to the scale induced by the parameters' initialization. In addition, the optimization problem \eqref{eq:min_prob} is not convex and a good initialization is crucial to enable the algorithm's convergence.
In the case of small movements, it is reasonable to initialize the pose with $\Rb_k=I$ and $\tb_k=0$. Therefore, it is necessary to have a good initialization of the plane parameters. Our method relies on an initialization based on a very coarse and low-resolution estimation of the scene depth in the reference image. Starting from a depth map $\mathbf{z} = (z_i)_{i \in \Gb^{(0)}}$ on a very low-resolution grid $\Gb^{(0)}$ (typically $16 \times 16$), we can initialize the plane map as $n^{(0)}=([0,0,1/z_i])_{i \in \Gb^{(0)}}$. This corresponds to initializing the planes as fronto-parallel and located at a distance $z_i$ from the reference camera. It is important to note that this initialization resolves the scale ambiguity and initializes in a good region, thereby avoiding certain local minima. However, it does not need to be extremely precise. As observed in Section \ref{sec:experiment}, the performance of our method is minimally impacted by the quality of the initialization. This paper uses the smallest monocular network of shallow resolution from \cite{ranftl2020towards}, which has negligible inference cost, to initialize the algorithm.

From this initialization, we adopt a \emph{coarse-to-fine} strategy as in \cite{Lei2009, ipolinverse} for optimizing the plan map. Specifically, we define a sequence $\Gb^{(0)}, \ldots, \Gb^{(L)}$ of $L$ regular grids, each twice as fine as the previous one, with $\Gb^{(L)}$ having the same resolution as the burst $(I_k)_{k=0..K}$. We also denote $I_k^{(l)}$ as the downsampled version of the image $I_k$ to the resolution of $\Gb^{(l)}$. Our optimization strategy is as follows:
\begin{itemize}
    \item We perform the steps for poses $\mathbf{R},\mathbf{t}$ using the high-resolution grid $\Gb^{(L)}$, a linear interpolation of the current estimate of the plane map $\nb^{(l)}$ to the resolution of $\Gb^{(L)}$, and using the high-resolution images $(I_k)_{k=0..K}$. For these steps, we employ a proximal Gauss-Newton algorithm tailored to the fact that rotation matrices belong to the Lie group $SO(3)$ and the minimization problem \eqref{eq:min_prob} is a robust nonlinear least squares problem \cite{nocedal}. The small number of variables (six times the number of images) makes the computation of the required Jacobians tractable. Details about the Gauss-Newton step and the closed form of the Jacobians are provided in Appendix \ref{app:cf_jac}. Using the Lie group exponential representation of rotation and employing a second-order optimization method are crucial elements of our method for achieving high precision. We empirically show the advantages of these choices in the ablation study presented in Appendix \ref{app:abl_stdy}.
    \item We perform the steps on the plane map parameters $\mathbf{n}^{(l)}$ at scale $l$ using the gradient descent variation \textbf{Adam} \cite{kingma2014adam}, with the loss calculated using the grid $\Gb^{(l)}$, $\mathbf{R},\mathbf{t}$, and the images at resolution $l$: $(I_k^{(l)})_{k=0..K}$. Using a method with moments like Adam accelerates the convergence of the procedure.
    \item Every few alternate steps on $\mathbf{R},\mathbf{t}$ on one side and $\mathbf{n}^{(l)}$ on the other side, we double the resolution of the plane map $\mathbf{n}^{(l)}$ and move to the next scale with $\mathbf{n}^{(l+1)}$.
\end{itemize}
The procedure is summarized in the pseudocode in Algorithm \ref{alg:minimization}. In the case of a dense approach, a coarse-to-fine strategy is crucial. Since we use a photometric loss, the gradients and Jacobians depend on the spatial gradients of the $I_k$ images and contain only sub-pixel information. When the alignment error is larger than one pixel, this can lead to convergence issues. At lower scales of the coarse-to-fine approach, pixels cover a larger area, allowing us to benefit from the information. As we move to higher scales, we increase the precision we aim to achieve. Finally, note that the original minimization problem is properly solved during the last stage of the coarse-to-fine approach. The previous stages can be interpreted as a procedure to generate the right initialization for the original minimization problem.
\begin{algorithm}
\small
\caption{\small 
Multiscale block coordinate descent}\label{alg:meta}
\begin{algorithmic}[1]
\Require $L \geq 0$, $N \geq 0, \beta$
\Require $I=(I_k)_{k=0..K}, d$  \Comment{Burst and $16\times16$ resolution depth map}
\State $d \gets \operatorname{Mono}(I_0)$ \Comment{Low resolution monocular depth estimation}
\State $n \gets [0,0,1/d]$ \Comment{dimension $16\times16$}
\State $R \gets I$ \Comment{$K$ matrices $3\times3$}
\State $t \gets 0$ \Comment{$K$ vectors of size $3$}
\State $l \gets 0$
\While{$l \leq L$} \Comment{Multiscale loop}
\State $n \gets \operatorname{interpole}(n,2\times\operatorname{resolution}(n))$ \Comment{Double resolution}
\State $I_{-} \gets \operatorname{sample}(I,\operatorname{resolution}(n))$ \Comment{Low resolution image}
\State $m \gets 0$
\While{$m \leq M$} \Comment{Block descent}
\State $n_{+} \gets \operatorname{interpole}(n,\operatorname{resolution}(I))$ \Comment{High resolution}
\State $R,t \gets \operatorname{PGN}(R,t,n_{+}, I)$\Comment{Pose Newton step}
\State $n \gets \operatorname{ADAM}(R,t,n,I_{-})$\Comment{Some steps with ADAM}
\State $m \gets m+1$
\EndWhile
\State $l\gets l+1$
\EndWhile
\State \Return $R,t,n$
\end{algorithmic}
\label{alg:minimization}
\end{algorithm}

\myparagraph{Outputting the flows, poses, depth map and normal map}

\begin{figure}
    \centering
        \includegraphics[trim=50 0 20 0, clip,width=0.5\textwidth]{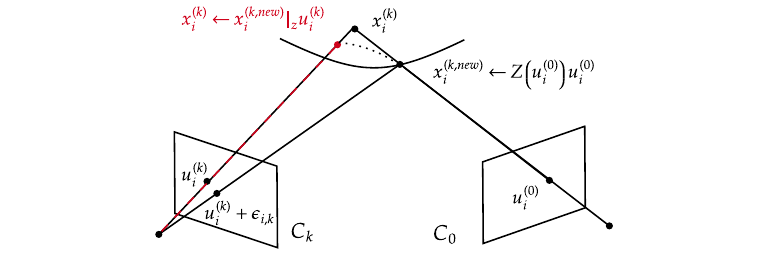}
    \caption{We have the depth map $(z^{(0)}_i)_{i\in\Gb^{(L)}}$ in the reference view $(C_0)$, with interpolation we construct $Z$ that gives the depth for any $\ub_0$. We initialize a depth map $(z^{(k)}_i)_{i\in\Gb^{(L)}}$ in view $k$ with a copy of the depth map in view $0$. Then, we use $z^{(k)}$ on the regular grid to induce a direct flow into the view $0$ and query the depth $Z(F(\ub_i^{(k)}))$. Reprojecting the obtained depth map gives the new iterate of $z^{(k)}$. At convergence, the direct flow induced by $z^{(k)}$ is the reverse flow from view $0$ to view $k$.}\label{fig:fp_model}
\end{figure}
After the convergence of the algorithm, we obtain $\mathbf{R}, \mathbf{t}$, and $\mathbf{n}^{(L)}$ that minimize the problem \eqref{eq:min_prob} on a grid of maximum resolution, thus minimizing the photometric error of the flow.
We then obtain an estimation of the \textbf{direct flow} for each image. For $\mathbf{u}_i^{(0)} \in \Gb^{(L)}$:
\begin{equation}
\mathbf{u}_i^{(k)} = \hat{H}_{i,k}(\mathbf{u}_i^{(0)}) = \psi(\mathbf{R}_k \mathbf{u}_i^{(0)} + 1/z_i^{(0)} \mathbf{t}_k),\label{eq:depthmap}
\end{equation}
where, from $\mathbf{n}^{(L)}$, we recover a high-resolution depth map in the reference view: $z_i^{(0)} = 1 / ([\mathbf{u}_i^{(0)}, 1]^\top\mathbf{n}_i^{(L)})$. Normalizing $\mathbf{n}^{(L)}$ also provides a high-resolution normal map.
Finally, our algorithm directly estimates the camera poses $\mathbf{R}, \mathbf{t}$.
For certain applications, such as super-resolution, we need the inverse flow, i.e., $\mathbf{u}_i^{(0)} = F(\mathbf{u}_i^{(k)})$. The flow inverse is generally unstable, so PyTorch~\cite{pytorch} does not implement the forward warp. As the movements are small for a burst, the depth map in view $C_k$ will be close to that in view $C_0$. Moreover, the depth map in view $C_k$ allows generating a direct flow $\mathbf{u}_i^{(0)} = F(\mathbf{u}_i^{(k)})$ as in \eqref{eq:depthmap} using the inverse poses: $[\mathbf{R}_k^\top, -\mathbf{R}_k^\top\mathbf{t}_k]$. So, if the depth map is correct, we should obtain the identity by composing with the initial direct flow. This allows designing a fixed-point algorithm presented in Fig.~\ref{fig:fp_model} and detailed in Appendix \ref{app:fp}. Pixels for which the fixed point does not converge correspond to the occluded pixel, and the occlusion masks are presented in Appendix \ref{app:occ}.

\begin{table*}[htbp]
    \centering
    \small
    \begin{tabular}{lccccc|ccccc}
    \toprule
Method   
& \begin{tabular}[c]{@{}c@{}}EPE  \\ $\downarrow$ \end{tabular} &\begin{tabular}[c]{@{}c@{}}RMSE  \\ $\downarrow$ \end{tabular} & 	\begin{tabular}[c]{@{}c@{}} NPE1  \\ $\uparrow$ \end{tabular} &\begin{tabular}[c]{@{}c@{}} NPE2  \\ 
$\uparrow$ \end{tabular} &	\begin{tabular}[c]{@{}c@{}} NPE3  \\ $\uparrow$ \end{tabular}

& \begin{tabular}[c]{@{}c@{}}EPE  \\ $\downarrow$ \end{tabular} &\begin{tabular}[c]{@{}c@{}}RMSE  \\ $\downarrow$ \end{tabular} & 	\begin{tabular}[c]{@{}c@{}} NPE1  \\ $\uparrow$ \end{tabular} &\begin{tabular}[c]{@{}c@{}} NPE2  \\ 
$\uparrow$ \end{tabular} &	\begin{tabular}[c]{@{}c@{}} NPE3  \\ $\uparrow$ \end{tabular}
\\ \midrule

 & \multicolumn{5}{c}{Blender 1 (small motion)} & \multicolumn{5}{c}{Blender 2 (micro motion)}\\  \midrule
DfUSMC \cite{ha2016high} *&  1.4466	&2.1723&0.5315	&0.7488&	0.8477 & 4.1356&4.5676	&0.2267	&0.4278	&0.5497 \\
RCVD \cite{kopf2021robust}*& 5.9556	&7.678	&0.0957&	0.2534&	0.3763 & {0.4007}	& {0.5316}	& {0.8676} &	0.9825&	0.9959\\
Saop \cite{chugunov2022shakes} *&  9.7262	&12.5891	&0.101	&0.2457	&0.3402 & 2.0430	&2.3563	&0.5684	&0.7645	&0.8424  \\ \midrule

Homography   &  2.8102	& 4.7107 &	0.4998 &	0.6627 &	0.7405 & \underline{0.3008}	& \underline{0.3772} &	\underline{0.9003} &	\underline{0.9921} &	\underline{0.9982}\\

Farnebäck  \cite{farneback2003two} &  2.6852	& 4.8478 &	0.5299 &	0.6612 &	0.7278 & 2.0892	& 3.8154 &	0.6480 &	0.7296 &	0.7642\\
RAFT \cite{teed2020raft}  & \underline{0.9013}&	\underline{1.5396}	&\underline{0.7348}&	\underline{0.9069}	&\underline{0.9443}   & 0.4857&	0.5765	&0.8664	& {0.9857}	&{0.9963} \\

Ours   &\textbf{0.7439 }& \textbf{1.4324}  &\textbf{0.7841} &\textbf{ 0.9084 } &\textbf{0.9456} & \textbf{0.2321} & \textbf{0.2820} &\textbf{ 0.9366} & \textbf{0.9972} & \textbf{1.0000}  \\

\bottomrule
\end{tabular}
\caption{Optical flow errors. The optical flow was predicted from the extrinsic camera parameters and depth maps for the models marked with an asterisk.}
\label{tab:flow_err}
\end{table*}

\section{Experiments}\label{sec:experiment}

\begin{table*}[htp]
    \centering
    \small
    \begin{tabular}{l|cccc|cccccc}
    \toprule
 &\multicolumn{4}{c}{Pose } &\multicolumn{6}{c}{Depth } \\ 

Method  
& \begin{tabular}[c]{@{}c@{}}Left l2 \\ (m)$\downarrow$ \end{tabular}& \begin{tabular}[c]{@{}c@{}}ATE  \\ (m) $\downarrow$ \end{tabular} &	\begin{tabular}[c]{@{}c@{}}Geom  \\ (m) $\downarrow$ \end{tabular}  &	 \begin{tabular}[c]{@{}c@{}}Biinvrot l2  \\ (deg) $\downarrow$ \end{tabular}    
& Abs rel $\downarrow$	&Sqr rel $\downarrow$&	RMSE$\downarrow$	&Delta 1$\uparrow$	&Delta 2	$\uparrow$&Delta 3  $\uparrow$\\

\midrule
Dataset &\multicolumn{9}{c}{Blender 1 (small motion)} \\ \midrule 
Colmap   \cite{schonberger2016structure} & \multicolumn{4}{c}{\xmark} &   \multicolumn{4}{c}{\xmark} \\

DfUSMC\cite{ha2016high}	& \underline{0.0117}	& \underline{0.0108} &	\underline{0.0094}	& \underline{0.1948} &\underline{0.2107}	&\underline{0.4864}&	\underline{0.9683}	&\underline{0.7723}	& \underline{0.8877}	&{0.9409}\\

Saop \cite{chugunov2022shakes} &0.0274&	0.0229&	0.0204	&0.6369 
&0.5818	&1.8768	&1.7900	&0.3958&	0.6009	&0.7198\\

RCVD \cite{kopf2021robust} & 0.0168	&0.0162	&0.0140	& 0.2158 
& 0.3111	&0.5382	&1.2368	&0.5294&	0.814&	\underline{0.9524}\\

Ours &\textbf{ 0.0066}  &\textbf{ 0.0056 } &\textbf{0.0050}  &\textbf{0.1806}  &\textbf{0.1381} & \textbf{0.2391}  &\textbf{0.8688} &\textbf{ 0.8358 } &\textbf{0.9263} & \textbf{0.9761}  \\

\midrule
Dataset &\multicolumn{9}{c}{Blender 2 (micro motion)} \\ \midrule 
Colmap   \cite{schonberger2016structure} & \multicolumn{4}{c}{\xmark} &   \multicolumn{4}{c}{\xmark} \\

DfUSMC\cite{ha2016high}	 & \underline{0.0046}	&\underline{0.0026}	&\underline{0.0024}	&\underline{0.1918} 
&0.3093	&0.9543&	2.0499	&{0.5722}	&0.7785&	0.9187\\

Saop \cite{chugunov2022shakes} & 0.0078	&0.0043	&0.0040	&0.2678 
&0.2936	&0.8326	&2.0020	&{0.5794}&	0.7976	&0.9263\\

RCVD \cite{kopf2021robust} &  0.0168	&0.0162	&0.0140	& 0.2158 
&\underline{ 0.1898}	&\underline{0.3492}	&\underline{1.3745}	&\underline{0.6726}	&\underline{0.8816}	&\underline{0.9693}\\

Ours &  \textbf{0.0022} & \textbf{0.0022} & \textbf{0.0020 } &\textbf{ 0.0245}  & \textbf{0.1383}  & \textbf{0.1962 }& \textbf{1.1521 } & \textbf{0.7996 }& \textbf{0.9819 }& \textbf{0.9983}  \\

\bottomrule
\end{tabular}
\caption{Pose and depth errors metrics on the two proposed synthetic bursts datasets.}
\label{tab:pose_err}
\end{table*}

\newcommand\wwww{0.19}
\begin{figure*}[htbp]
    \setlength\tabcolsep{0.5pt}
    \renewcommand{\arraystretch}{0.5}
    \centering
        \begin{tabular}{ccccc}
             
\multirow{2}{*}[4em]{\includegraphics[trim=0 0 0 0,clip,width=0.22\textwidth]{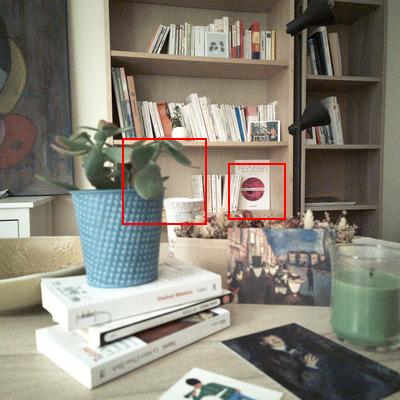}} &
            \includegraphics[trim=0 0 0 0,clip,width=\wwww\textwidth]{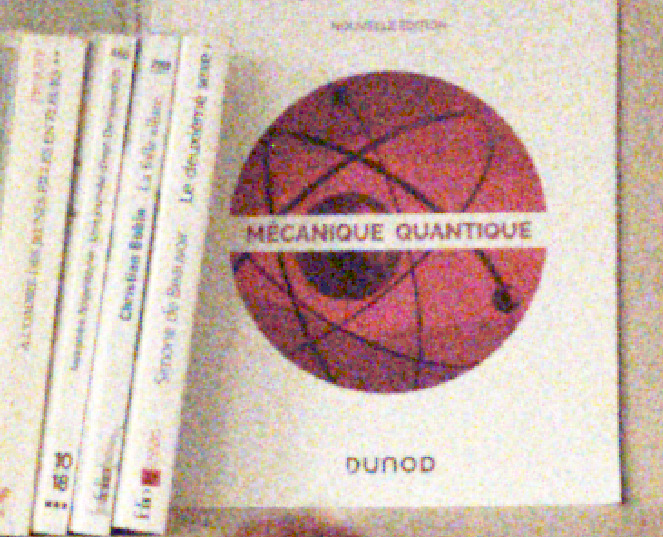} &
            \includegraphics[trim=0 0 0 0,clip,width=\wwww\textwidth]{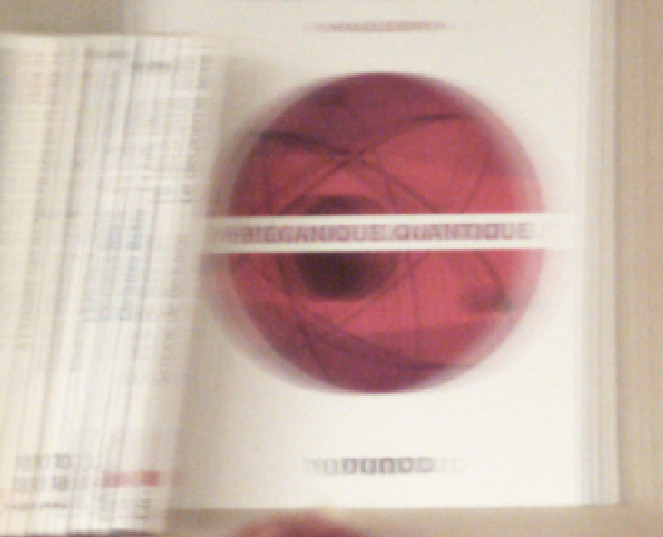} &
            \includegraphics[trim=0 0 0 0,clip,width=\wwww\textwidth]{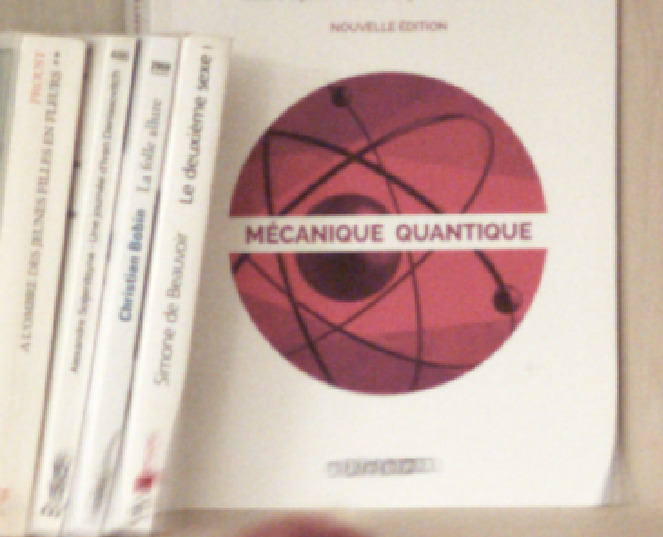} &
            \includegraphics[trim=0 0 0 0,clip,width=\wwww\textwidth]{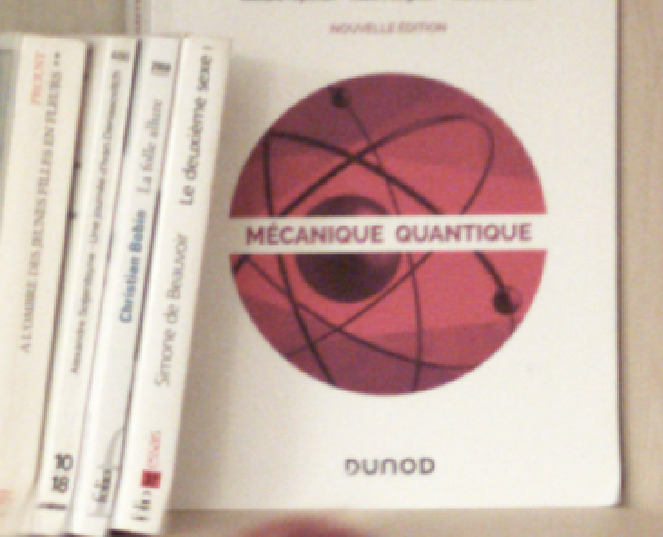} \\
            
&
            \includegraphics[trim=0 0 0 0,clip,width=\wwww\textwidth]{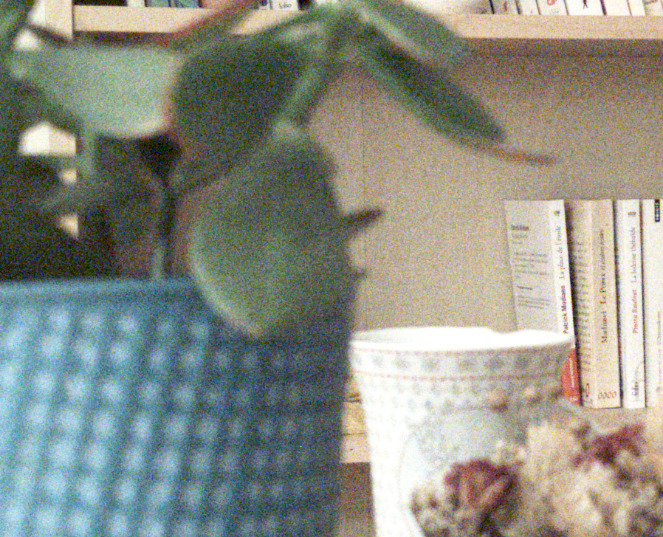} &
            \includegraphics[trim=0 0 0 0,clip,width=\wwww\textwidth]{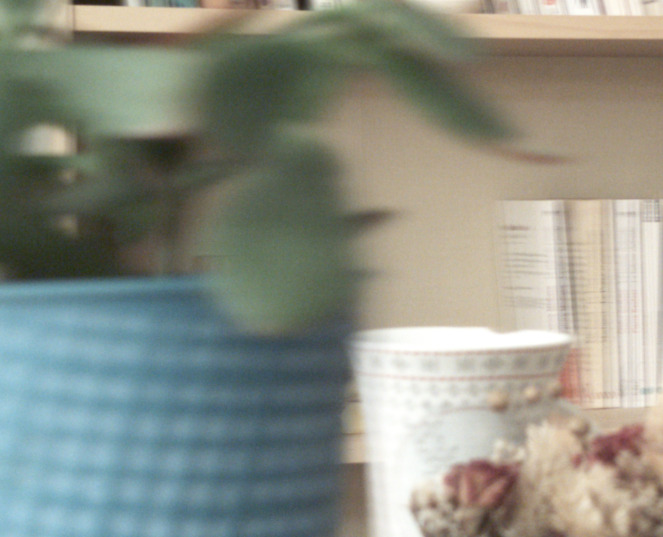} &
            \includegraphics[trim=0 0 0 0,clip,width=\wwww\textwidth]{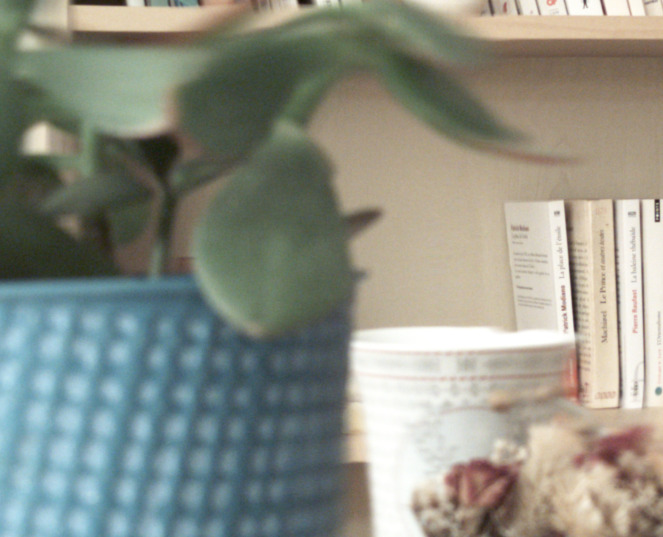} &
            \includegraphics[trim=0 0 0 0,clip,width=\wwww\textwidth]{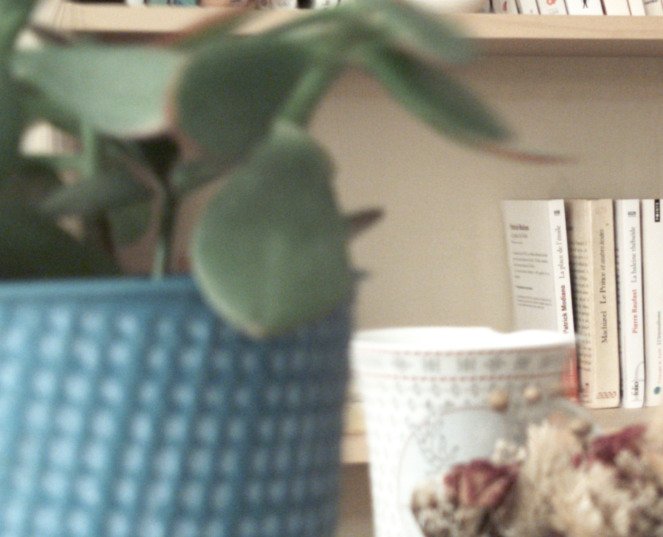} 
            \\
            Ref. image & Noisy & Homography & Farneback~\cite{farneback2003two} & Ours \\ 
            
\end{tabular}
    \caption{Burst denoising for night photography on real bursts exploiting alignment of various algorithms. Left: Full image with bounding boxes highlighting the region of interest. Top line: background region is misaligned for concurrent methods. Bottom line: The cup is misaligned for other methods. Homography misaligned the plant as well. It is best seen by zooming on a computer screen.}
    \label{fig:denoising}
\end{figure*}

\newcommand\xs{0.12}
 \begin{figure}[htbp]
         \setlength\tabcolsep{0.5pt}
     \renewcommand{\arraystretch}{0.5}
     \centering
         \begin{tabular}{cccc}
          
         \includegraphics[width=\xs\textwidth]{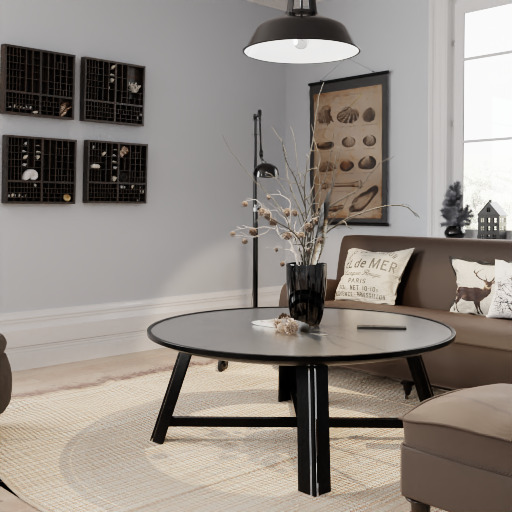}&
         \includegraphics[width=\xs\textwidth]{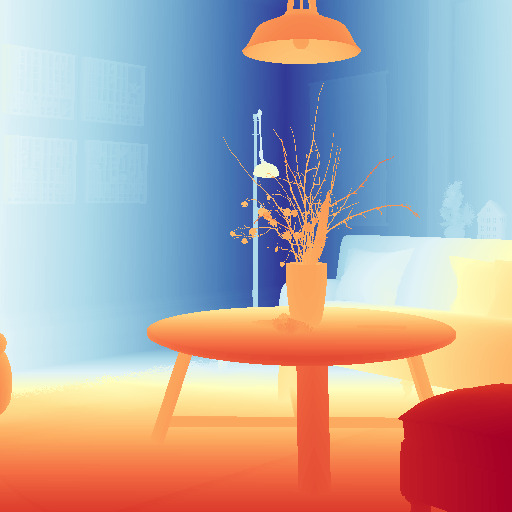}&
             \includegraphics[width=\xs\textwidth]{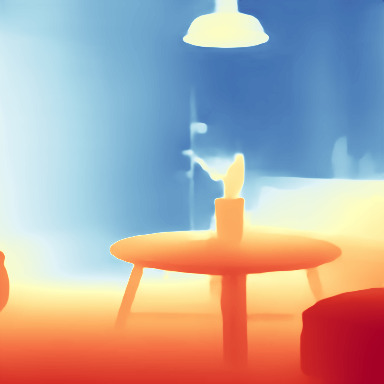}&
             \includegraphics[width=\xs\textwidth]{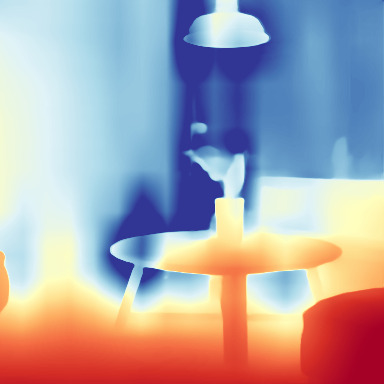}\\
             Ref. image & Groundtruth & Midas \cite{ranftl2020towards} & RCVD~\cite{kopf2021robust} \\
             \includegraphics[width=\xs\textwidth]{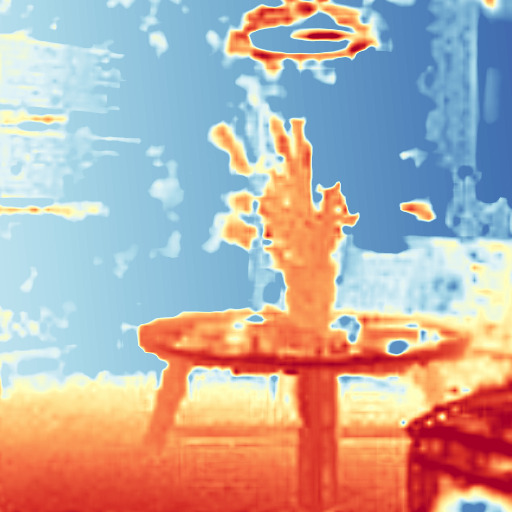}&
             \includegraphics[width=\xs\textwidth]{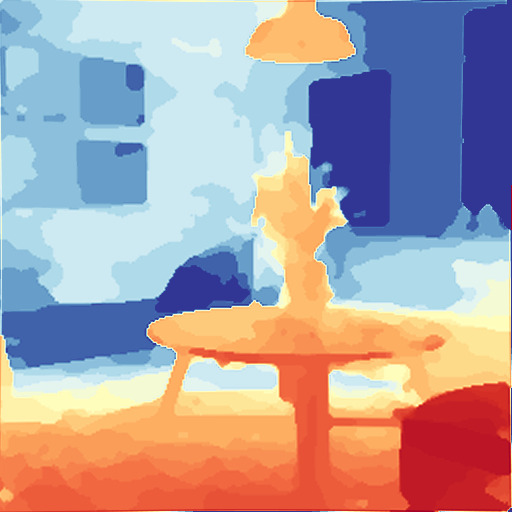}&
                  \includegraphics[width=\xs\textwidth]{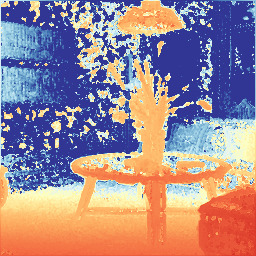} &
     \includegraphics[width=\xs\textwidth]{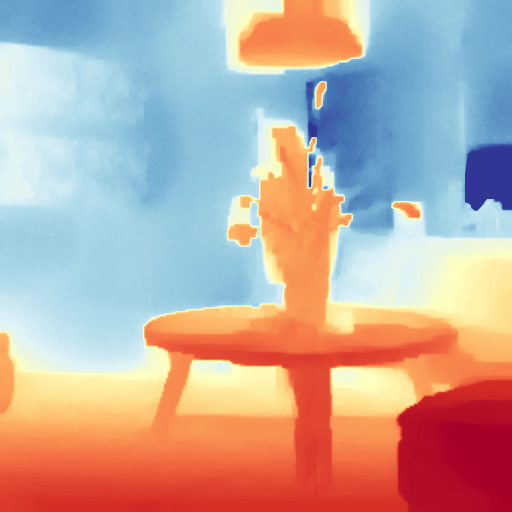} \\
             Saop \cite{chugunov2022shakes} & DfUSMC \cite{ha2016high} & Ours & Ours + reg\\ 
 \end{tabular}
     \caption{Depth estimation from a synthetic image burst. It is one of the scenes generated with Blender used in the dataset \emph{Blender 2}.  We present our result w/o regularisation (\emph{Ours}) and with determinant penalization (\emph{Ours + reg}) for smoother results, see the Appendix~\ref{app:det}.}
     \label{fig:depthmap3}
     \vspace{-0.4cm}
 \end{figure}

We conduct experiments on synthetic bursts and showcase practical applications using real bursts captured with a Pixel 6 Pro smartphone. These applications include night photography and 3D reconstruction, serving as proof of concept. Additionally, we have included preliminary experiments on burst super-resolution in Appendix~\ref{app:sr}.

\myparagraph{Synthetic burst simulation}
We require photorealistic bursts containing ground truth depth and camera poses for evaluating our approach and concurrent methods, but existing public multi-view stereo datasets we are aware of lack the needed characteristics due to non-static scenes or excessively large frame baselines that do not align with our specific use cases.

We generated two photorealistic synthetic datasets using CYCLES, the path tracing engine of Blender \cite{blender}. 
We used a set of twelve publicly available indoor scenes made by 3D artists, with detailed and varied scene compositions. 

Ten scenes come from~\cite{archinteriors}, and two scenes are from~\cite{archtopics}.
Each burst of the dataset consists of 20 frames, with a resolution of 512x512 pixels.

We skipped the post-processing denoising step at the end of the rendering to avoid temporal flickering artifacts and mitigated render noise by using a large number of samples (4096).
The camera trajectories and orientations are crafted as follows: a few keyframes was positioned manually to outline the global path, and the other keyframes were obtained with Bezier interpolation. 

We generated two datasets: Blender 1 with small baselines and Blender 2 with micro-baselines. The first dataset exhibits larger parallax effects, while the second dataset has reduced parallax effects. Detailed characteristics of these datasets are provided in the Appendix~\ref{app:detail}.

\myparagraph{Evaluation on synthetic data}
We initialize our algorithm on synthetic data with a $16\times16$ coarse depth map using the shallow network from~\cite{ranftl2020towards}.
For the evaluation, we follow the standard practice to evaluate pose, depth, and flow, described in \cite{kopf2021robust,geiger2012we}. 
For all the methods, as depth estimation and pose are known up to an unknown scale, we align the predicted depth and the ground truths using median scaling. For pose evaluation, we compute the scale factor as $s = \argmin_s \| T - s \hat{T}\|^2$, where $T=\begin{bmatrix}\tb_0, \cdots , \tb_N \end{bmatrix}$. In addition, we use the canonic left-invariant distance in $SE(3)$ that combines rotational and translation parts in one quantity; see \cite{Chirikjian2009, left_spatial} for details. We report the distance between the ground truth pose and the estimated pose. It reads $d([R,t], [R',t'])^2 = \|t'-t\|_2^2 + \lambda\|\log(R^\top R')\|_2^2$. For $\lambda$, we use the median value of the ground truth depth. $\|\log(R^\top R')\|_2$ is the canonic metric on the set of rotation $SO(3)$ and is also reported independently.
Unlike other methods in the literature \cite{kopf2021robust}, we choose not to present relative pose error (RPE) as a good RPE may not correlate with good alignment metrics and rely on a time coherent burst. To evaluate the ATE, we did not align the estimated poses with the ground truth poses with rigid transformation, as is common in the SLAM community. Indeed, our loss \ref{eq:min_prob} and, more generally, the flow is not invariant by a solid transformation of the poses. As the final goal of our method is alignment, performance evaluation up to a rigid transformation would not be informative.

For optical flow evaluation, we conducted comparisons on our synthetic datasets. We utilized a state-of-the-art deep optical flow method \cite{teed2020raft} by registering all frames pairwise with a reference. Additionally, we employed a standard homography and the Farneback optical flow \cite{farneback2003two} for comparison. Furthermore, we computed optical flow errors for other concurrent methods \cite{chugunov2022shakes,ha2016high,kopf2021robust} using the camera projection model as in Eq.~\eqref{eq:depthmap} and their estimated pose and depth maps.
Leveraging the assumption of a static scene, our method consistently outperformed \cite{teed2020raft} regarding flow accuracy.

We conducted comparisons of our pose and depth estimation method with methods introduced in \cite{kopf2021robust}, \cite{chugunov2022shakes}, and \cite{ha2016high}, utilizing publicly available codebases. To ensure a fair comparison, we initialized the method from \cite{chugunov2022shakes} with the same depth map as the one we used for our own initialization.

We compare our method with a monocular depth estimation model Midas \cite{ranftl2020towards}. However, monocular methods estimate depth up to an affine transformation, whereas flow estimation is not invariant by affine reparameterization. Using affine registration lacks full relevance to evaluate the quality of the result, so the performances in Table \ref{tab:pose_err} are obtained after rescaling only. For a fair depth map comparison, we also evaluate our method and others against Midas with an affine registration. Results are presented in Appendix~\ref{app:comp_mono}.

\myparagraph{3D reconstructions quality on synthetic data and real bursts}

We evaluate qualitatively our depth reconstructions on synthetic data from our dataset and real bursts captured with a Pixel 6 Pro smartphone.

Visualizations of reconstructed depth maps are provided in Figure~\ref{fig:depthmap3}. Our depth map can have a noisy aspect on a texture-less structure. This is a normal feature as our optimization is not well conditioned on uniform surfaces, as small variations in inferred depth will not affect the reprojection photometric loss. This noisy effect can be mitigated by adding spatial regularization for the scene steps. But this trades with lower performance in terms of flow and pose metrics on synthetic data. We observed that no spatial regularization plan parameters give the best image alignment and pose estimation results. We detail the spatial regularization in Appendix \ref{app:det}.

\begin{figure}
    \centering
    \includegraphics[width=0.5\textwidth]{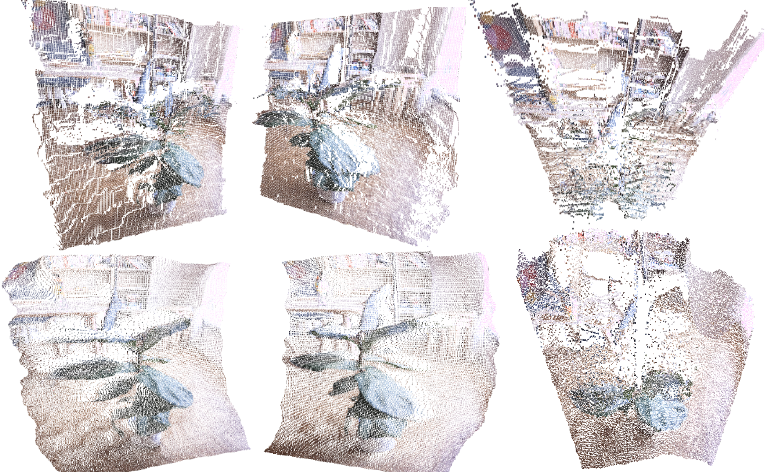}
    \caption{\textbf{Top}~: point cloud reconstruction with DfUSMC~\cite{ha2016high}. \textbf{Bottom}~: point cloud reconstructed with our method. We show respectively left, right, and top views of the two point clouds.}
    \label{fig:pcd} 
       \vspace{-0.3cm}
\end{figure}

For real scenes, we showcase the high-quality depth reconstructions achievable with our method in Figure~\ref{fig:depthmap}. We input RAW image bursts from the Pixel 6 Pro smartphone, and perform demosaicking using bilinear filtering. We initialize our algorithm with a low-resolution depth map from the phone sensor. We compare our results with depth maps obtained from a monocular method \cite{ranftl2020towards}, RCVD \cite{kopf2021robust}, Saop \cite{chugunov2022shakes}, and DfUSMC \cite{ha2016high}. Furthermore, we provide visualizations of reconstructed point clouds in Figure~\ref{fig:pcd}.

\myparagraph{Low-light photography on real bursts}

To demonstrate the robustness and accuracy of our alignment method for downstream tasks, we conducted a low-light photography experiment as a proof of concept. This scenario is challenging as it involves aligning frames with a low signal-to-noise ratio. We captured night bursts using a Pixel 6 Pro smartphone under low light conditions, using a short exposure time and high ISO settings to reduce motion blur. We aligned these frames using our method and other concurrent alignment algorithms, including a simple homography and dense optical flow using the Farneback implementation from OpenCV~\cite{bradski2000opencv}.

To reduce noise, we averaged the aligned frames, using a straightforward denoising approach. While our focus was on highlighting the registration quality of our method, it's worth noting that a more sophisticated fusion algorithm could be employed to enhance image quality and reduce artifacts, as seen in previous works \cite{liba2019handheld,hasinoff2016burst}.

In Figure \ref{fig:denoising}, we provide visual comparisons of our results. We observed that due to the nonplanar nature of the scene, the homography-based approach failed to align objects in the foreground and background, resulting in a blurry appearance in the denoised image. In contrast, the optical flow model exhibited greater flexibility, successfully aligning objects in both the foreground and background. However, some elements, such as the white book in the background or certain patterns on the white cup in the foreground, were still not perfectly aligned.

\newcommand\x{0.15}
\begin{figure}[htbp]
        \setlength\tabcolsep{0.5pt}
    \renewcommand{\arraystretch}{0.5}
    \centering
        \begin{tabular}{ccc}
       
        \includegraphics[width=\x\textwidth]{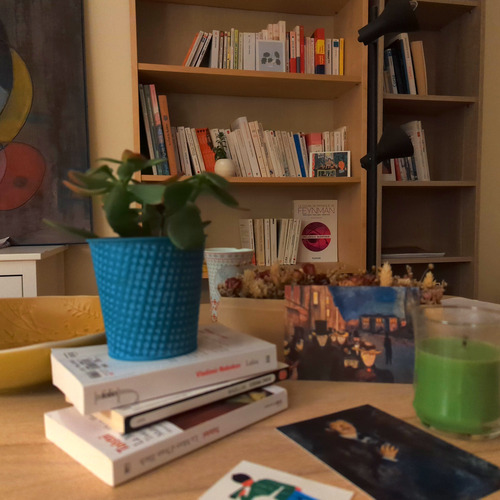}&
            \includegraphics[width=\x\textwidth]{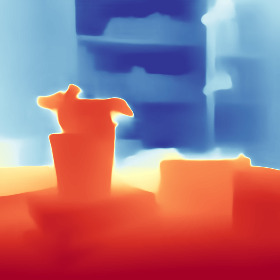}&
            \includegraphics[width=\x\textwidth]{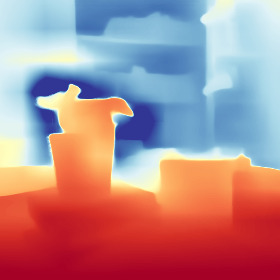}\\
                        Ref. image & Midas~\cite{ranftl2020towards} & RCVD~\cite{kopf2021robust} \\
            \includegraphics[width=\x\textwidth]{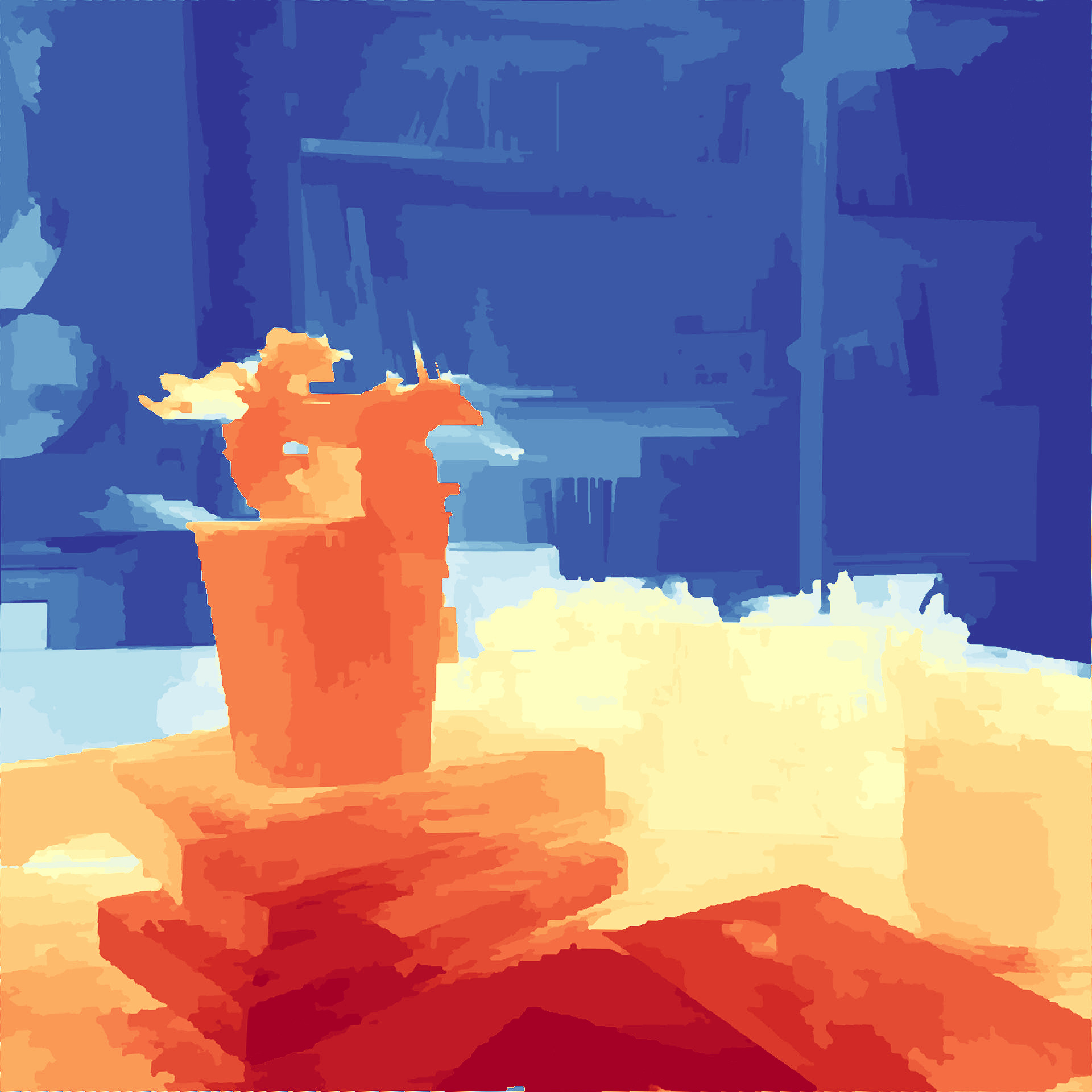} &
            \includegraphics[width=\x\textwidth]{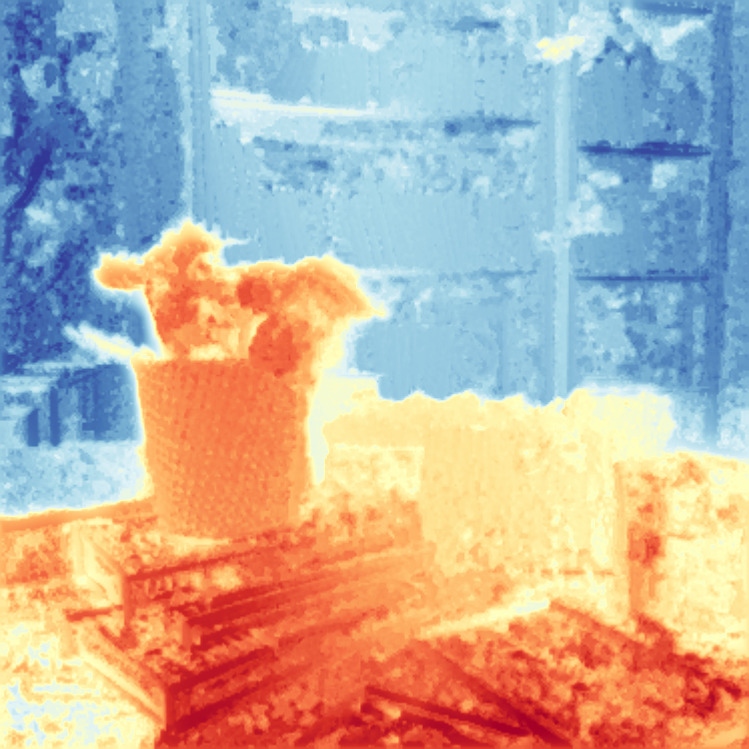} &
                        \includegraphics[width=\x\textwidth]{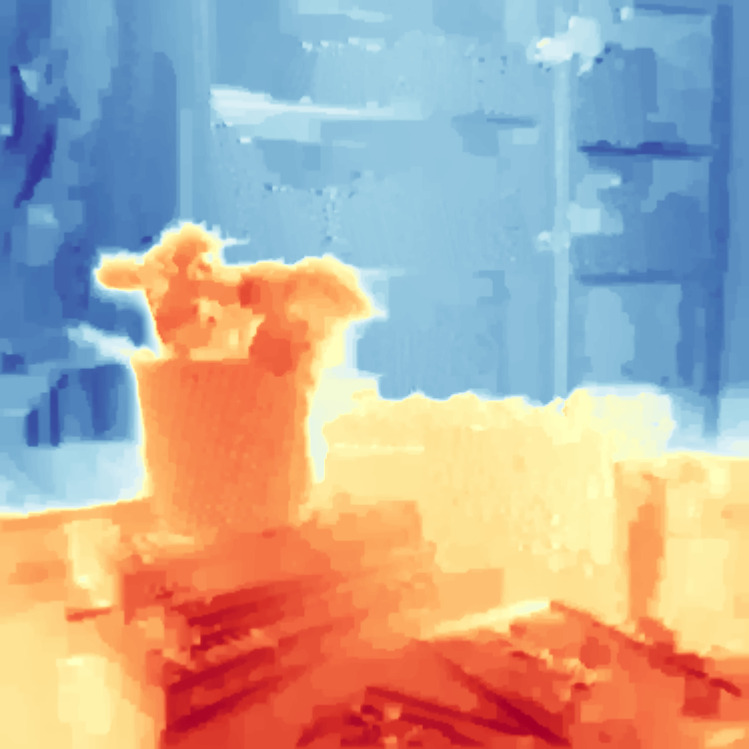} 
            \\
            DfUSMC~\cite{ha2016high} & Ours & Ours+reg\\ 
\end{tabular}
    \caption{Depth estimation from a real burst.  We present our result w/o regularisation (\emph{Ours}) and with determinant penalization (\emph{Ours + reg}) for smoother results; see Appendix~\ref{app:det} for more details.}
    \label{fig:depthmap}
\end{figure}

\begin{figure}[htbp] 
   \centering
   \includegraphics[trim=0 10 0 10,clip,width=0.5\textwidth]{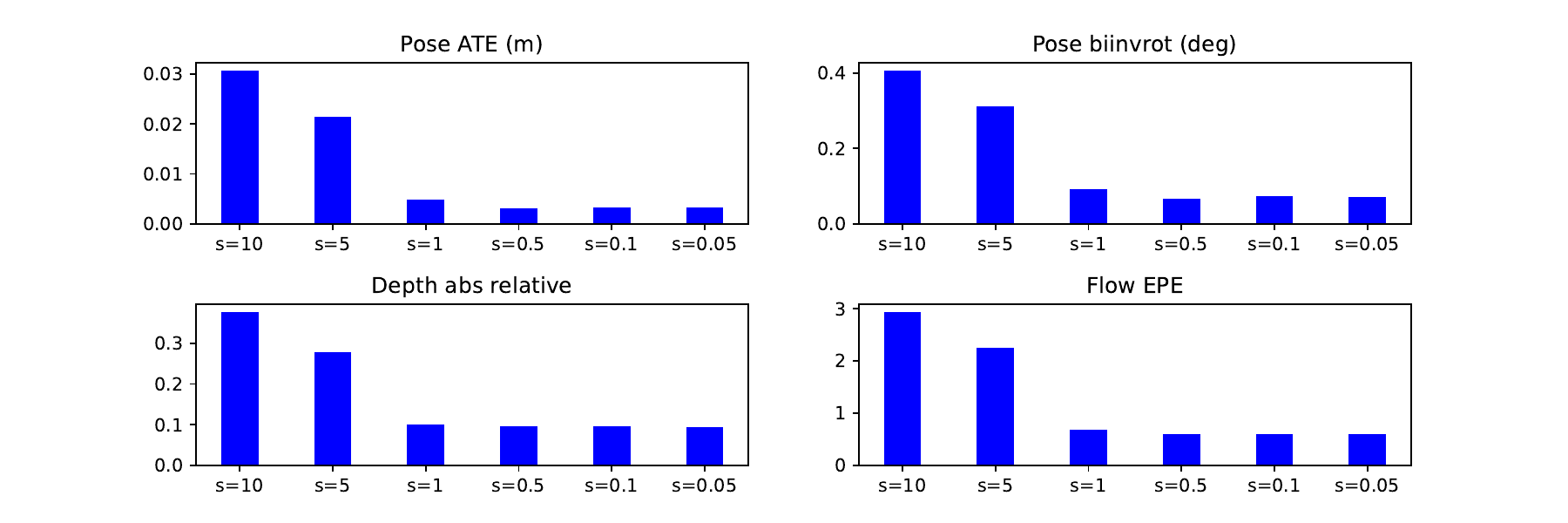}
  \caption{Noise on the initialization depth map. Our method is robust to noise; the performance degrades when the noise's variance is larger than 1 meter.}
   \label{fig:snr}
   \vspace{-0.3cm}
\end{figure}

\myparagraph{Depth initialization}
Figure \ref{fig:snr} shows the impact of the depth map's initialization on our method's performance. We gradually increase the variance of a Gaussian random noise added to the $16\times16$ initialization depth map and evaluate the performance of our algorithm on our synthetic dataset with various depth, pose, and alignment metrics. This experiment demonstrates that our method is robust to noise on the initialization depth map. The model only requires a noisy estimate to converge to the right solution.

\section{Conclusion }

Our approach offers a comprehensive and versatile solution for burst photography. It excels in accurately estimating flow, depth, and pose, setting a new benchmark for processing small motion scenes. 
Future enhancements include integrating intrinsic camera parameter estimation like focal length to improve accuracy, refining our model for lens-induced distortions, and exploring more advanced camera models such as thin-lens to account for defocus effects. 

\subsection*{Acknowledgments}

This work was funded in part by the French government under management of Agence Nationale
de la Recherche as part of the ``Investissements d'avenir'' program, reference ANR-19-P3IA-0001
(PRAIRIE 3IA Institute). JM was supported by the ERC grant number 101087696
(APHELAIA project) and by ANR 3IA MIAI@Grenoble Alpes (ANR-19-P3IA-0003). JP was partly supported by the Louis Vuitton/ENS chair in artificial intelligence and a Global Distinguished Professor appointment at the Courant Institute of Mathematical Sciences and the Center for Data Science of New York University.

{
    \small
    \bibliographystyle{ieeenat_fullname}
    \bibliography{references}
}

\clearpage
\appendix
\onecolumn
\section*{Appendix}
\addcontentsline{toc}{section}{Appendices}
\renewcommand{\thesubsection}{\Alph{subsection}}
\subsection{Closed form Jacobean for Gauss-Newton 
step}\label{app:cf_jac}
From Eq. \eqref{eq:min_prob}, we recall that the residual of the robust least square for which we have to compute the Jacobian is the flat vector $\rb$ with coordinates indexed by $k=1..K$, $i\in \Gb$, $\ub'\in P_i$ with total dimension noted $K D$ with $D=\#(\Gb) \#(P)$ size of grid by size of patch:
\begin{equation}
    r_{k,i,\ub'} = I_0(\ub') - I_k(\hat{H}_{i,k}(\ub')).
\end{equation}
We note the twist $\xi_k$ such that $\Rb_k, \tb_k = \operatorname{Exp}(\xi_k)$. We want to find the Jacobian $\operatorname{J}_\rb$ with the variable flat vector $\mathbf{\xi}=(\xi_k)_{k=1..K}$. Then we note that $\operatorname{J}_\rb$ is of dimensions $KD,K$ and that $\operatorname{J}_\rb$ is diagonal by $K$ block of dimension $D,K$. We note $\operatorname{J}_k$ these blocks.
Using the expression of the homography matrix in \eqref{eq:homo},  the block $\operatorname{J}_{k}$ have $D$ rows of the form $(\nabla \phi_{k,i,\ub'})^\top$ where:
\begin{align}
        \phi_{k,i,\ub'}(\xi) &= I_k(\psi((\Rb + \tb \nb_i^\top)\ub'))\\
    &= I_k\left(\psi\left(\Rb \frac{1}{\nb_i^\top\ub'}\ub' + \tb\right)\right),
\end{align}
with $\psi(x,y,z)=[x/z, y/z]^\top$ and $\Rb, \tb = \operatorname{Exp}(\xi)$. So if we note $\Xb_{i,\ub'} = \frac{1}{\nb_i^\top\ub'}\ub'$ in $\mathbb{R}^3$ and $\Lambda_{\Xb}:SE(3)\rightarrow \mathbb{R}^3$ the action on $\Xb\in\mathbb{R}^3$ that takes an element of $[\Rb, \tb]$ in $SE(3)$ and gives its action on $\Xb$: $\Lambda_{\Xb}(\Rb, \tb) = \Rb\Xb + \tb$ we can simplify $\phi_{k,i,\ub'}$ to a simple composition and compute its gradient with a chain rule:
\begin{align}
    \phi_{k,i,\ub'} &= I_k\circ\psi\circ\Lambda_{\Xb_{i,\ub'}}\circ\operatorname{Exp}\\
    (\nabla \phi_{k,i,\ub'})^\top &= \nabla I_k^\top \operatorname{J}_{\psi}\operatorname{J}_{\Lambda_{\Xb_{i,\ub'}}}\operatorname{J}_{\operatorname{Exp}}.
\end{align}
Note that $\Lambda_{\Xb}$ takes input on the group $SE(3)$, and $\operatorname{Exp}$ has an output on the same group. However, as described in \cite{microlie}, using the so-called \emph{left jacobian} suffices. $\nabla I_k$ is the spatial gradient of the image $I_k$ calculated using a convolution and a Sobel kernel and evaluated in a coordinate using bilinear interpolation. The individual Jacobians are reported dropping indexes in Table \ref{tab:cf}.

In practice, exploit the diagonal structure of $\operatorname{J}_{\rb}$ in our implementation.
\begin{table}[htp]
    \centering
    
    \begin{tabular}{lcc}
    \toprule
Domains & Function & Jacobian\\ \midrule
$\mathbb{R}^3\rightarrow \mathbb{R}^2$&
$\psi(x,y,z)=[x/z, y/z]^\top$&
$\operatorname{J}_{\psi} = \frac{1}{z}[\operatorname{I}_2|-\psi(x,y,z)]$\\
$SE(3)\rightarrow \mathbb{R}^3$&
$\Lambda_{\Xb}(\Rb,\tb)=\Rb\Xb + \tb$&
$\operatorname{J}_{\Lambda_{\Xb}} = [\Rb|-\Rb[\Xb]_{\times}]$\\

$\mathbb{R}^6\rightarrow SE(3)$&
$\operatorname{Exp}(\xi)$ as eq (172) in \cite{microlie} &
$\operatorname{J}_{Exp}$ as eq (179a) in \cite{microlie}\\
\bottomrule
\end{tabular}
\caption{Closed form of functions needed to calculate the residual jacobian. $\operatorname{I}_2$ is the identity matrix of size $2$}
\label{tab:cf}
\end{table}

\subsection{Fixed point algorithm for reverse flow estimation}\label{app:fp}
We have a depth map in the reference view $ (z_i^{(0)})_{i\in\Gb}$ and we note $\gamma_i^{(0)}= 1/z_i^{(0)}$ the associated disparity.

Given a disparity $\gamma$, a relative pose $\Rb,\tb$ and $\ub$ a point in the camera plane of the first view, we can calculate the image $\Bar{u}$ a point on the second camera plane and $\Bar{\gamma}$ the projected disparity in the second view frame:
\begin{align}    
    \Bar{u}\left(\ub,\gamma, \Rb, \tb\right) &= \psi\left(\Rb [\ub,1] + \gamma \tb\right)\label{eq:flow_pixel}\\
    \Bar{\gamma}\left(\ub,\gamma, \Rb, \tb\right) &= \gamma \omega \left(\Rb [\ub,1] + \gamma \tb\right).\label{eq:disp_reproj}
\end{align}

In particular, given a regular grid of $\ub_i^{(0)}$ in the reference view and the relative position of other views, $\Rb_k,\tb_k$, we have the direct flow:
\begin{align}    
    \Bar{\ub}_i^{(k)} = \Bar{u}\left(\ub_i^{(0)},\gamma_i^{(0)}, \Rb_k, \tb_k\right),
\end{align}
$\Bar{\ub}_i^{(k)}$ is not a regular grid in the view $C_k$, it is the image of a regular grid in the view $C_0$. The direct flow warp $I_k$ as an image $I_k^(0)$ in the camera plane $C_0$. It is called a backward warp. But for some applications, we also need the warp of the image $I_0$ as an image $I_0^{(k)}$ in the view $C_k$. This can be done using the direct flow $\Bar{\ub}_i^{(k)}$ and a forward warp, but it is known as not numerically stable. Instead, it can be computed using a backward warp and the \textit{reverse flow}.
The reverse flow is the other way around; the regular grid is $\ub_i^{(k)}$ in the camera plane $C_k$ and we want to find its antecedent $\Bar{\ub}_i^{(0)}$ in the camera plane of $C_0$. The reverse can be computed using \eqref{eq:flow_pixel} using the inverse of the relative position and the disparity map in the view $C_k$. The inverse of the relative position is the inverse in $SE(3)$, and it is $\Rb_k^\top, -\Rb_k^\top\tb_k$. On the other hand, the disparity map in the view $C_k$ is not known. However, using the inverse relative position, the disparity $\gamma^{(k)}$ in a point $\ub^{(k)}$ in the camera plane of $C_k$ must match the known one in $C_0$:
\begin{align}
    \Bar{\gamma}(\ub^{(k)}, \gamma^{(k)},\Rb^\top_k,-\Rb^\top_k\tb_k) = \Gamma^{(0)}\left(\Bar{\ub}(\ub^{(k)}, \gamma^{(k)},\Rb^\top_k,-\Rb^\top_k\tb_k)\right),\label{eq:implicit_oter_disp}
\end{align}
where $\Gamma^{(0)}$ is the disparity function on the whole camera plane of $C_0$ using interpolation and the depth map $(z_i^{(0)})_{i\in\Gb}$. This equation can be interpreted as the reprojection of the disparity in $C_k$ must match the disparity in $C_0$ evaluated in the flow induced by the disparity in $C_k$. It is an implicit equation for $\gamma^{(k)}$. Using \eqref{eq:disp_reproj} again from $C_0$ to $C_k$, it can be converted as a fixed point equation $\gamma^{(k)} = F\left(\gamma^{(k)}\right)$ when defining $F$ as :
\begin{align}
    F\left(\gamma\right) = \Bar{\gamma}\left(\ub, \Gamma^{(0)}(\ub),\Rb_k, \tb_k\right)\quad
    \text{with}\quad\ub=\Bar{\ub}\left(\ub^{(k)}, \gamma,\Rb^\top_k,-\Rb^\top_k\tb_k\right).\nonumber
\end{align}
Then we can estimate the disparity map $(\gamma_i^{(k)})_{i\in \Gb}$ in view $C_k$ using a regular grid $\ub_i^{(k)}$ in the view $C_k$ and using a fixed point algorithm with a function $F$ for every pixel using $\ub_i^{(k)}$.

We build the sequence for $\gamma_{i, m}^{(k)}$ with $m>0$ as:
\begin{align}
    \gamma^{(k)}_{i, m+1} = \Bar{\gamma}\left(\ub_{i,k, m} , \Gamma^{(0)}(\ub_{i,k, m}), \Rb_k, \tb_k\right) \quad \text{with}\quad
    \ub_{i,k, m} = \Bar{u}\left(\ub_i^{(k)}, \gamma^{(k)}_{i,m}, \Rb_k^\top, - \Rb_k^\top\tb_k\right),
\end{align}
as the motion baseline is small, we initialize the disparity map in view $C_k$ by the one in $C_0$: $\gamma^{(k)}_{i, 0}=\gamma^{(0)}_{i}$ and we can use the composition of the two flows (direct and reverse) as a convergence error:
\begin{align}
\epsilon_{i,k,m} = \left\|\ub_i^{(k)} - \Bar{\ub}\left(\ub_{i,k,m},\Gamma^{(0)}(\ub_{i,k, m}), \Rb_k, \tb_k\right)\right\|
\end{align}
The value of $i$ and $k$ for which the sequence does not converge correspond to the occlusion of the element projected in $\ub_i^{(k)}$ between view $C_k$ and $C_0$. We can build an occlusion mask using the convergence criterion. Examples of these masks are available in appendix \ref{app:occ}. For the value of $i$ and $k$ for which the sequence does converge $\gamma^{(k)}_{i}$.

Finally, the reverse flow is given by:
\begin{align}
\Bar{\ub}_i^{(0)}=\Bar{u}\left(\ub_i^{(k)}, \gamma^{(k)}_{i}, \Rb_k^\top, - \Rb_k^\top\tb_k\right).
\end{align}

\subsection{Determinant regularization}\label{app:det}
\begin{figure}
    \centering
        \includegraphics[trim=0 0 0 0, clip, width=0.9\textwidth]{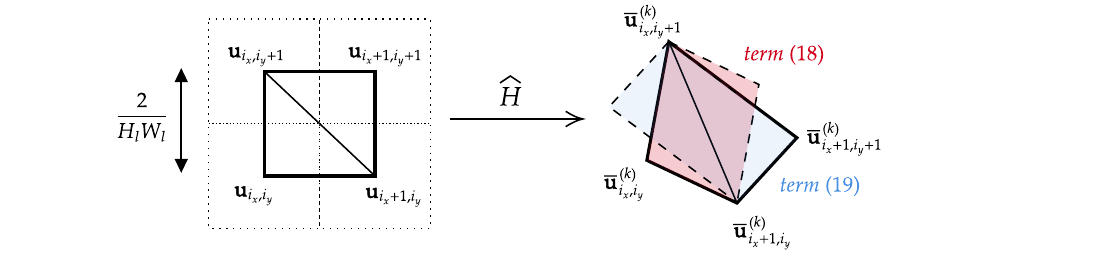}
    \caption{Illustration of the determinant regularization.}\label{fig:determinant}
    \vspace{-0.5cm}
\end{figure}

The idea behind this regularization is that when the gradient is small, we will favor the direction of descent for the structure that deforms the current flow the least. To do this, we look at the flow effect on the center of the patches regularly distributed on the $\Gb^{(l)}$ grid. We note $i=i_x,i_y$ the $i$ elements of the grid $\Gb^{(l)}$ with $i_x=1..W_l$ and $i_y=1..H_l$. We note $(\ub_{i_x,i_y})$ the center point of the pixel in the corresponding image plane, and we suppose that the coordinates of $\ub_{i_x,i_y}$ are normalized and evolve in a range $[-1,1]$. A parallelogram constituted by the points $(\ub_{i_x,i_y}, \ub_{i_x+1,i_y}, \ub_{i_x+1,i_y+1}, \ub_{i_x,i_y+1})$ thus has a normalized area of $4 / (H_lW_l)$. We compare independently, for each view $k$ and each grid mesh element, the normalized area of the mesh after application of the local homographic flow and the constant area noted $\Bar{\ub}_{i_x,i_y}^{(k)} = \hat{H}_{(i_x,i_y),k}(\ub_{i_x,i_y})$. We penalize the ratio of the area of each parallelogram before and after the homography flow to 1. The penalization reads:
\begin{align}
P =& \sum_{k=1}^K\sum_{i_x=1}^{W_l-1}\sum_{i_y=1}^{H_l-1}\left|\frac{\mathcal{A}_{i_x,i_y}^{(k)}/2}{4 / (H_lW_l)} - 1\right|\\
\mathcal{A}_{i_x,i_y}^{(k)} =& \operatorname{det}(\Bar{\ub}_{i_x+1,i_y}^{(k)} - \Bar{\ub}_{i_x,i_y}^{(k)},\Bar{\ub}_{i_x,i_y+1}^{(k)} - \Bar{\ub}_{i_x,i_y}^{(k)})\\&+ \operatorname{det}(\Bar{\ub}_{i_x,i_y+1}^{(k)} - \Bar{\ub}_{i_x+1,i_y+1}^{(k)}, \Bar{\ub}_{i_x+1,i_y}^{(k)} - \Bar{\ub}_{i_x+1,i_y+1}^{(k)}),
\end{align}
where $\mathcal{A}_{i_x,i_y}$ is the double of the area of the parallelogram $(\Bar{\ub}_{i_x,i_y}^{(k)}, \Bar{\ub}_{i_x+1,i_y}^{(k)}, \Bar{\ub}_{i_x+1,i_y+1}^{(k)}, \Bar{\ub}_{i_x,i_y+1}^{(k)})$ using determinant on the two halves triangle as illustrated in figure \ref{fig:determinant}.

\subsection{Additional details on the datasets and on the experiments}\label{app:detail}

\paragraph{Proposed datasets.} We generated two datasets: Blender 1 with small baselines and Blender 2 with micro-baselines.
The first dataset exhibits larger parallax effects, while the second dataset has reduced parallax effects.
Detailed characteristics of these datasets are provided in Table \ref{app:tab_dataset}.

\begin{table}[b]
    \centering
    \footnotesize
    \begin{tabular}{lccccccc}
    \toprule
Dataset & Scenes & Frames &  \begin{tabular}[c]{@{}c@{}}Std baselines \\ (m) \end{tabular} & \begin{tabular}[c]{@{}c@{}}Std rotations \\ (deg) \end{tabular}  & \begin{tabular}[c]{@{}c@{}}Max depth \\ (m) \end{tabular}  &\begin{tabular}[c]{@{}c@{}}Min depth \\ (m) \end{tabular}  &\begin{tabular}[c]{@{}c@{}}Mean depth \\ (m) \end{tabular} \\ \midrule
Blender 1 & 15 & 20 & 0.116 & 0.20 & 0.316 & 11.234 & 3.73\\
Blender 2 & 10 & 20 & 0.010 & 0.29 & 1.92 & 19.453 & 6.21\\
\bottomrule
\end{tabular}
\caption{Main characteristics of the two proposed datasets.}
\label{app:tab_dataset}
\end{table}

\paragraph{Experiments.} In our experiments, whose results are reported in Table \ref{tab:flow_err} and Table \ref{tab:pose_err}, we evaluated the performance of the Saop method \cite{chugunov2022shakes} by calculating the average results across all scenes where Saop successfully converged.  On the \textit{blender 1} dataset, we excluded one scene where Saop did not converge. Excluding this scene for Saop does not change the methods' ranking and our experiments' conclusion.
\begin{table}[htp]
    \centering
    \footnotesize
    \begin{tabular}{lcccccccc}
    \toprule
Method   &Abs rel $\downarrow$  &Sqr rel $\downarrow$&  RMSE$\downarrow$    &Delta 1$\uparrow$  &Delta 2    $\uparrow$&Delta 3  $\uparrow$\\ \midrule
 \multicolumn{7}{c}{Blender 1 (small motion)} \\ \midrule
 Midas \cite{ranftl2020towards} &\underline{0.1589}&    1.0747& \underline{1.3148}& \textbf{0.8019}&    \underline{0.951}&  \underline{0.9824} \\
RCVD \cite{kopf2021robust}& 0.2038& \underline{1.0622}& 1.3888& 0.698&  0.9191& 0.9684\\
Ours &\textbf{ 0.1544}& \textbf{0.2229}&\textbf{ 0.9258}&   \underline{0.7881}&\textbf{ 0.9544}&    \textbf{0.9911}     \\
\midrule
 \multicolumn{7}{c}{Blender 2 (micro motion)} \\ \midrule
Midas \cite{ranftl2020towards} & \textbf{0.0790}    &\textbf{0.0786} &  \textbf{0.7166}&    \textbf{0.9429}&    \textbf{0.9929} & \textbf{0.9986}\\
RCVD \cite{kopf2021robust}& \underline{0.0971}  &\underline{0.1131}&    \underline{0.8244}& \underline{0.9149}& \underline{0.988} & 0.9973\\
Ours  &0.1763   &0.2875&    1.3711& 0.6857  &0.9594 &\underline{0.9976}\\
\bottomrule
\end{tabular}
\caption{Depth errors metrics on the two proposed synthetic bursts datasets.}
\label{tab:depth_err_aff}
\end{table}

\subsection{Ablation study}\label{app:abl_stdy}
We make an ablation study to understand the impact of the different choices in our modeling and algorithm. We compare the global algorithm to an identical algorithm using the same hyperparameters but, respectively, without the exponential parametrization of the motion, without the newton step, using spatial regularization (total variation and determinant), without the plan parametrization, with patches of size one, i.e., a pixel-wise loss and without the multiscale approach.
We report the performance on the fllow estimate in Table. \ref{tab:abl_flow_err}, and depth/pose in Table. \ref{tab:abl_pose_err}.

\begin{table*}[htbp]
    \centering
    \footnotesize
    \begin{tabular}{lccccc|ccccc}
    \toprule
Method   
& \begin{tabular}[c]{@{}c@{}}EPE  \\ $\downarrow$ \end{tabular} &\begin{tabular}[c]{@{}c@{}}RMSE  \\ $\downarrow$ \end{tabular} & 	\begin{tabular}[c]{@{}c@{}} NPE1  \\ $\uparrow$ \end{tabular} &\begin{tabular}[c]{@{}c@{}} NPE2  \\ 
$\uparrow$ \end{tabular} &	\begin{tabular}[c]{@{}c@{}} NPE3  \\ $\uparrow$ \end{tabular}

& \begin{tabular}[c]{@{}c@{}}EPE  \\ $\downarrow$ \end{tabular} &\begin{tabular}[c]{@{}c@{}}RMSE  \\ $\downarrow$ \end{tabular} & 	\begin{tabular}[c]{@{}c@{}} NPE1  \\ $\uparrow$ \end{tabular} &\begin{tabular}[c]{@{}c@{}} NPE2  \\ 
$\uparrow$ \end{tabular} &	\begin{tabular}[c]{@{}c@{}} NPE3  \\ $\uparrow$ \end{tabular}
\\ \midrule

 & \multicolumn{5}{c}{Blender 1 (small motion)} & \multicolumn{5}{c}{Blender 2 (micro motion)}\\  \midrule

Base   &\textbf{0.7439 }& \textbf{1.4324}  &\textbf{0.7841} &\textbf{ 0.9084 } &\textbf{0.9456} & \textbf{0.2321} & \textbf{0.2820} &\textbf{ 0.9366} & \textbf{0.9972} & \textbf{1.0000}  \\

with regularization & 0.7641 & 1.4596 & 0.7732 & 0.9024 & 0.9432 & 0.2660 & 0.3286 & 0.9297 & 0.9937 & 0.9997 \\

with $k=1$ (pixelwise) & 0.8102 & 1.4705 & 0.7512 & 0.8909 & 0.9377 & 0.2834 & 0.3482 & 0.9220 & 0.9940 & 0.9997 \\

w/o plan parametrization & 544.3828 & 3151.1105 & 0.2983 & 0.4619 & 0.5569 & 219.6589 & 1234.7524 & 0.6671 & 0.7301 & 0.7420 \\

w/o exponential parametrization & 0.7685 & 1.4629 & 0.7721 & 0.9013 & 0.9421 & 0.2658 & 0.3294 & 0.9294 & 0.9937 & 0.9997 \\

w/o newton step & 0.7676 & 1.4630 & 0.7725 & 0.9015 & 0.9422 & 0.2741 & 0.3402 & 0.9272 & 0.9933 & 0.9997 \\

\bottomrule
\end{tabular}
\caption{Optical flow errors.}
\label{tab:abl_flow_err}
\end{table*}

\begin{table*}[htp]
    \centering
    \footnotesize
    \begin{tabular}{l|cccc|cccccc}
    \toprule
 &\multicolumn{4}{c}{Pose } &\multicolumn{6}{c}{Depth } \\ 

Method  
& \begin{tabular}[c]{@{}c@{}}Left l2 \\ (m)$\downarrow$ \end{tabular}& \begin{tabular}[c]{@{}c@{}}ATE  \\ (m) $\downarrow$ \end{tabular} &  \begin{tabular}[c]{@{}c@{}}Geom  \\ (m) $\downarrow$ \end{tabular}  &    \begin{tabular}[c]{@{}c@{}}Biinvrot l2  \\ (deg) $\downarrow$ \end{tabular}    
& Abs rel $\downarrow$  &Sqr rel $\downarrow$&  RMSE$\downarrow$    &Delta 1$\uparrow$  &Delta 2    $\uparrow$&Delta 3  $\uparrow$\\

\midrule
Dataset &\multicolumn{9}{c}{Blender 1 (small motion)} \\ \midrule

Base &\textbf{ 0.0066}  &\textbf{ 0.0056 } &\textbf{0.0050}  &\textbf{0.1806}  &\textbf{0.1381} & \textbf{0.2391}  &\textbf{0.8688} &\textbf{ 0.8358 } &\textbf{0.9263} & \textbf{0.9761}  \\

with regularization & 0.0072 & 0.0062 & 0.0053 & 0.1850 & 0.1399 & 0.2462 & 0.8777 & 0.8344 & 0.9236 & 0.9759 \\

with $k=1$ (pixelwise) & 0.0073 & 0.0062 & 0.0054 & 0.1883 & 0.1538 & 0.2673 & 0.9217 & 0.8087 & 0.9220 & 0.9734 \\

w/o plan parametrization & 0.0267 & 0.0250 & 0.0221 & 0.4317 & 0.4518 & 0.7926 & 1.7602 & 0.2319 & 0.5076 & 0.7142 \\

w/o exponential parametrization & 0.0073 & 0.0062 & 0.0054 & 0.1865 & 0.1392 & 0.2398 & 0.8694 & 0.8340 & 0.9252 & 0.9763 \\

w/o newton step & 0.0073 & 0.0062 & 0.0054 & 0.1861 & 0.1393 & 0.2397 & 0.8696 & 0.8342 & 0.9253 & 0.9763 \\

\midrule
Dataset &\multicolumn{9}{c}{Blender 2 (micro motion)} \\ \midrule 

Base &  \textbf{0.0022} & \textbf{0.0022} & \textbf{0.0020 } &\textbf{ 0.0245}  & \textbf{0.1383}  & \textbf{0.1962 }& \textbf{1.1521 } & \textbf{0.7996 }& \textbf{0.9819 }& \textbf{0.9983}  \\

with regularization & 0.0023 & 0.0022 & 0.0020 & 0.0256 & 0.1766 & 0.2935 & 1.3640 & 0.6943 & 0.9498 & 0.9948 \\

with $k=1$ (pixelwise) & 0.0024 & 0.0024 & 0.0022 & 0.0287 & 0.1750 & 0.2928 & 1.3629 & 0.6932 & 0.9503 & 0.9952 \\

w/o plan parametrization & 0.0040 & 0.0039 & 0.0037 & 0.0430 & 0.2571 & 0.5270 & 1.7727 & 0.5005 & 0.8881 & 0.9857 \\

w/o exponential parametrization & 0.0022 & 0.0022 & 0.0020 & 0.0261 & 0.1755 & 0.2908 & 1.3590 & 0.6981 & 0.9510 & 0.9949 \\

w/o newton step & 0.0023 & 0.0022 & 0.0021 & 0.0258 & 0.1818 & 0.3040 & 1.3837 & 0.6807 & 0.9458 & 0.9946 \\

\bottomrule
\end{tabular}
\caption{Pose and depth errors metrics on the two proposed synthetic bursts datasets.}
\label{tab:abl_pose_err}
\end{table*}

\subsection{Comparison with monocular method}\label{app:comp_mono}
Monocular depth estimation methods can only estimate depth up to an affine transformation. Therefore,  we evaluate them up to an affine correction. It does not make sense to compare them to the binocular method with linear correction as in Table \ref{tab:pose_err}. On the other hand, to compare them to the latter, we must recalculate the error of each of the methods in Table \ref{tab:pose_err} with an affine correction. The results are reported in Table \ref{tab:depth_err_aff}.

\subsection{Estimated occlusion mask}\label{app:occ}
We use the fixed point algorithm described in \ref{app:fp} on the depth map obtained at the optimization's last step and note the points for which the fixed point algorithm does not converge.
We use a threshold and a maximum number of iterations to construct the non-convergent set. This set constitutes a partial occlusion mask. It can be used in downstream tasks to avoid aggregating erroneous information because it is occluded. Fig. \ref{fig:occ_mask} shows examples of masks on synthetic data.

\newcommand\xda{0.272}
\newcommand\xdb{0.728}
 \begin{figure}[htbp]
    \setlength\tabcolsep{0.5pt}
    \renewcommand{\arraystretch}{0.5}
    \centering
         \begin{tabular}{cc}
            \includegraphics[width=\xda\textwidth]{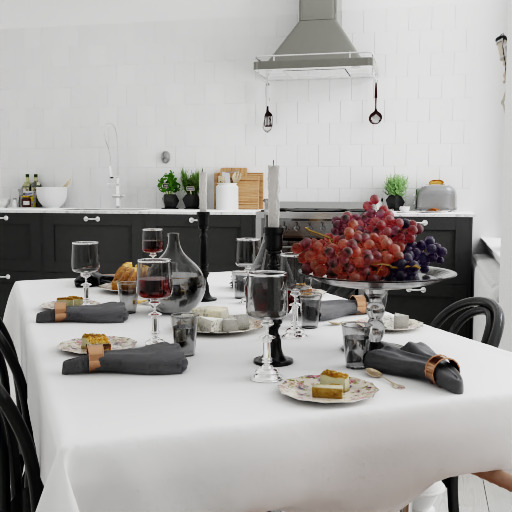}&
            \includegraphics[width=\xdb\textwidth]{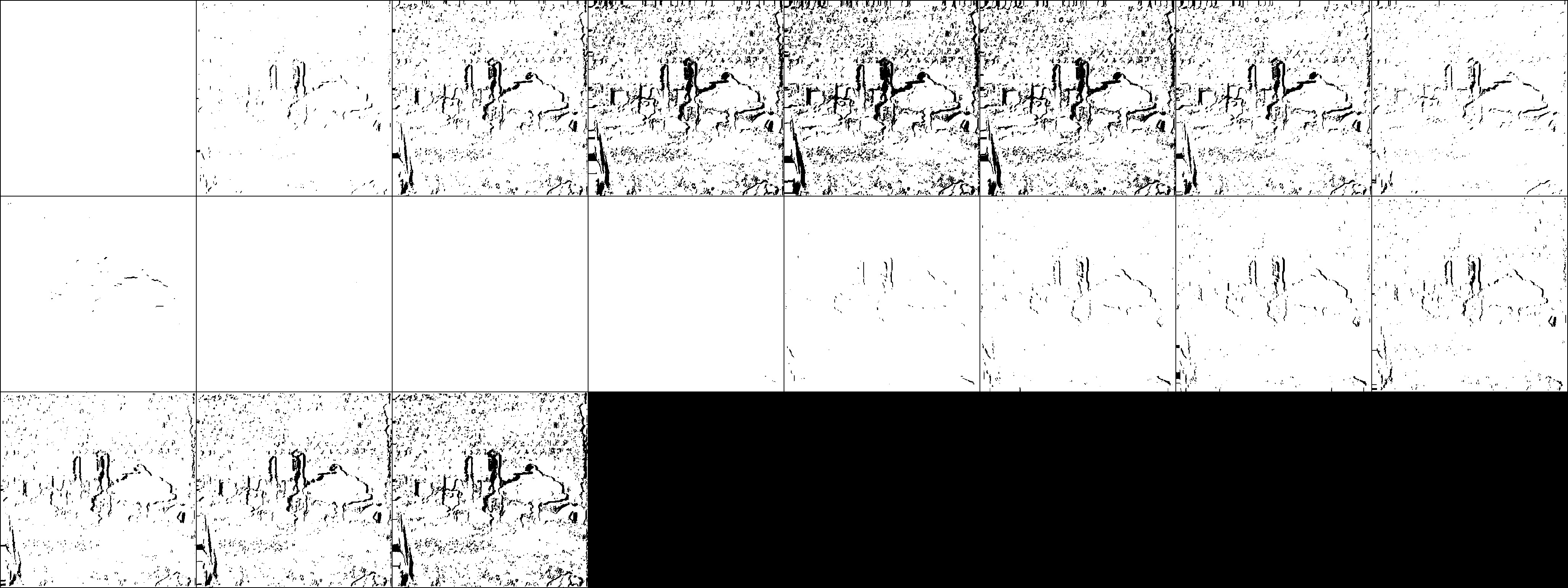}\\
            \includegraphics[width=\xda\textwidth]{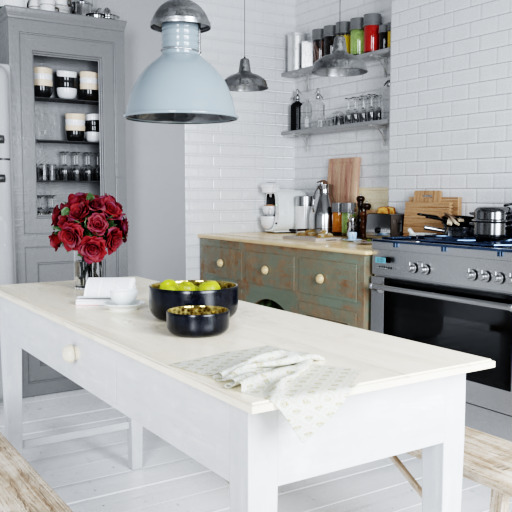}&
            \includegraphics[width=\xdb\textwidth]{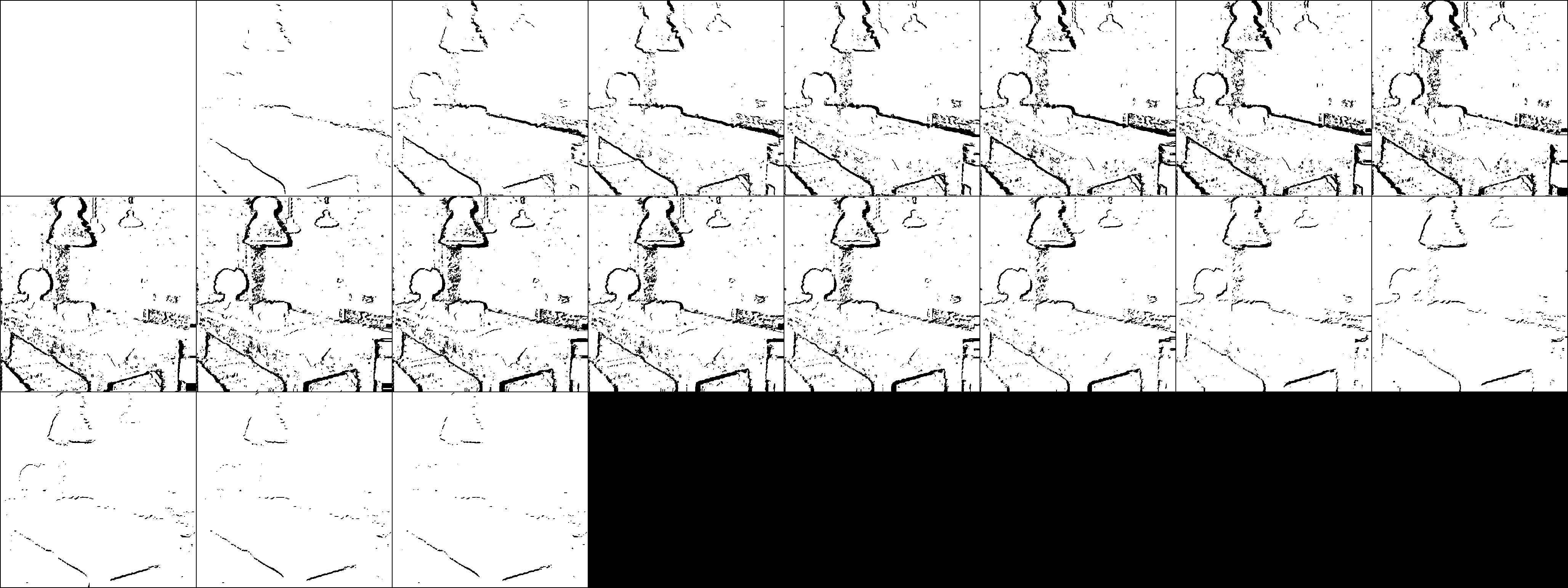}\\
            \includegraphics[width=\xda\textwidth]{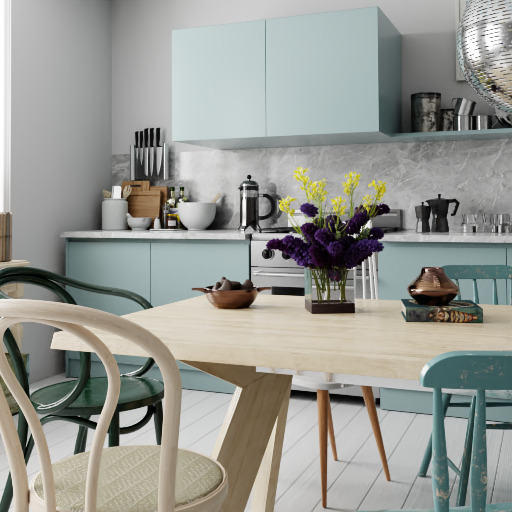}&
            \includegraphics[width=\xdb\textwidth]{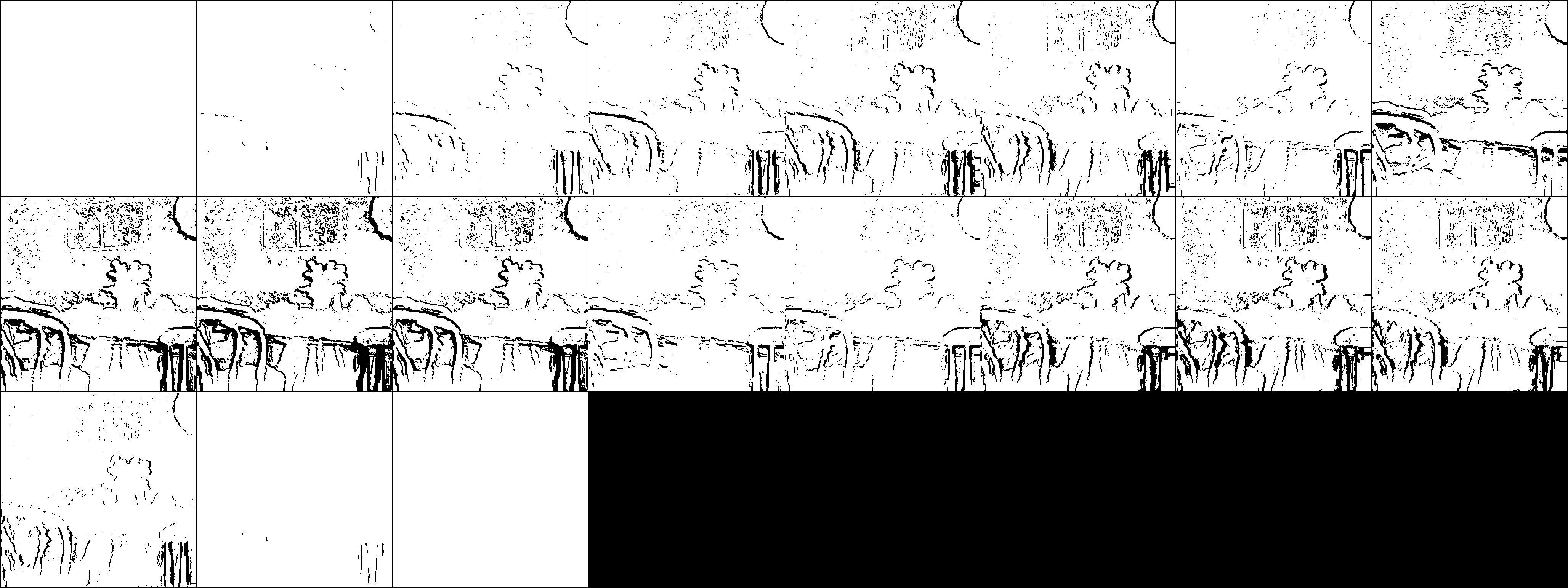}\\
            \includegraphics[width=\xda\textwidth]{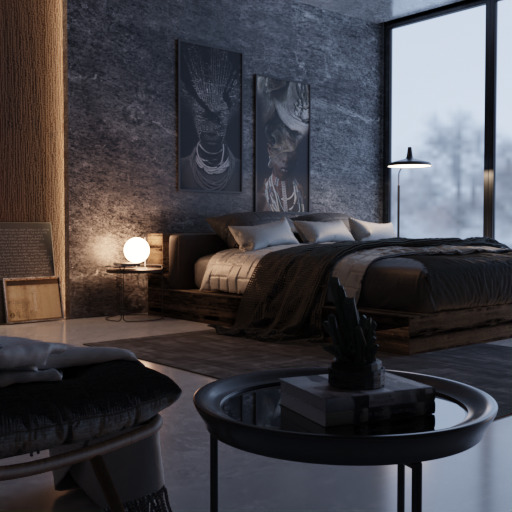}&
            \includegraphics[width=\xdb\textwidth]{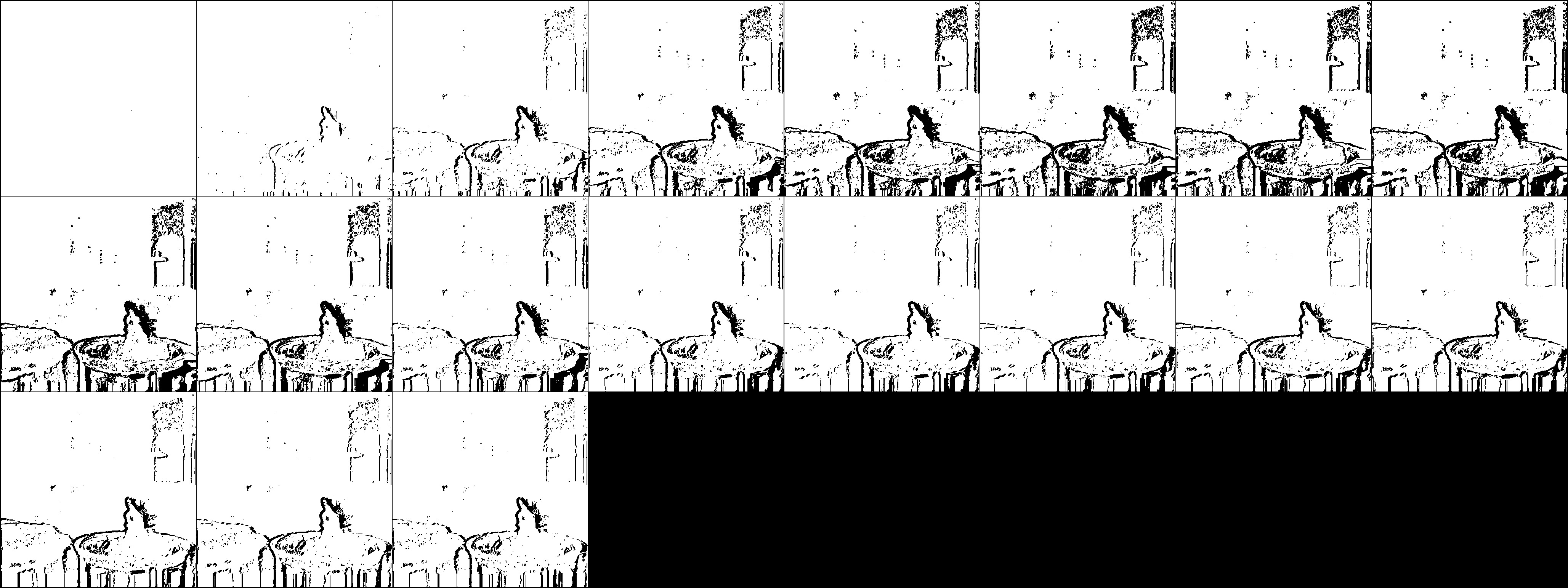}
 \end{tabular}
     \caption{Partial occlusion mask obtained using the fixed point algorithm for four examples of the Blender 2 dataset.}
     \label{fig:occ_mask}
 \end{figure}

\subsection{Depthmaps}\label{app:depth}

We provide additional examples of depth maps from both synthetic bursts (Fig.~\ref{fig:depthmap_synthetic_appendix}) and real bursts (Fig.~\ref{fig:depthmap_real_appendix}).
All disparity maps were aligned to the ground truth with an affine transform by using the least square criterion of \cite{ranftl2020towards}.

For a fair comparison, we also show the results of DfUSMC without their additional depth map filtering, which is essential to obtain a visually appealing depth map.
However, this step introduces a stratification of the depth map, which is not present in our method.

\newcommand\xd{0.11}
\newcommand\fs{\tiny}
 \begin{figure}[htbp]
         \setlength\tabcolsep{0.5pt}
     \renewcommand{\arraystretch}{0.5}
     \centering
         \begin{tabular}{cc|cc|ccccc}
            \multicolumn{2}{c}{} &\multicolumn{2}{c}{Learning based} & \multicolumn{5}{c}{Optimization based} \\
            \includegraphics[width=\xd\textwidth]{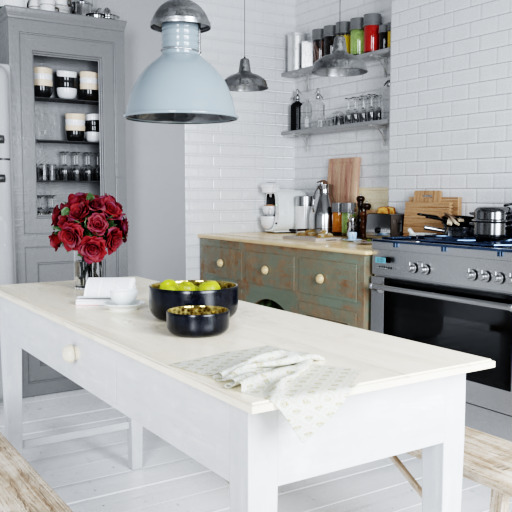}&
            \includegraphics[width=\xd\textwidth]{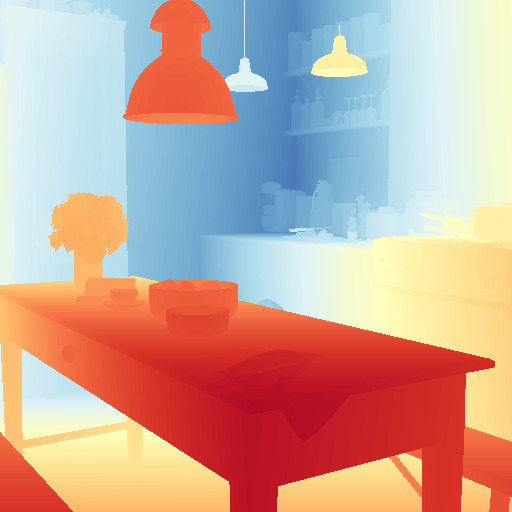}&
            \includegraphics[width=\xd\textwidth]{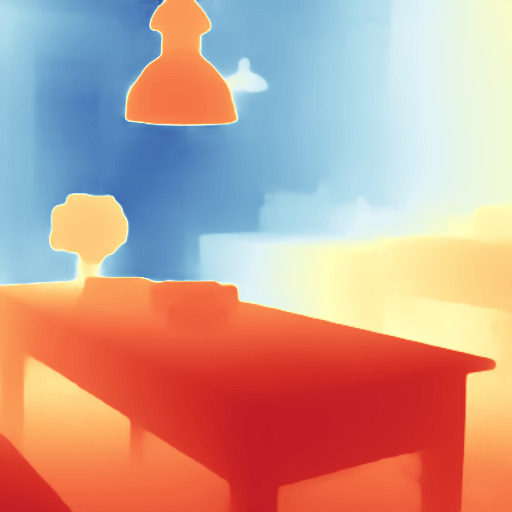}&
            \includegraphics[width=\xd\textwidth]{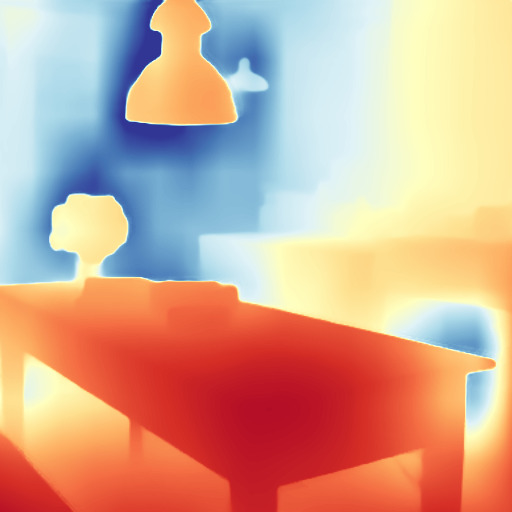}&
            \includegraphics[width=\xd\textwidth]{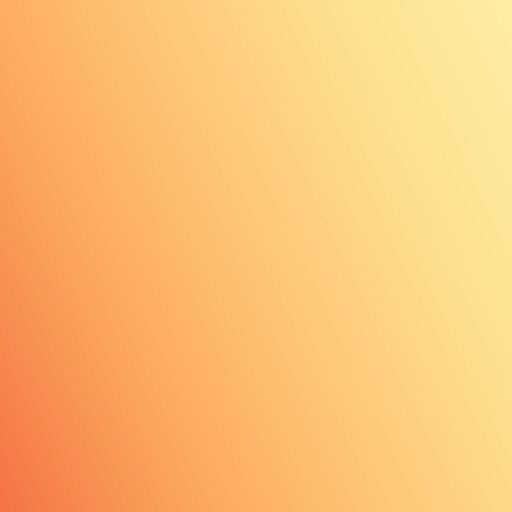}&
            \includegraphics[width=\xd\textwidth]{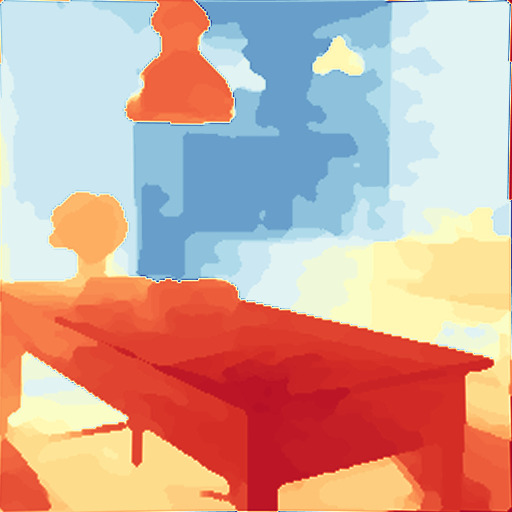}&
            \includegraphics[width=\xd\textwidth]{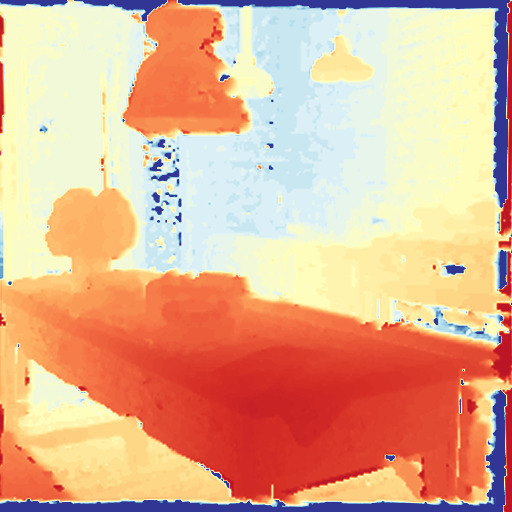}&
            \includegraphics[width=\xd\textwidth]{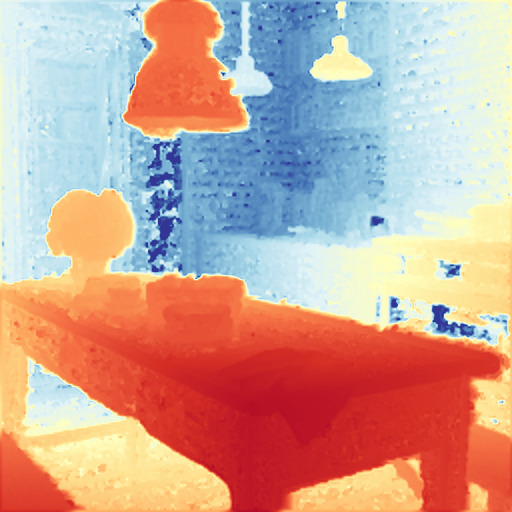}  &         \includegraphics[width=\xd\textwidth]{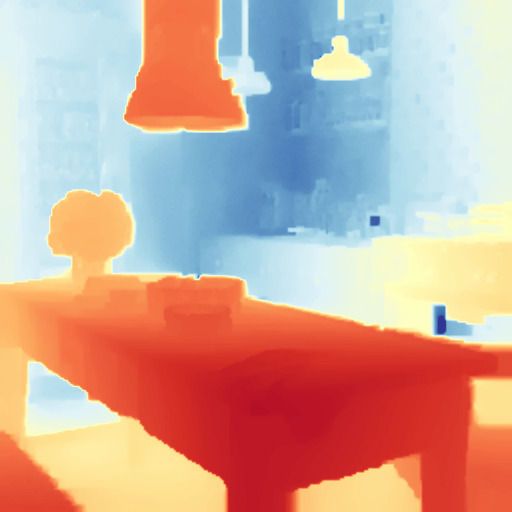} \\
            \includegraphics[width=\xd\textwidth]{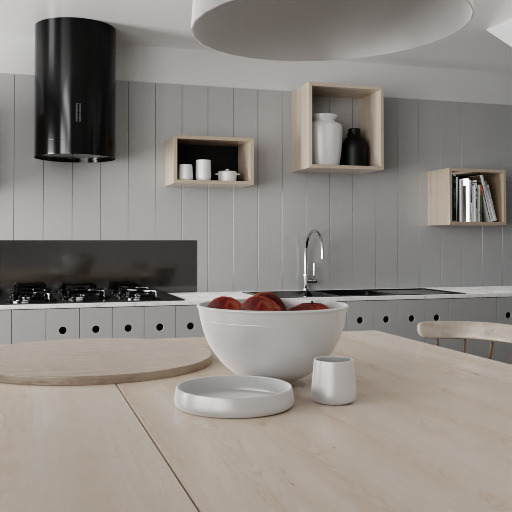}&
            \includegraphics[width=\xd\textwidth]{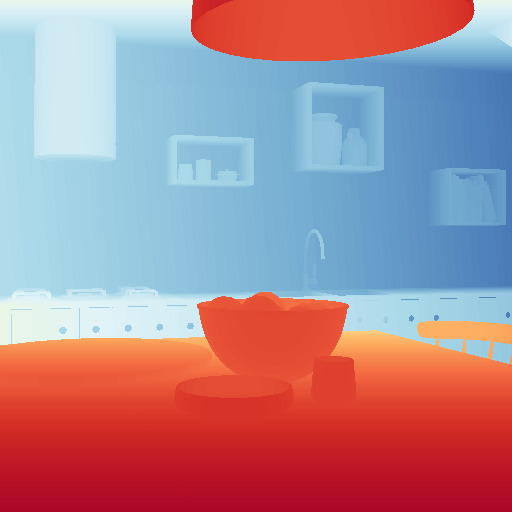}&
            \includegraphics[width=\xd\textwidth]{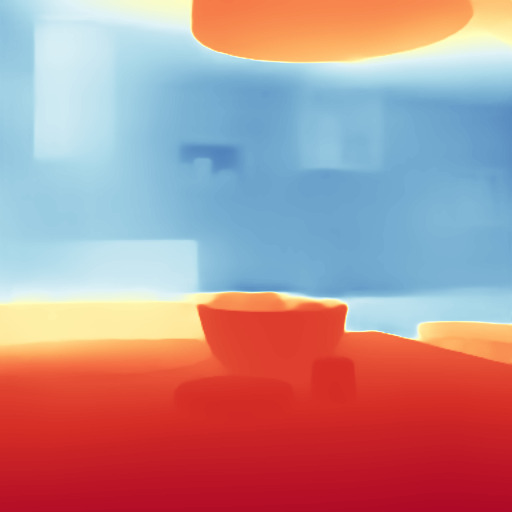}&
            \includegraphics[width=\xd\textwidth]{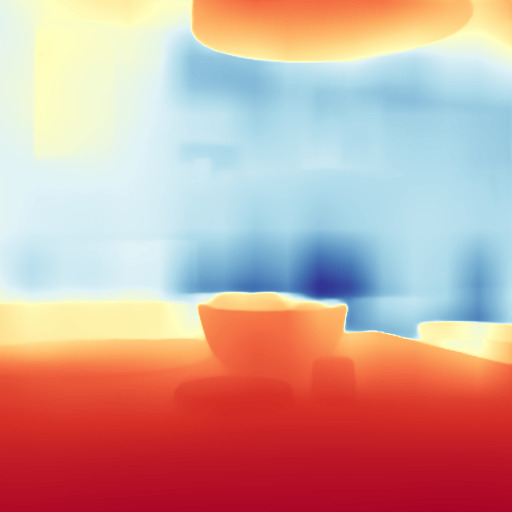}&
            \includegraphics[width=\xd\textwidth]{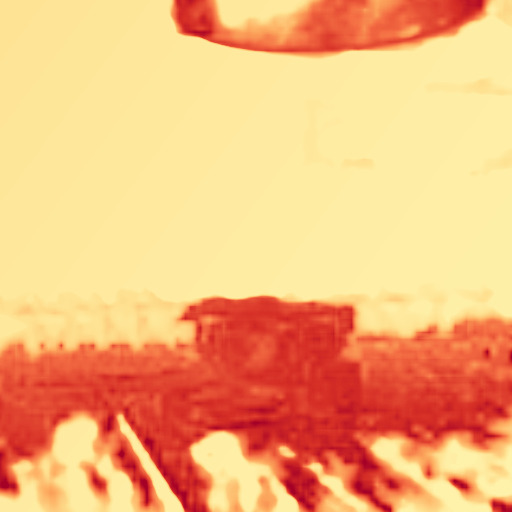}&
            \includegraphics[width=\xd\textwidth]{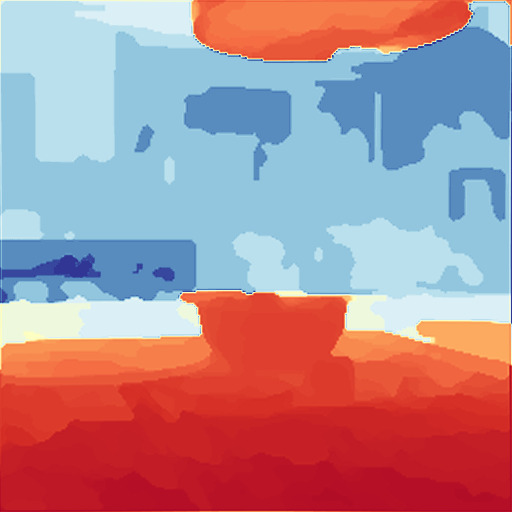}&
            \includegraphics[width=\xd\textwidth]{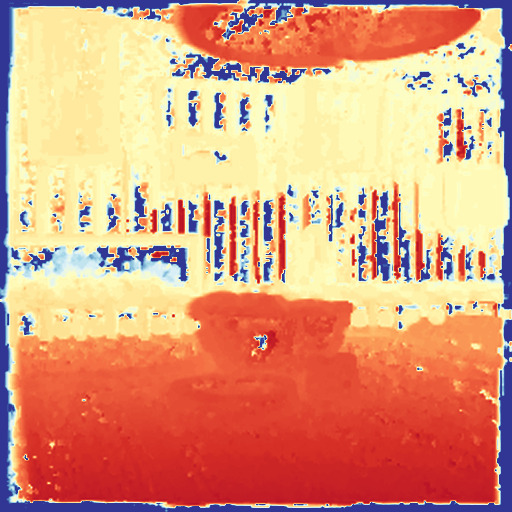}&
            \includegraphics[width=\xd\textwidth]{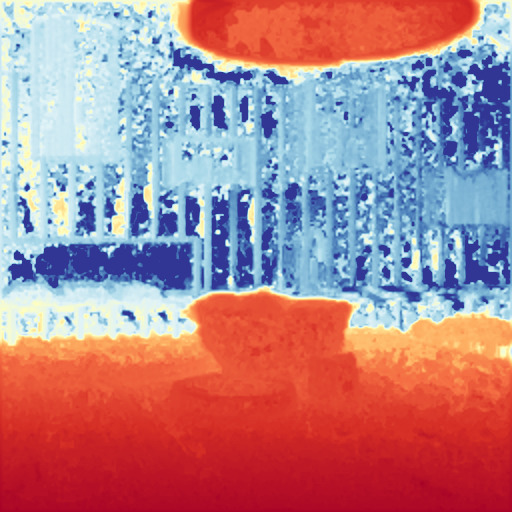}  &         \includegraphics[width=\xd\textwidth]{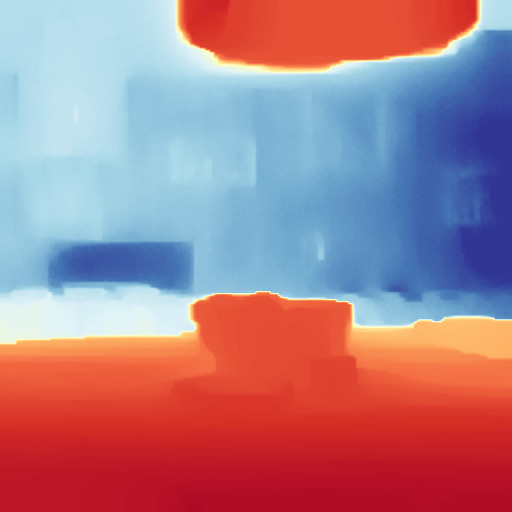} \\
            \includegraphics[width=\xd\textwidth]{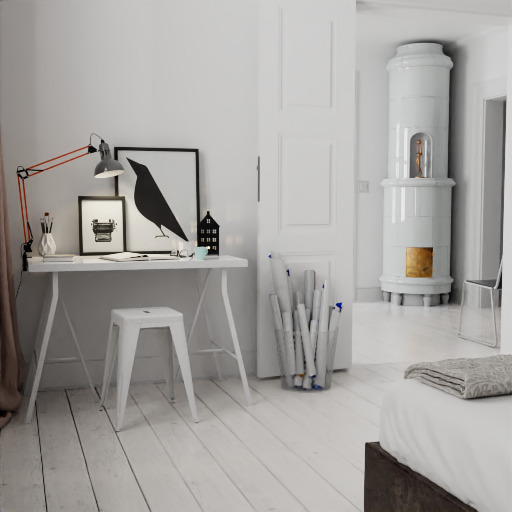}&
            \includegraphics[width=\xd\textwidth]{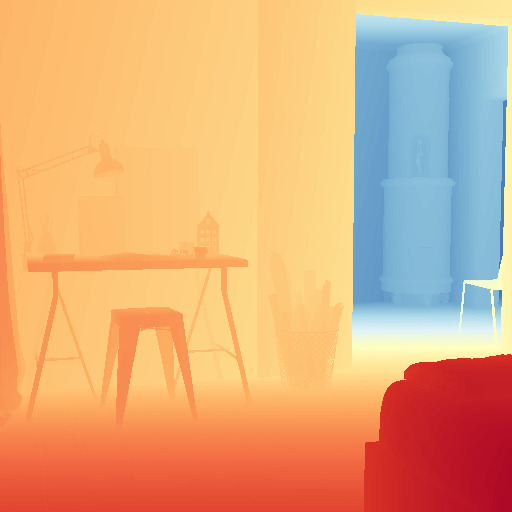}&
            \includegraphics[width=\xd\textwidth]{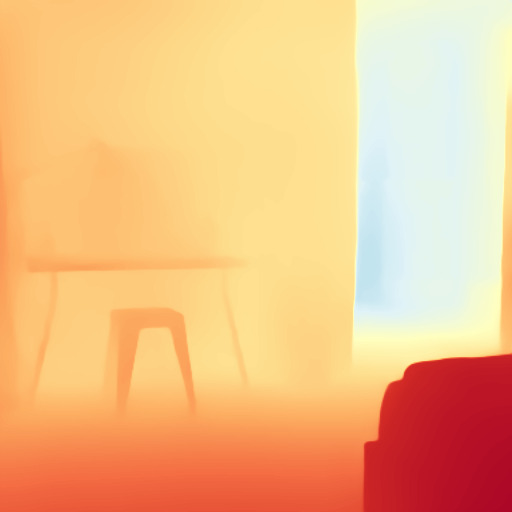}&
            \includegraphics[width=\xd\textwidth]{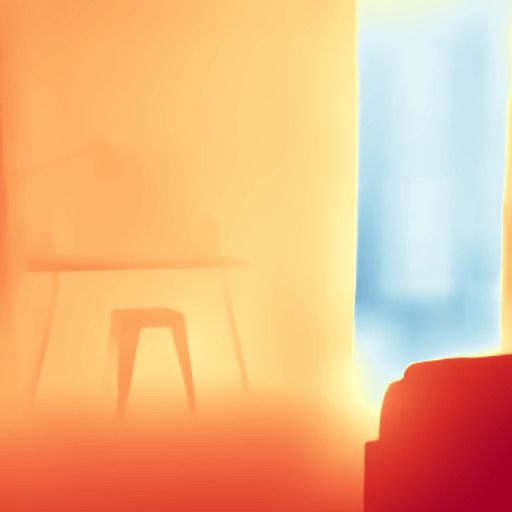}&
            \includegraphics[width=\xd\textwidth]{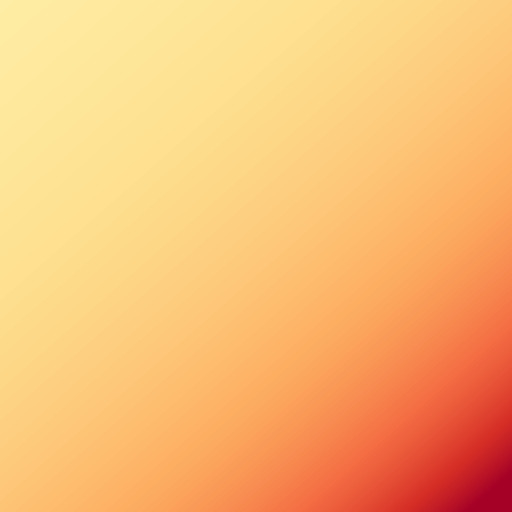}&
            \includegraphics[width=\xd\textwidth]{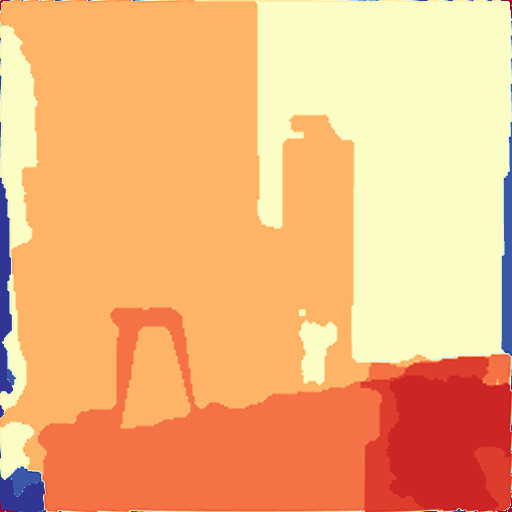}&
            \includegraphics[width=\xd\textwidth]{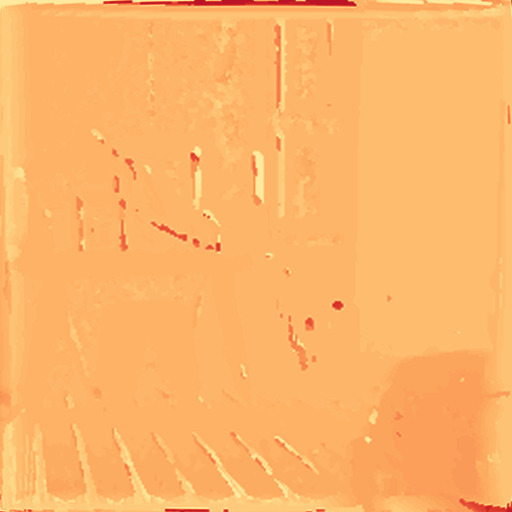}&
            \includegraphics[width=\xd\textwidth]{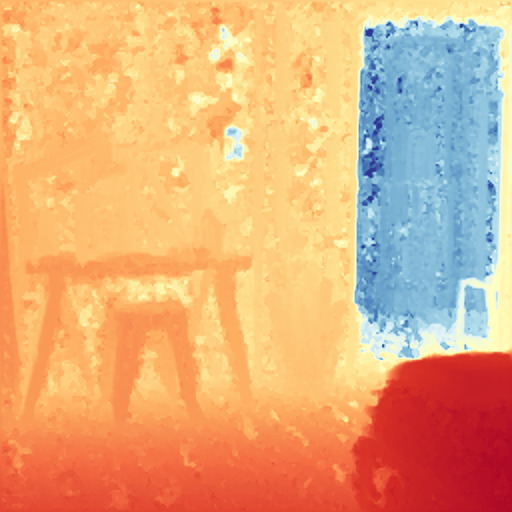}  &         \includegraphics[width=\xd\textwidth]{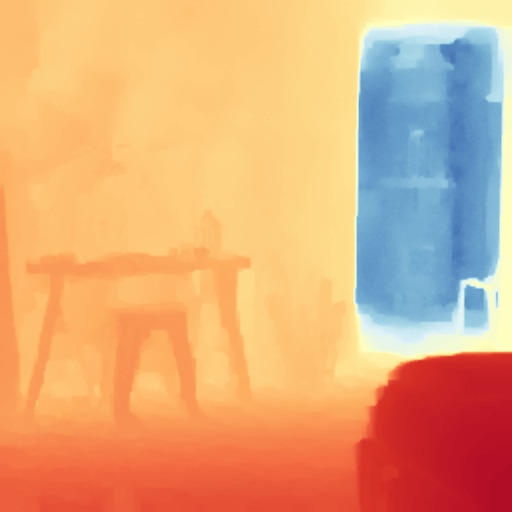} \\
            \includegraphics[width=\xd\textwidth]{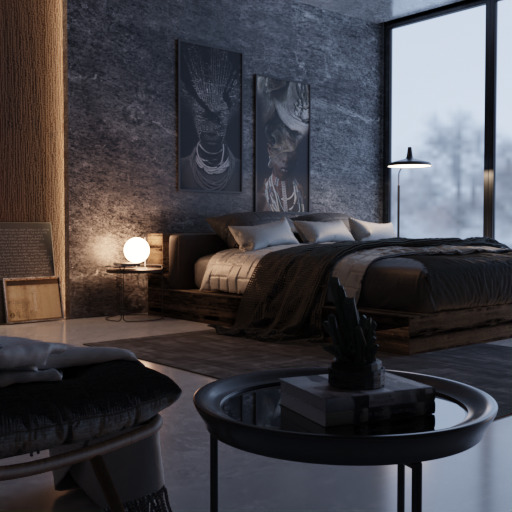}&
            \includegraphics[width=\xd\textwidth]{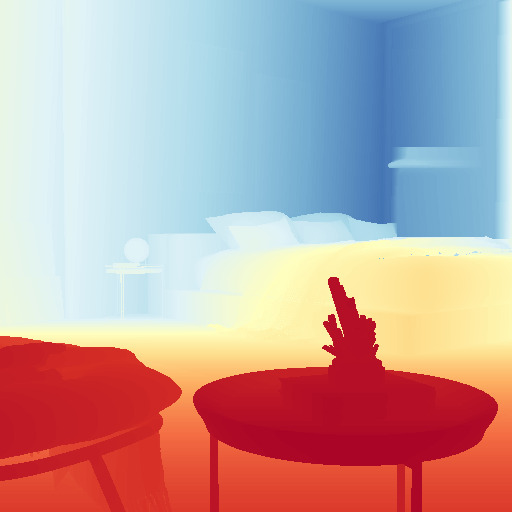}&
            \includegraphics[width=\xd\textwidth]{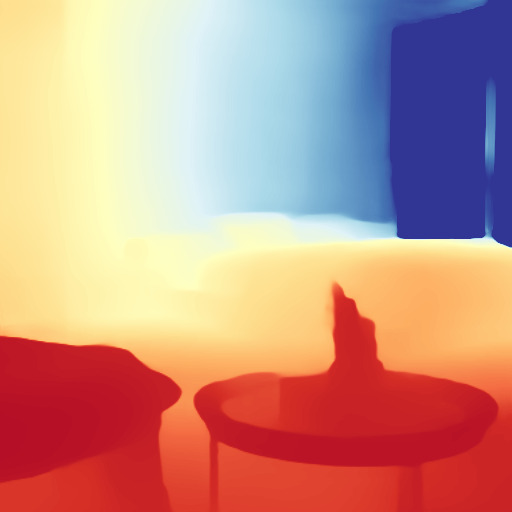}&
            \includegraphics[width=\xd\textwidth]{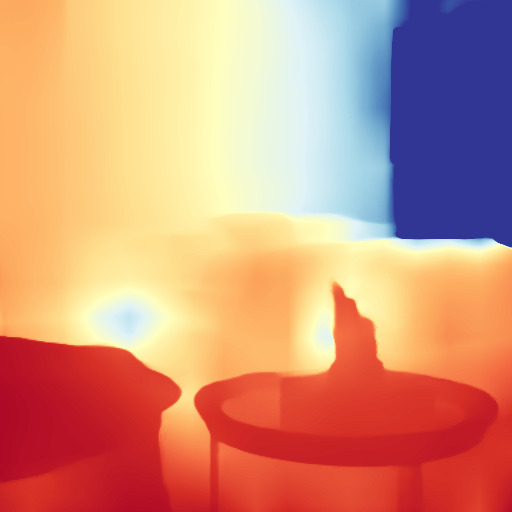}&
            \includegraphics[width=\xd\textwidth]{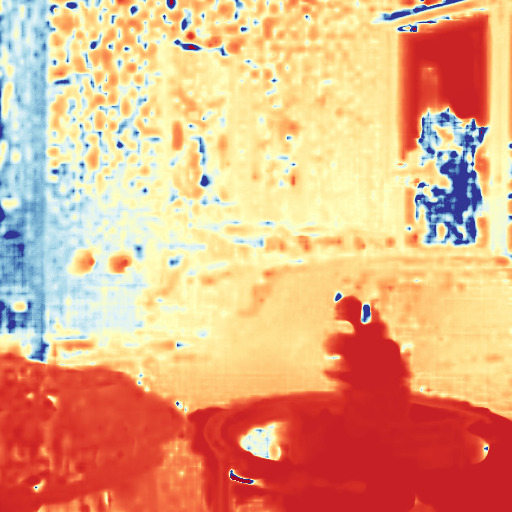}&
            \includegraphics[width=\xd\textwidth]{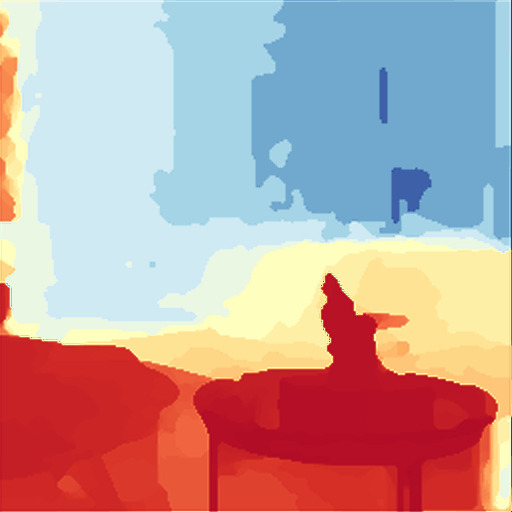}&
            \includegraphics[width=\xd\textwidth]{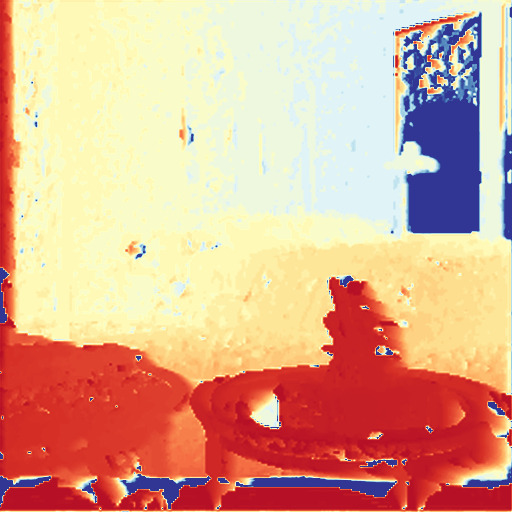}&
            \includegraphics[width=\xd\textwidth]{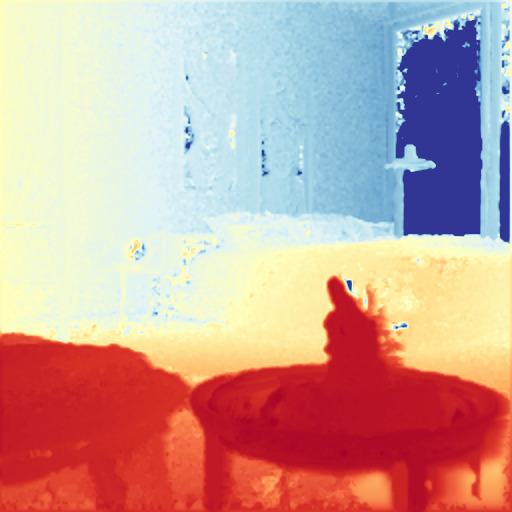}  &         \includegraphics[width=\xd\textwidth]{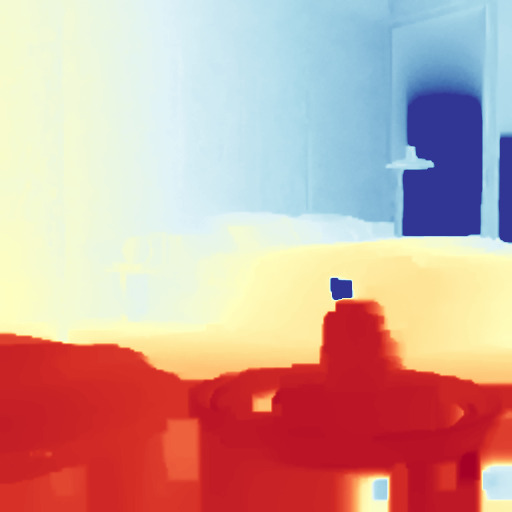} \\
            \fs Ref. image & 
            \fs Groundtruth &  
            \fs Midas \cite{ranftl2020towards} & 
            \fs RCVD\cite{kopf2021robust} &
            \fs Saop \cite{chugunov2022shakes} &
            \fs DfUSMC \cite{ha2016high} &
            \fs DfUSMC no filt.&
            \fs Ours &
            \fs Ours + reg.\\ 
 \end{tabular}
     \caption{Depth estimation from synthetic bursts (\emph{Blender 2} dataset).}
     \label{fig:depthmap_synthetic_appendix}
 \end{figure}

\newcommand\xdr{0.18}

 \begin{figure}[htbp]
         \setlength\tabcolsep{0.5pt}
     \renewcommand{\arraystretch}{0.5}
     \centering
         \begin{tabular}{ccccc}
            
            \includegraphics[width=\xdr\textwidth]{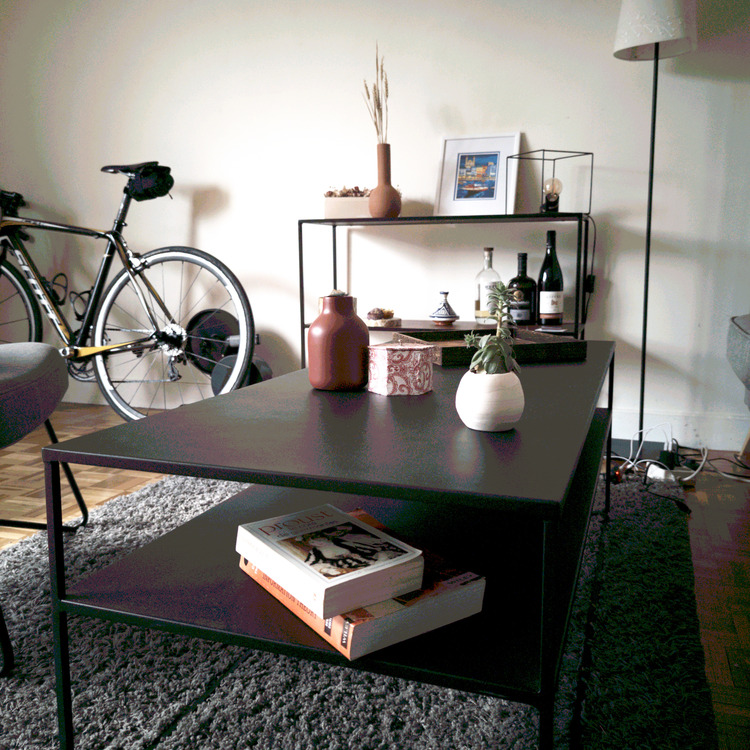}&
            \includegraphics[width=\xdr\textwidth]{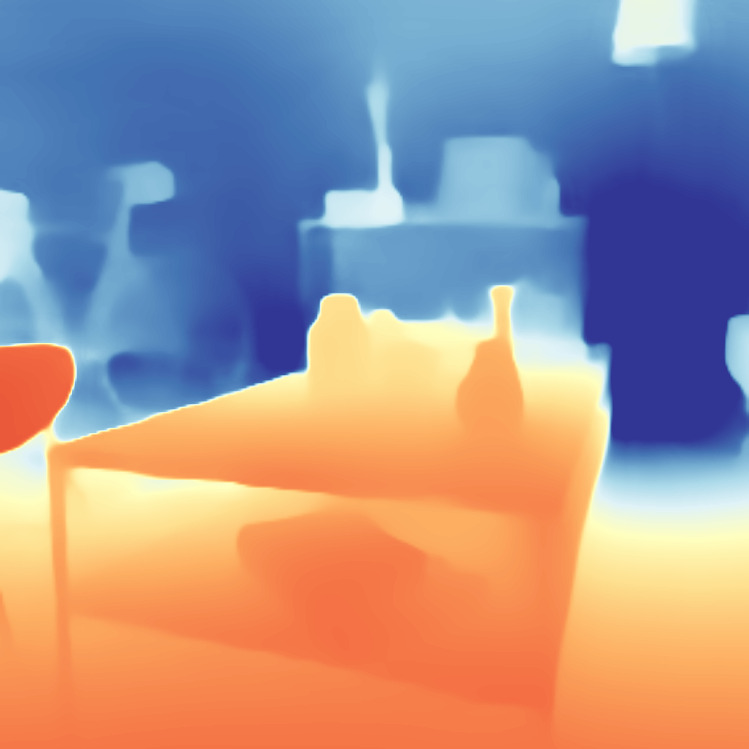} &
            \includegraphics[width=\xdr\textwidth]{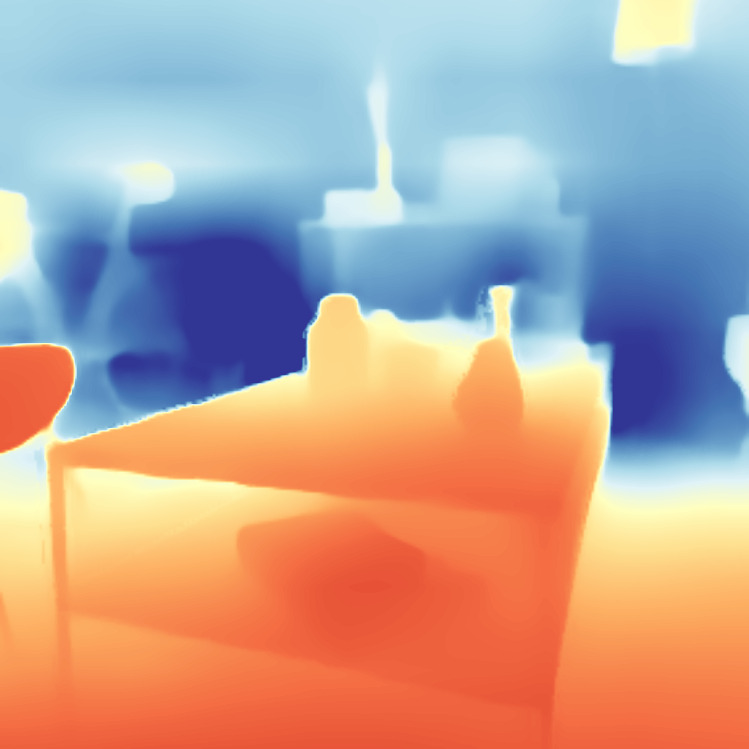} &
            \includegraphics[width=\xdr\textwidth]{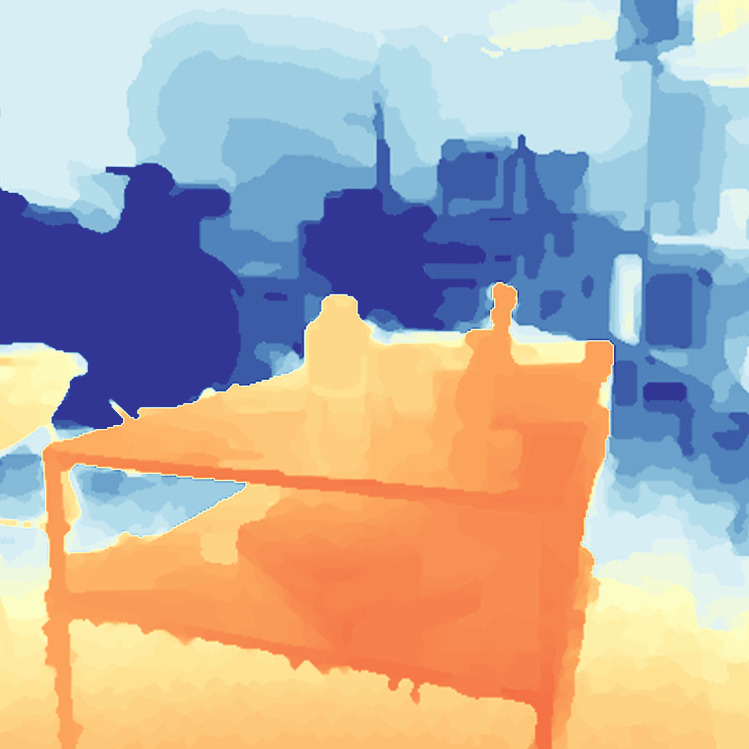} &
            \includegraphics[width=\xdr\textwidth]{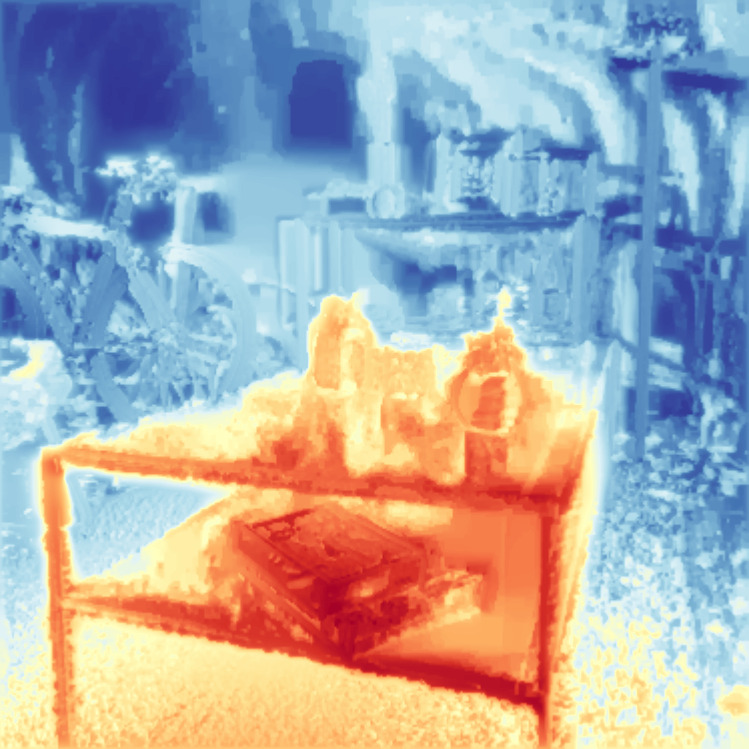} \\
            \includegraphics[width=\xdr\textwidth]{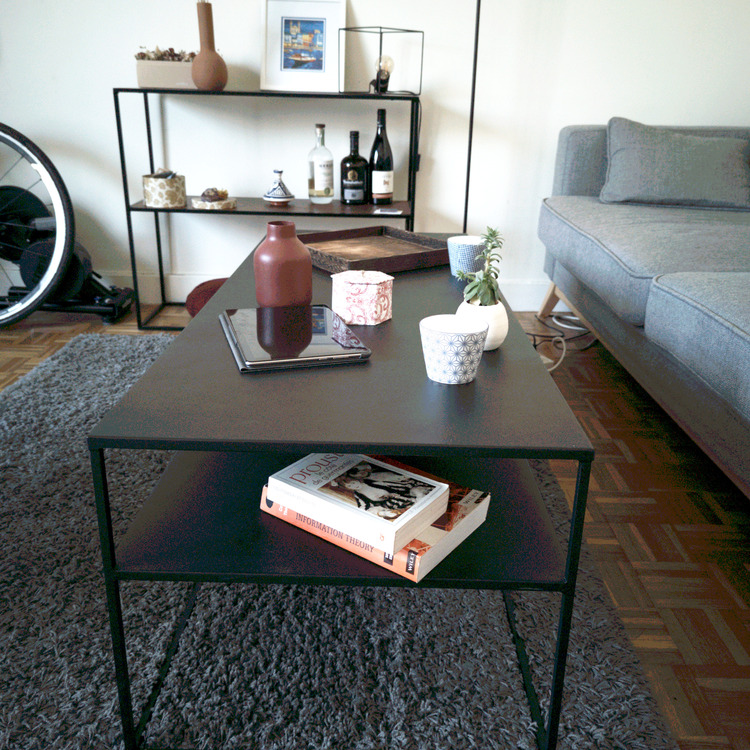}&
            \includegraphics[width=\xdr\textwidth]{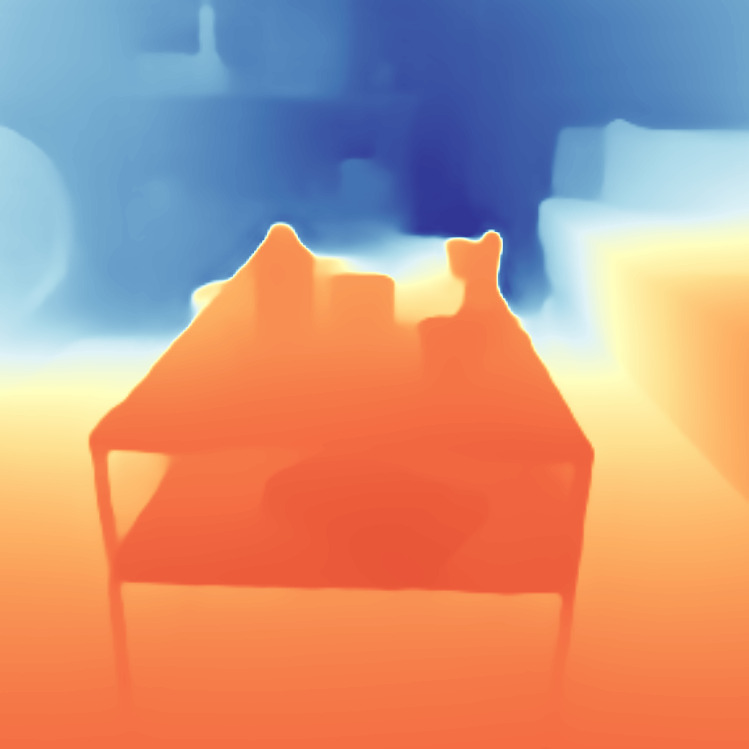} &
            \includegraphics[width=\xdr\textwidth]{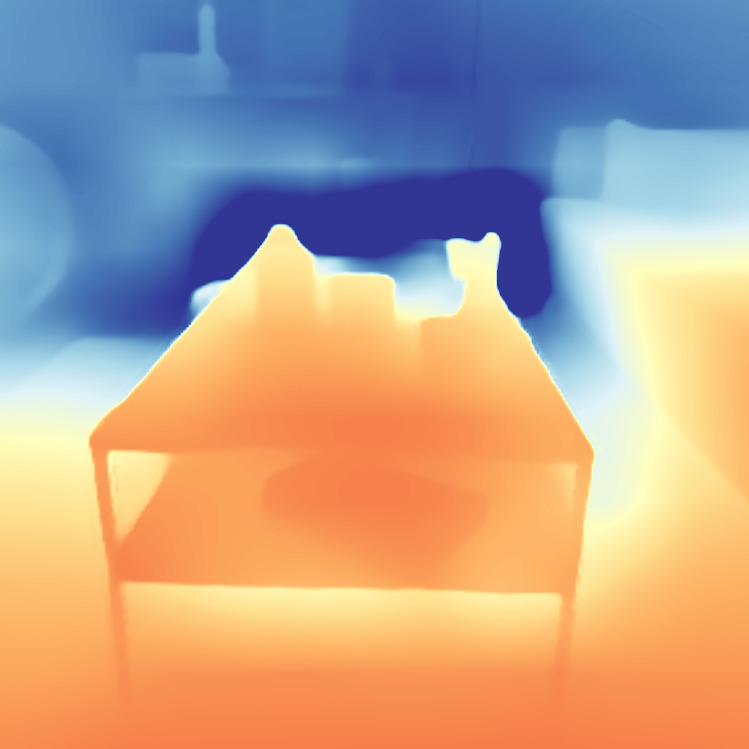} &
            \includegraphics[width=\xdr\textwidth]{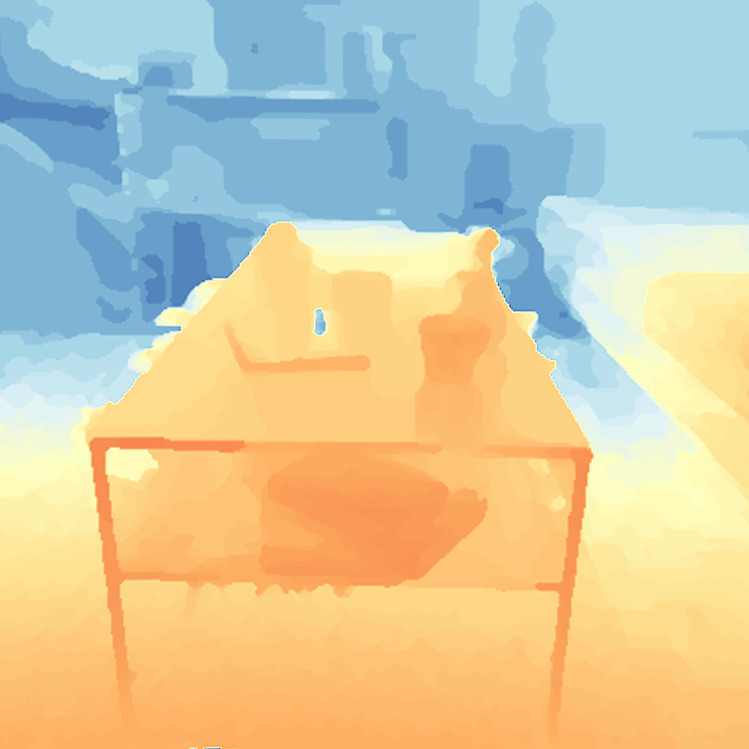} &
            \includegraphics[width=\xdr\textwidth]{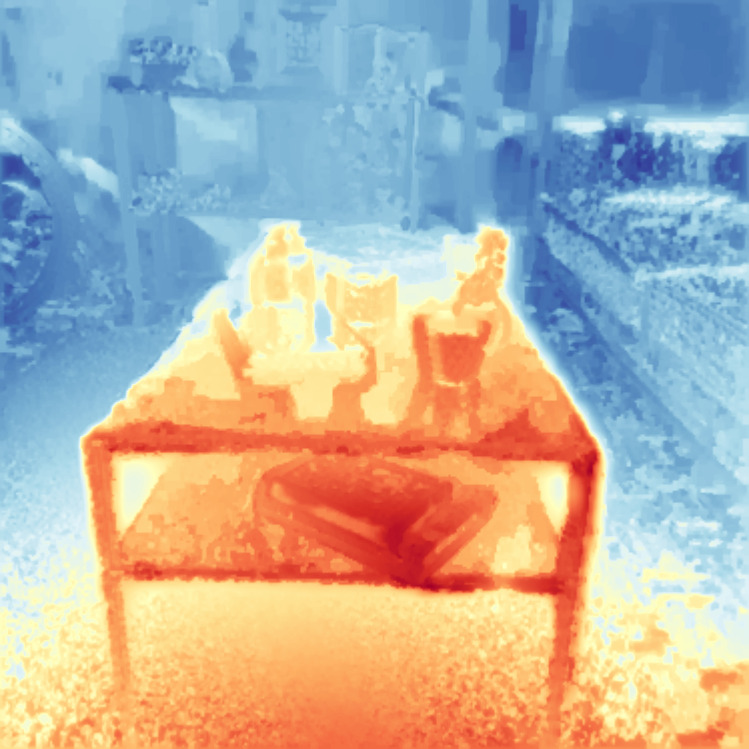} \\
            Ref. image &  
            Midas \cite{ranftl2020towards} & 
            RCVD\cite{kopf2021robust} &
            DfUSMC \cite{ha2016high} &
            Ours \\
 \end{tabular}
     \caption{Depth estimation from real bursts.}
     \label{fig:depthmap_real_appendix}
 \end{figure}

\subsection{Pose estimation visualization}\label{app:pose}
To visualize the positions the algorithm approximates, we can look at the translation part of the positions. Because our images come from a burst, we use the temporal coherence of the series of pictures and can trace the trajectory of the camera center during the burst. After rescaling, we compare the trajectory approximated by the algorithm to the trajectory used to create the burst in Blender. Fig. \ref{fig:traj} shows examples of trajectories for different images of the Blender 2 dataset during the last three stages.

\newcommand\xcc{0.275}
 \begin{figure}[htbp]
         \setlength\tabcolsep{0.5pt}
     \renewcommand{\arraystretch}{0.5}
     \centering
         \begin{tabular}{c|ccc}
            Scene & Stage 0 & Stage 1 & Stage 2\\ \hline
            \raisebox{0.35\height}{\rotatebox[origin=l]{90}{Fig. \ref{fig:occ_mask} row 1}} &
            \includegraphics[width=\xcc\textwidth]{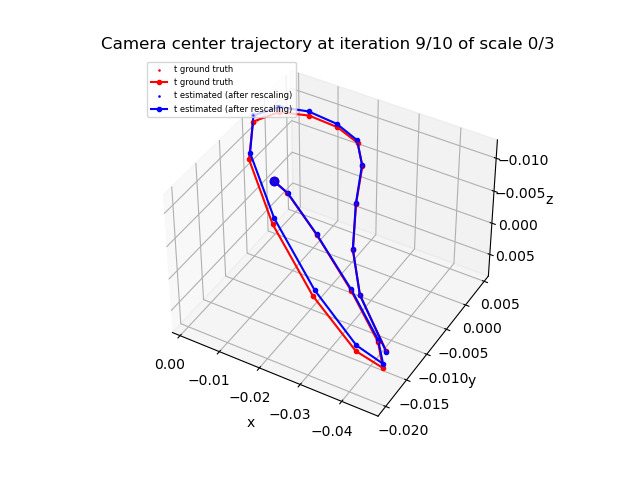} &
            \includegraphics[width=\xcc\textwidth]{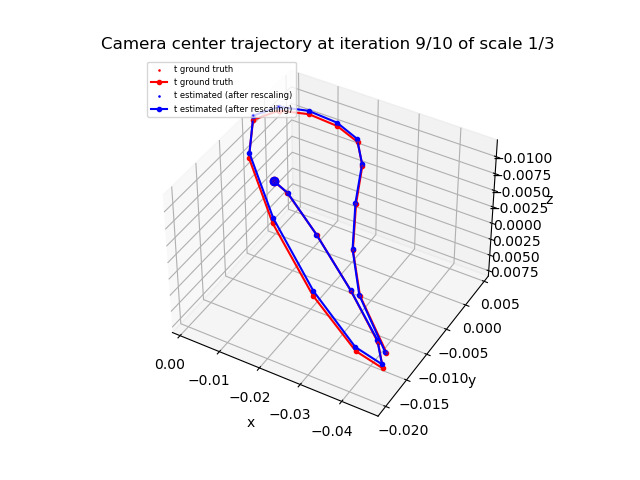} &
            \includegraphics[width=\xcc\textwidth]{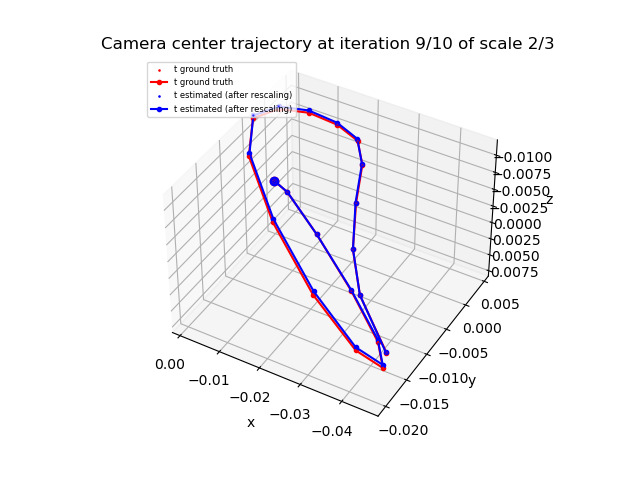}\\
            \raisebox{0.35\height}{\rotatebox[origin=l]{90}{Fig. \ref{fig:occ_mask} row 2}} &
            \includegraphics[width=\xcc\textwidth]{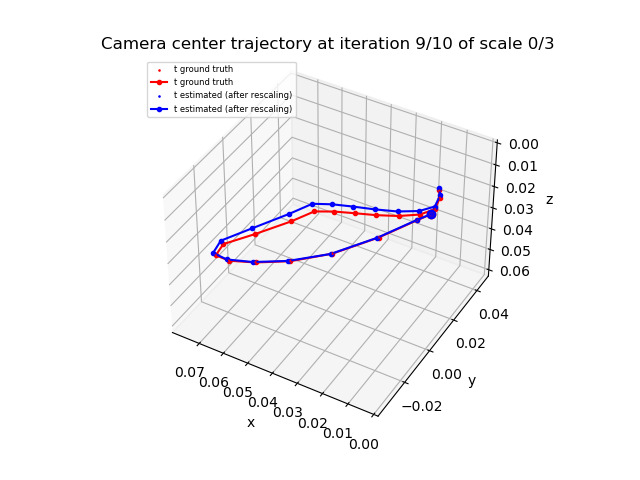} &
            \includegraphics[width=\xcc\textwidth]{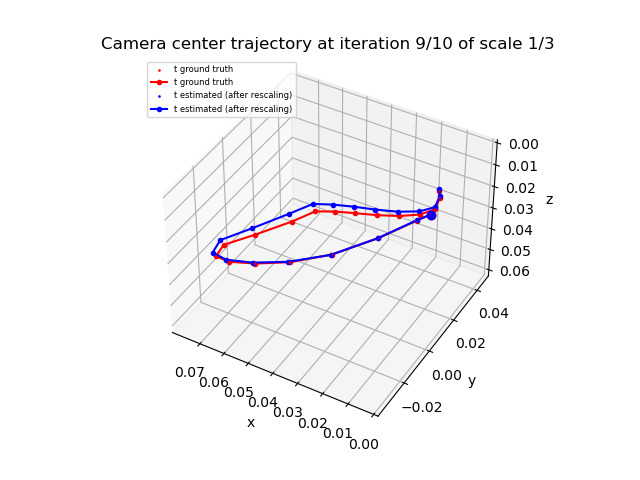} &
            \includegraphics[width=\xcc\textwidth]{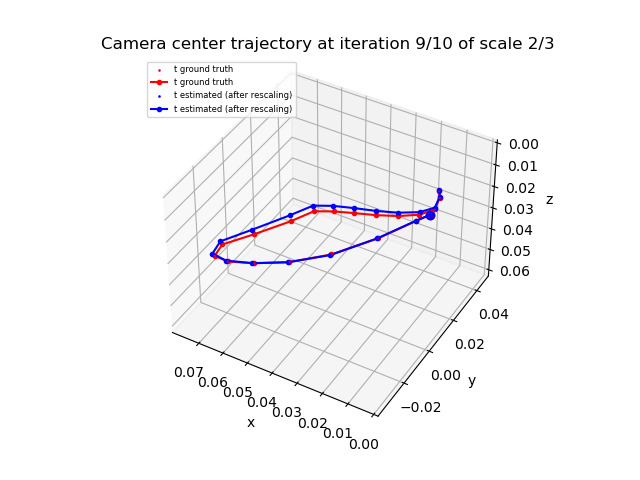}\\
            \raisebox{0.35\height}{\rotatebox[origin=l]{90}{Fig. \ref{fig:occ_mask} row 3}} &
            \includegraphics[width=\xcc\textwidth]{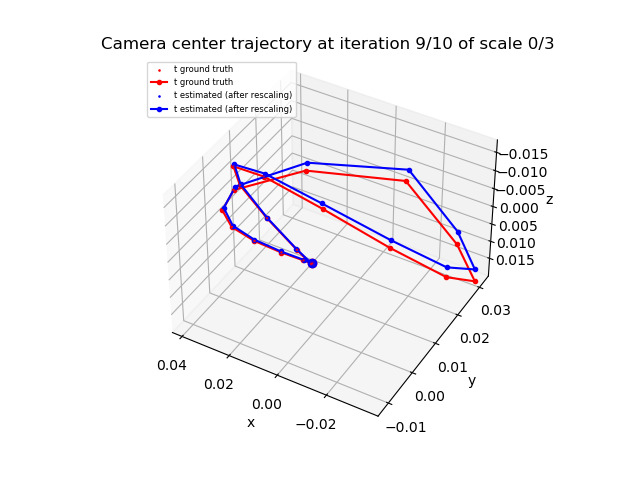} &
            \includegraphics[width=\xcc\textwidth]{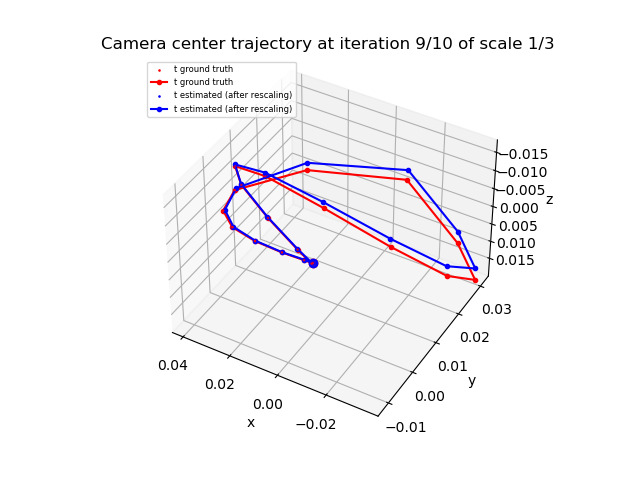} &
            \includegraphics[width=\xcc\textwidth]{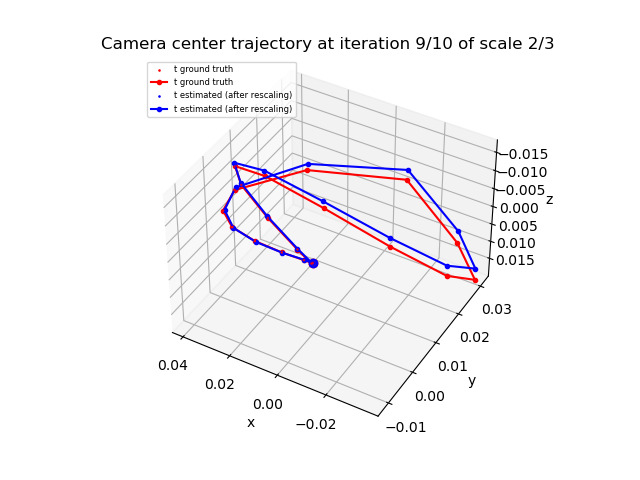}\\
            \raisebox{0.35\height}{\rotatebox[origin=l]{90}{Fig. \ref{fig:occ_mask} row 4}} &
            \includegraphics[width=\xcc\textwidth]{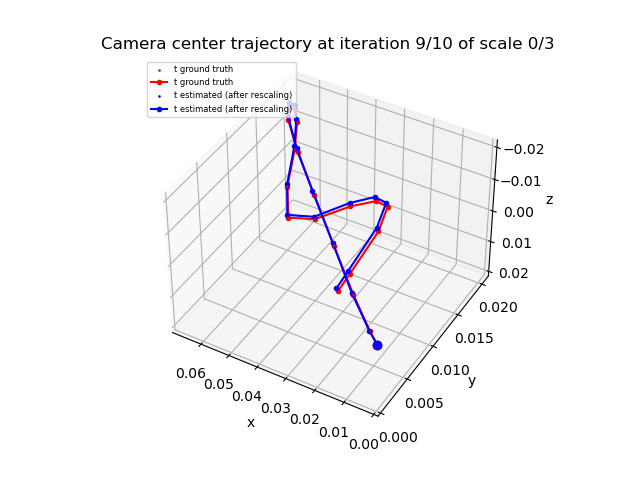} &
            \includegraphics[width=\xcc\textwidth]{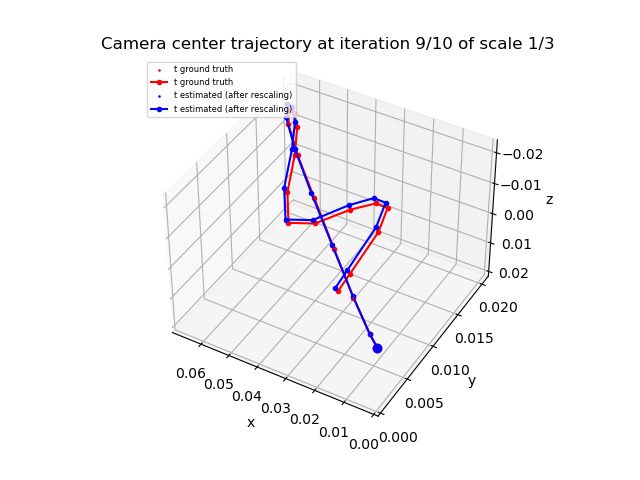} &
            \includegraphics[width=\xcc\textwidth]{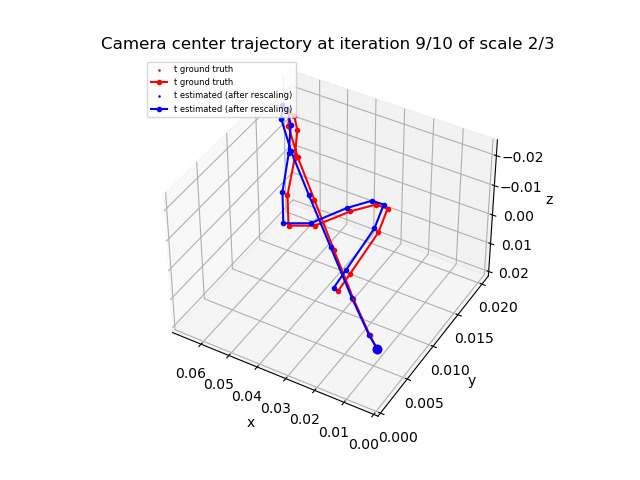}\\
            \raisebox{0.35\height}{\rotatebox[origin=l]{90}{Fig. \ref{fig:depthmap_synthetic_appendix} row 2}} &
            \includegraphics[width=\xcc\textwidth]{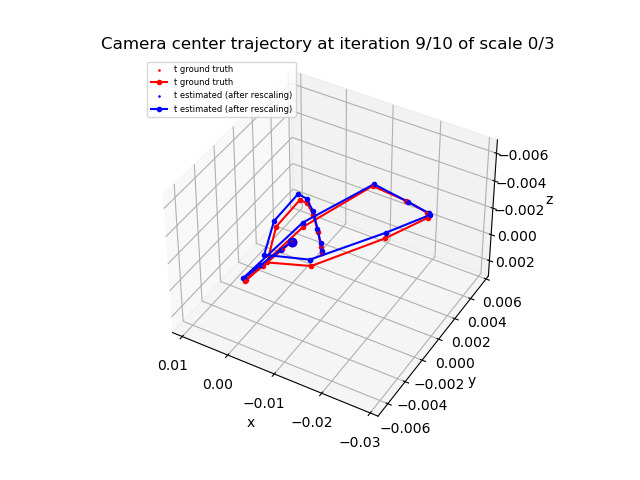} &
            \includegraphics[width=\xcc\textwidth]{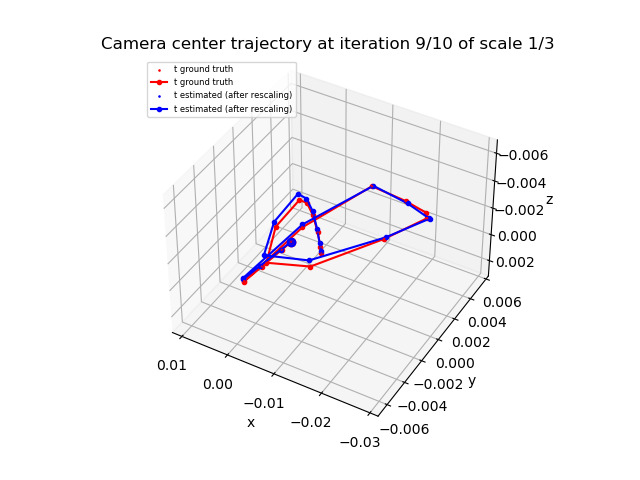} &
            \includegraphics[width=\xcc\textwidth]{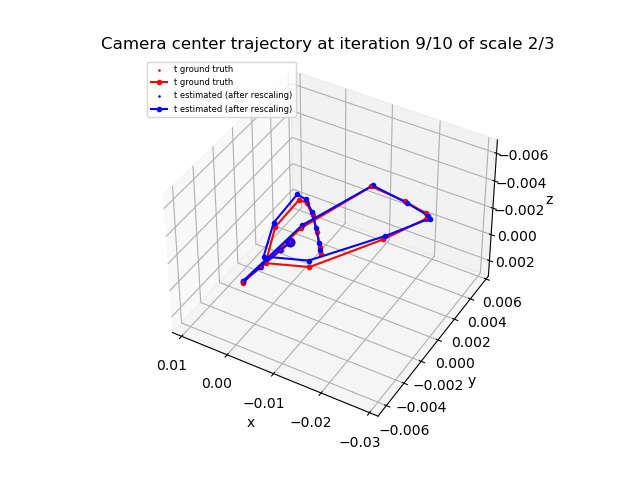}\\

            \raisebox{0.35\height}{\rotatebox[origin=l]{90}{Fig. \ref{fig:depthmap_synthetic_appendix} row 3}} &
            \includegraphics[width=\xcc\textwidth]{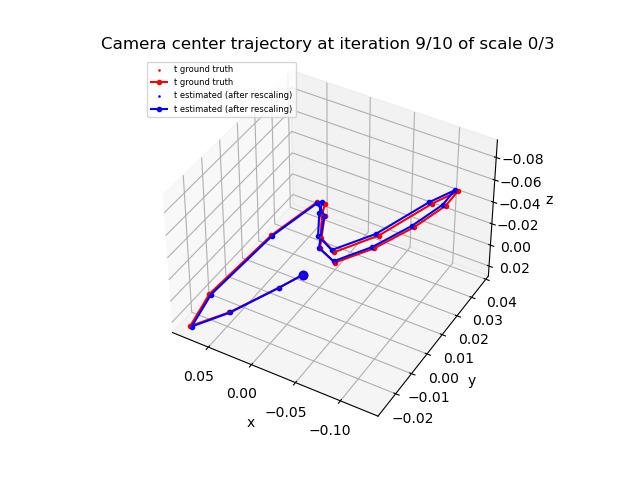} &
            \includegraphics[width=\xcc\textwidth]{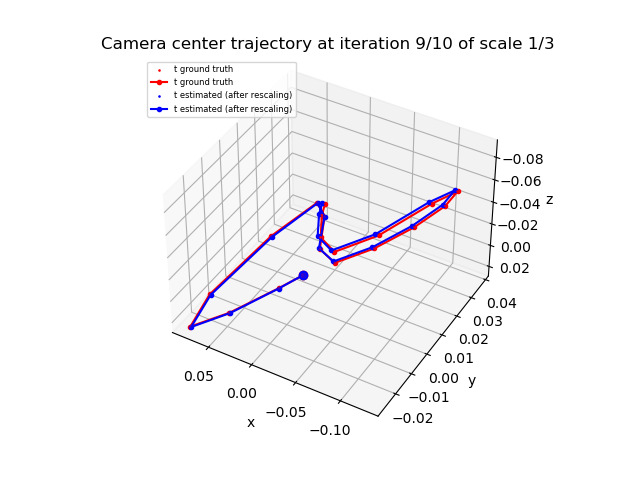} &
            \includegraphics[width=\xcc\textwidth]{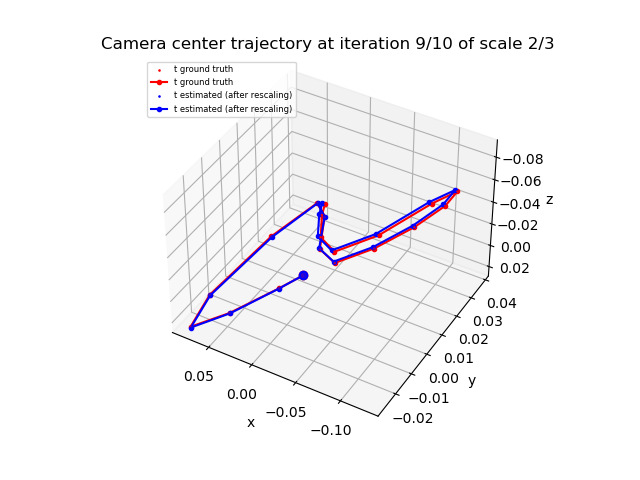}\\

 \end{tabular}
     \caption{Trajectory at different scales of the coarse to fine approach for all the scenes shown in Fig. \ref{fig:occ_mask} and Fig. \ref{fig:depthmap_synthetic_appendix}.}
     \label{fig:traj}
 \end{figure}

\subsection{Visual inspection of the registration of real frames}\label{app:anaglyph}

\begin{figure}
    \centering
    \includegraphics[width=1.\textwidth]{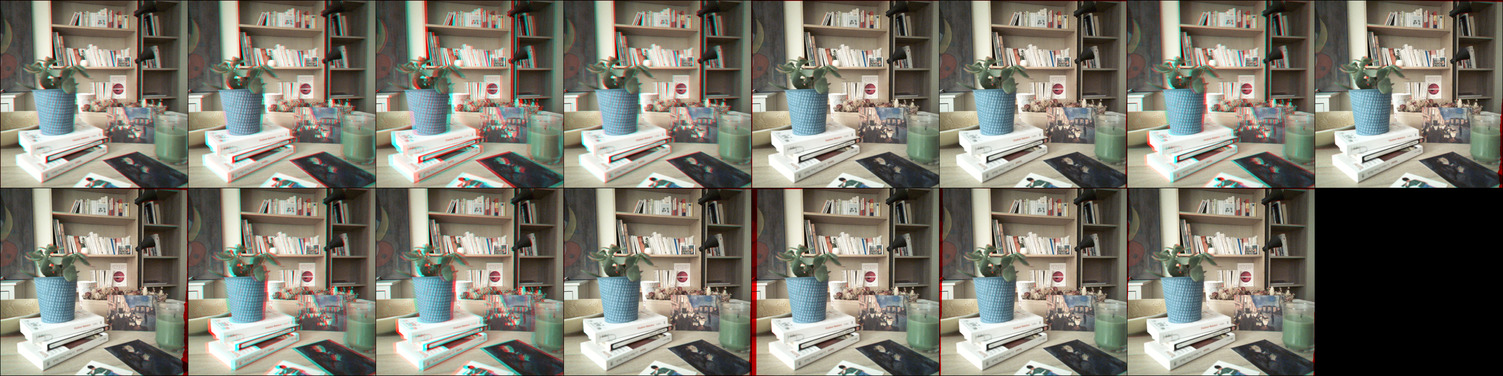}
    \caption{Qualitative alignment results of our method on a real burst. Images are generated by superimposing the
warped source images on the target image.}
    \label{fig:anaglyph}
\end{figure}

Fig. \ref{fig:anaglyph} visually demonstrates the alignment quality achieved with our method on a real burst. To assess the alignment quality, we generate images by overlaying the green and blue channels of the warped source images onto the red channel of the target image, following a similar approach as \cite{chugunov2022implicit} In this example, we observe that the majority of the frames exhibit a good alignment, while a few frames (5 out of 15) show inadequate alignment particularly in certain regions of the foreground (see for example the books or the plant). 

\subsection{Super-resolution on real bursts}\label{app:sr}

To showcase the ability of our method to produce fine alignments on real images, we perform burst super-resolution (SR) with our alignments. To achieve the task, we use the popular inverse problem framework employed in \cite{farsiu2004fast,lecouat2021lucas}. To recover the high-resolution image $\xb$ from a set of $K$ noisy and low-resolution observations $\yb_i$ with $ i \in [0,K]$ we solve the minimization problem $ \min_\xb \sum_i^K \| DBW_i \xb -\yb_i \|_2^2,$ with a gradient descent algorithm.

$D$ is a decimation operator that reduces spatial resolution, $B$ is a blurring operator, and $W$ is a warp parametrized by the optical flow. In our experiments, $DB$ is chosen as the average pooling operator following \cite{lecouat2021lucas}. 
The gradient can be derived as $ \sum_i^K W_i^\top B^\top D^\top( DBW_i \xb -\yb_i)$. 

The optical flow to warp the reference high-resolution image $\xb$ candidate is estimated in two steps using our method and then the fixed point algorithm presented in Sec. \ref{sec:meth} to infer the motion field of interest.
We perform super-resolution on RGB images in linear space demosaicked RAW frames with bilinear filtering.  Joint super-resolution and demosaicking is left for future work.

We visually compare our results in Figure \ref{fig:sr}. Our algorithm can recover fine details, including, for instance, the fine texture on the rum bottle or the hair of the doll, that were not distinguishable in the original frames.

\newcommand\www{0.22}
\begin{figure*}[htbp]
    \setlength\tabcolsep{0.5pt}
    \renewcommand{\arraystretch}{0.5}
    \centering
        \begin{tabular}{cccc}
        \multirow{2}{*}[5.5em]{\includegraphics[trim=0 0 0 0,clip,width=0.20\textwidth]{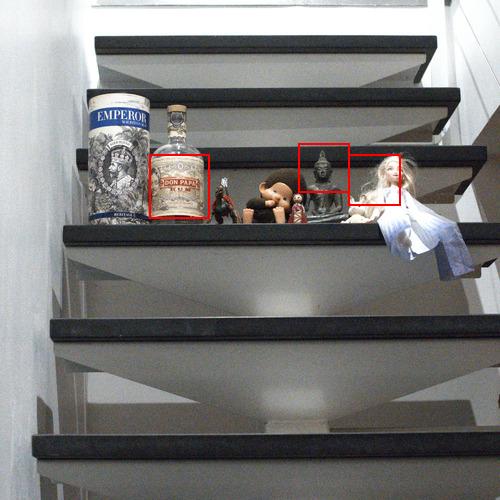}} &
            \includegraphics[trim=0 175 0 175,clip, width=\www\textwidth]{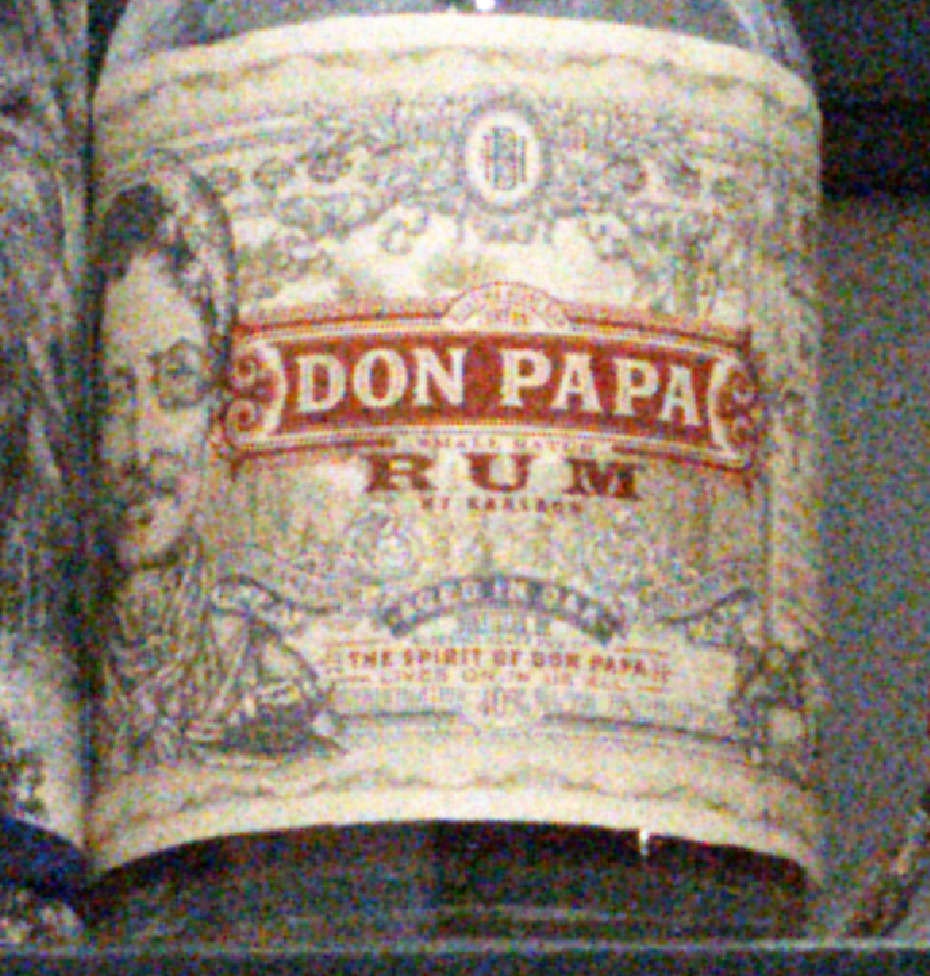} &
            \includegraphics[trim=0 225 0 125,clip, width=\www\textwidth]{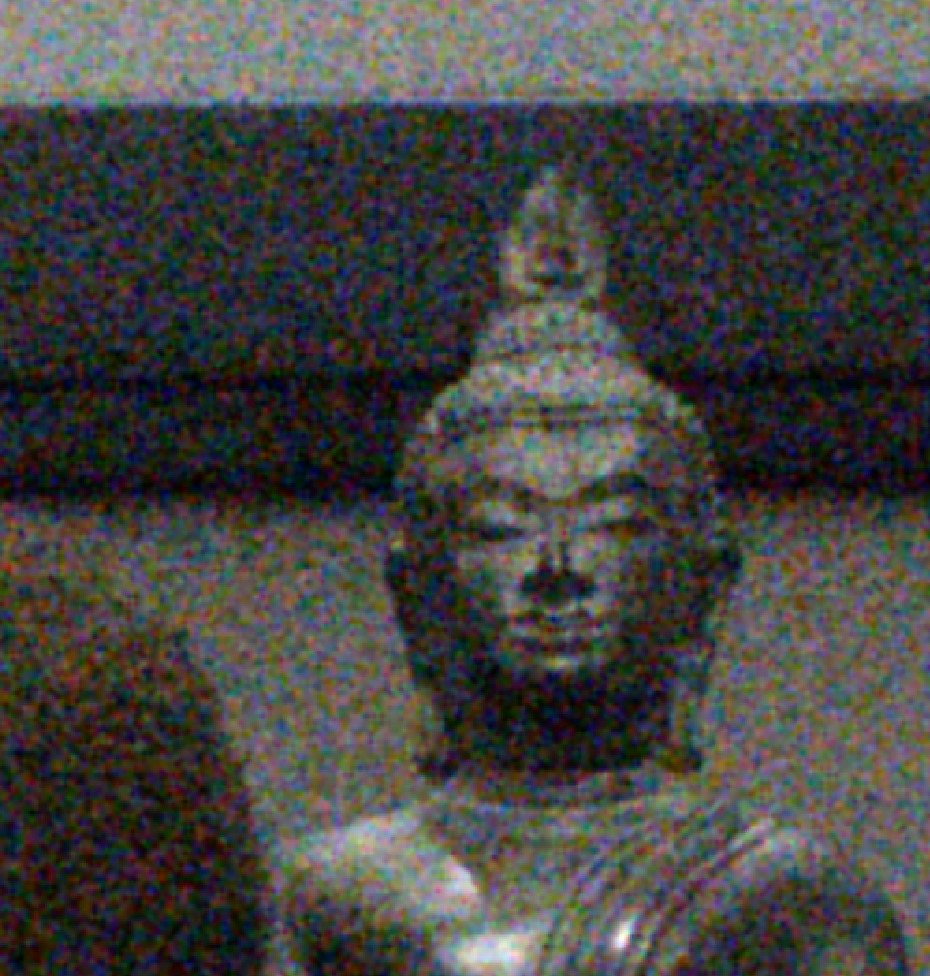} &
            \includegraphics[trim=0 175 0 175,clip, width=\www\textwidth]{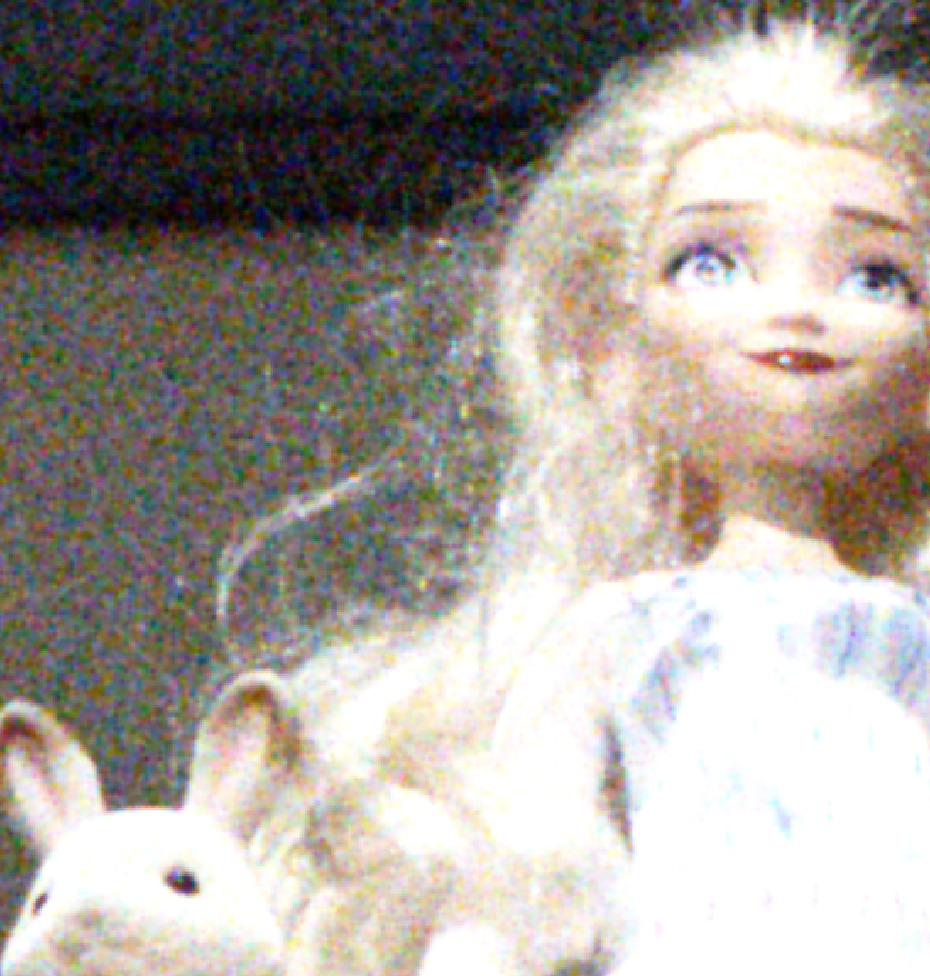} 
            \\
            &
            \includegraphics[trim=0 175 0 175,clip, width=\www\textwidth]{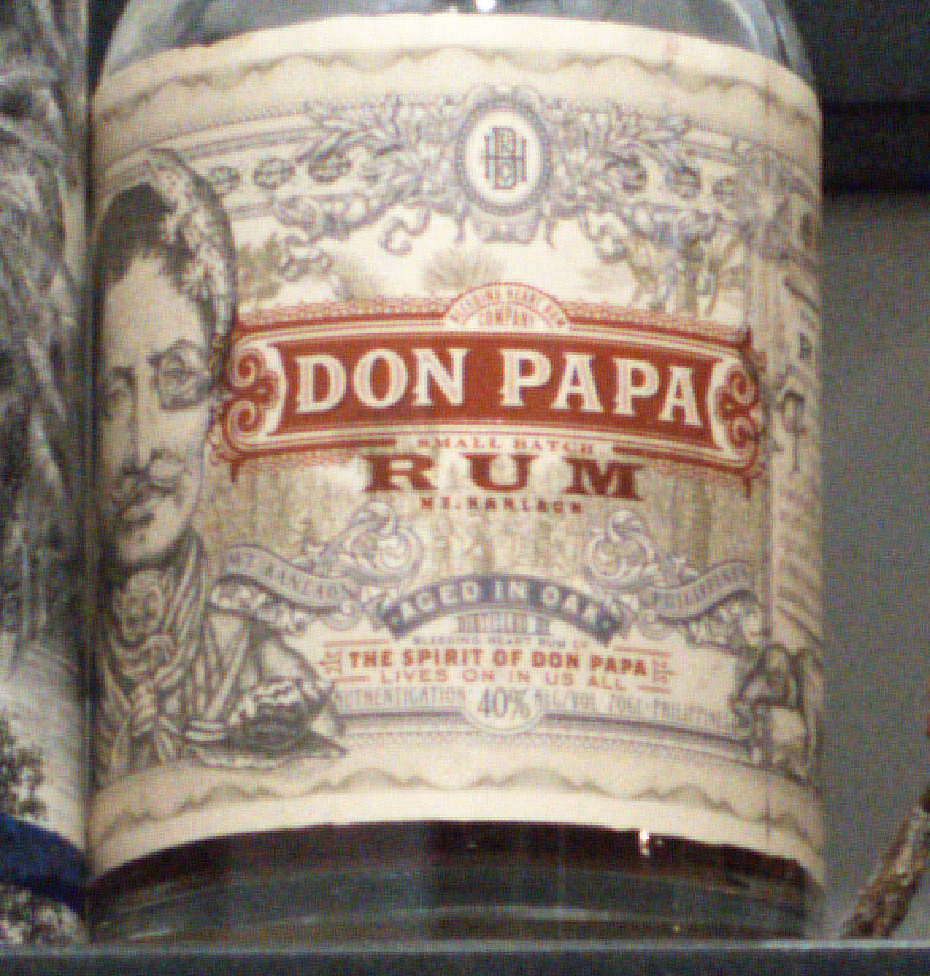} &
            \includegraphics[trim=0 225 0 125,clip, width=\www\textwidth]{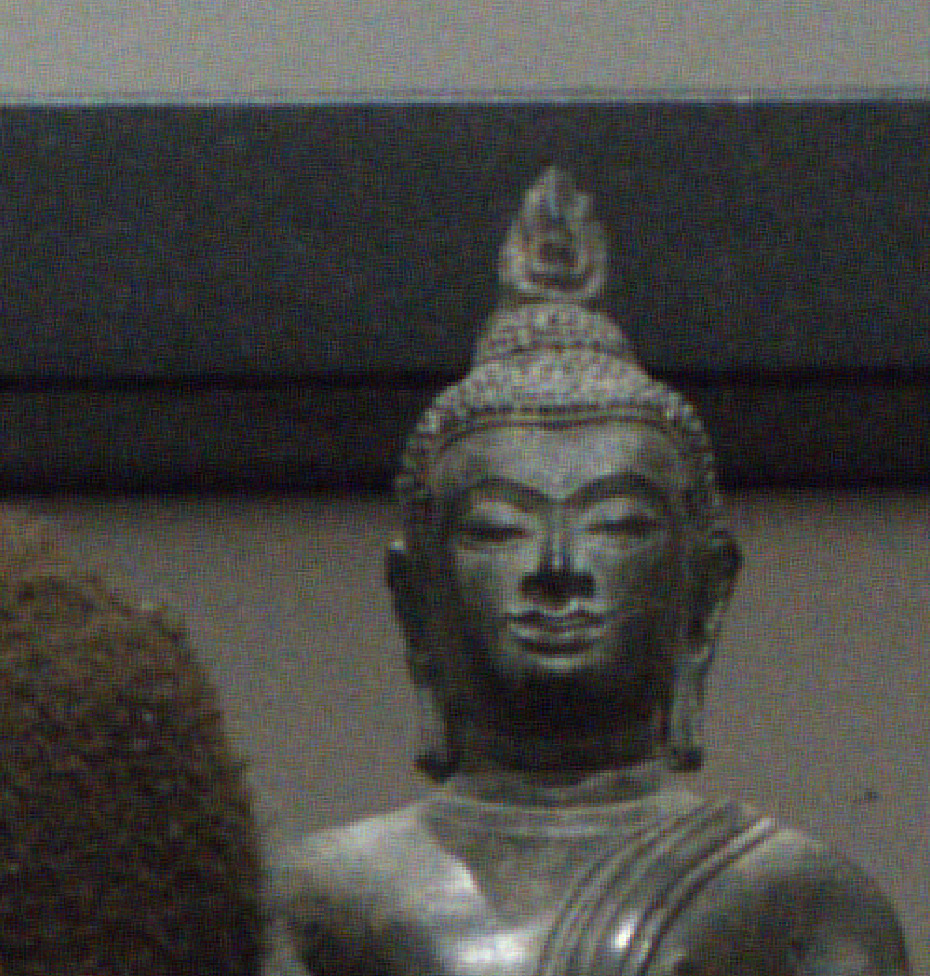} &
            \includegraphics[trim=0 175 0 175,clip, width=\www\textwidth]{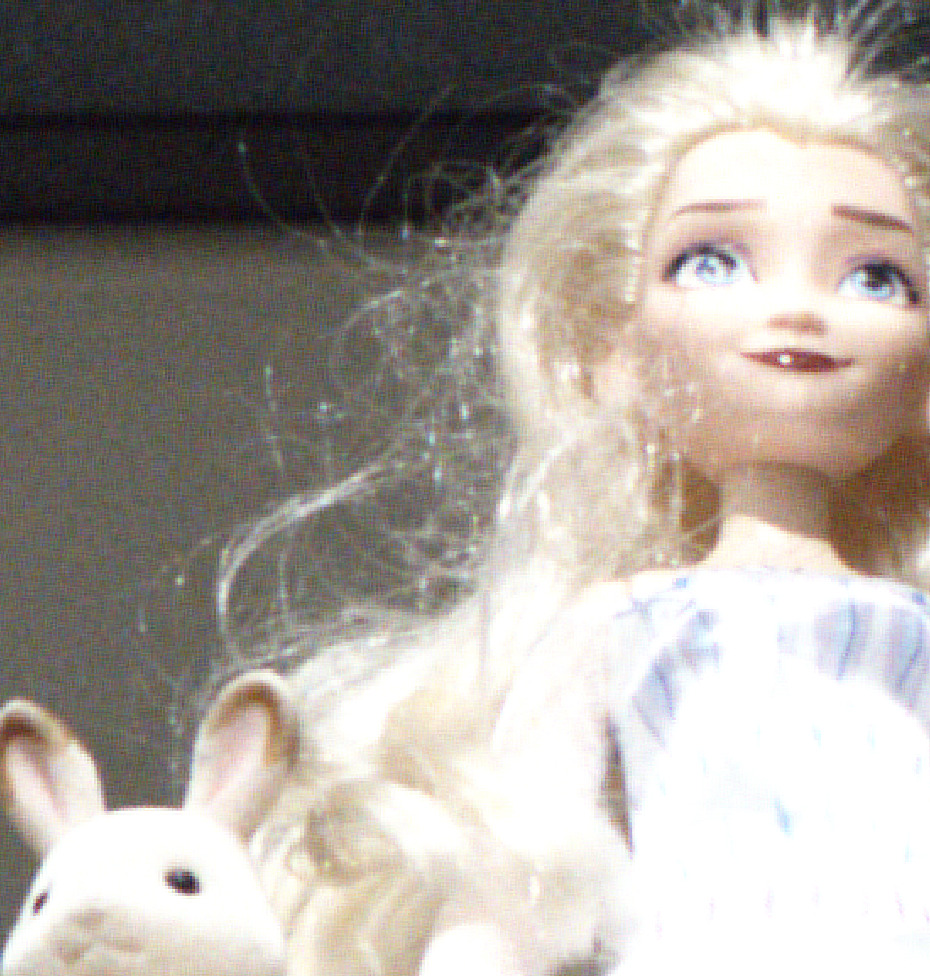} \\

\end{tabular}
    \caption{burst super-resolution on real raw bursts exploiting our alignment method. Top: low-resolution crops. Bottom: super-resolution exploiting our alignment method. Data kindly provided by the authors of \cite{lecouat2021lucas}. It is best seen by zooming aggressively on a computer screen.}
    \label{fig:sr}
\end{figure*}

\end{document}